%% file: multiviewsgm.tex
\documentclass[preprint,12pt,authoryear]{elsarticle}

\usepackage{amssymb}
\usepackage[hidelinks]{hyperref}
\usepackage{amsmath}

\input{mathops.tex}
\input{macros.tex}

\usepackage[acronyms]{glossaries} 

\usepackage[nameinlink, capitalize]{cleveref}

\usepackage{booktabs}
\usepackage{multirow}

\usepackage{subfig}

\usepackage{import} 
\usepackage{overpic}

\usepackage{url}

\usepackage{todonotes}

\usepackage[ruled,linesnumbered]{algorithm2e}
\SetKw{KwBy}{by}
\crefname{algocf}{alg.}{algs.}
\Crefname{algocf}{Algorithm}{Algorithms}
\usepackage[english]{babel}
\usepackage{blindtext}

\begin{document}

\begin{frontmatter}



\title{FaSS-MVS - Fast Multi-View Stereo with Surface-Aware Semi-Global Matching from UAV-borne Monocular Imagery}


\author[1,2]{Boitumelo Ruf\corref{cor1}}
\ead{boitumelo.ruf@kit.edu, boitumelo.ruf@iosb.fraunhofer.de}
\author[2]{Martin Weinmann}
\ead{martin.weinmann@kit.edu}
\author[2]{Stefan Hinz}
\ead{stefan.hinz@kit.edu}

\affiliation[1]{organization={Fraunhofer Institute of Optronics, System Technologies and Image Exploitation (IOSB), Fraunhofer Center for Machine Learning}, %Department and Organization
            city={Karlsruhe},
            postcode={76133}, 
            country={Germany}}
\affiliation[2]{organization={Institute of Photogrammetry and Remote Sensing, Karlsruhe Institute of Technology (KIT)}, %Department and Organization
            city={Karlsruhe},
            postcode={76133}, 
            country={Germany}}
\cortext[cor1]{Corresponding author}

\input{contents/00_abstract.tex}




\end{frontmatter}


% definition of acronyms
\input{acronyms.tex} 

\input{contents/01_introduction.tex}

\input{contents/02_related_work.tex}
\input{contents/03_methodology.tex}
\input{contents/04_experiments.tex}
\input{contents/05_discussion.tex}

\input{contents/06_conclusion.tex}


\input{contents/08_appendix.tex}

\section*{Acknowledgements}
\noindent We would like to thank the team from Arbeiter-Samariter-Bund (ASB) Baden-W\"{u}rttemberg e.V. from Karlsruhe for providing us with the use-case-specific dataset of the fire brigade exercise.

\section*{Funding}
\noindent We acknowledge support by the KIT-Publication Fund of the Karlsruhe Institute of Technology.

\bibliographystyle{elsarticle-harv} 
\bibliography{multiviewsgm.bib}

\end{document}

%% file: mathops.tex
\DeclareMathOperator*{\argmin}{arg\,min}

%% file: macros.tex
\usepackage{xspace}

\newcommand{\ie}{i.e.\ }

\newcommand{\eg}{e.g.\ }

\newcommand{\cf}{cf.\ }

\newcommand{\mm}{\,mm\xspace}

\newcommand{\m}{\,m\xspace}

\newcommand{\ms}{\,ms\xspace}

\newcommand{\s}{\,s\xspace}

\newcommand{\Hz}{\,Hz\xspace}

\newcommand{\GHz}{\,GHz\xspace}

\newcommand{\fps}{\,FPS\xspace}

\newcommand{\px}{\,pixels\xspace}

\newcommand{\percent}{\,\%\xspace}

\newcommand{\COLMAP}{COLMAP\xspace}

\newcommand{\xSGM}{SGM$^{\mathrm{x}}$\xspace}

\newcommand{\piSGM}{SGM$^{\Pi}$\xspace}

\newcommand{\fpSGM}{SGM$^{\mathrm{fp}}$\xspace}

\newcommand{\snSGM}{SGM$^{\Pi\text{-}\mathrm{sn}}$\xspace}

\newcommand{\pgSGM}{SGM$^{\Pi\text{-}\mathrm{pg}}$\xspace}

\newcommand{\LOne}{$\text{L1}$\xspace}

\newcommand{\LOneAbs}{$\text{L1-abs}$\xspace}

\newcommand{\LOneRel}{$\text{L1-rel}$\xspace}

\newcommand{\FASSMVS}{FaSS-MVS\xspace}

%% file: contents/00_abstract.tex
% KAO: Use times symbol
\begin{abstract}
With \FASSMVS, we present an approach for fast multi-view stereo with surface-aware Semi-Global Matching that allows for rapid depth and normal map estimation from monocular aerial video data captured by UAVs. %
The data estimated by \FASSMVS, in turn, facilitates online $3$D mapping, meaning that a $3$D map of the scene is immediately and incrementally generated while the image data is acquired or being received. %
\FASSMVS is comprised of a hierarchical processing scheme in which depth and normal data, as well as corresponding confidence scores, are estimated in a coarse-to-fine manner, allowing to efficiently process large scene depths which are inherent to oblique imagery captured by low-flying UAVs. %
The actual depth estimation employs a plane-sweep algorithm for dense multi-image matching to produce depth hypotheses from which the actual depth map is extracted by means of a surface-aware semi-global optimization, reducing the fronto-parallel bias of SGM. %
Given the estimated depth map, the pixel-wise surface normal information is then computed by reprojecting the depth map into a point cloud and calculating the normal vectors within a confined local neighborhood. 
In a thorough quantitative and ablative study we show that the accuracies of the $3$D information calculated by \FASSMVS is close to that of state-of-the-art approaches for offline multi-view stereo, with the error not even being one magnitude higher than that of \COLMAP. %
At the same time, however, the average run-time of \FASSMVS to estimate a single depth and normal map is less than $14$\percent of that of \COLMAP, allowing to perform an online and incremental processing of Full-HD imagery at $1$-$2$\Hz. %
\end{abstract}

%% file: acronyms.tex
\newacronym{ADAS}{ADAS}{advanced driver assistance systems}
\newacronym{ASG}{ASG}{{Average Shading Gradient}}
\newacronym{AAE}{AAE}{average angular error}
\newacronym{AR}{AR}{augmented reality}
\newacronym{ALS}{ALS}{airborne laser scanning}

\newacronym{BPU}{BPU}{{Branch Prediction Unit}}

\newacronym[shortplural={CNNs}, firstplural={convolutional neural networks (CNNs)}, longplural={convolutional neural networks}]{CNN}{CNN}{convolutional neural network}
\newacronym{COTS}{COTS}{commercial off-the-shelf}
\newacronym[shortplural={CRFs}, longplural={{Conditional Random Fields}}]{CRF}{CRF}{{Conditional Random Field}}
\newacronym{CT}{CT}{{census transform}}

\newacronym{DIM}{DIM}{dense image matching}
\newacronym{DLT}{DLT}{{Direct Linear Transformation}}
\newacronym{DoG}{DoG}{{Difference-of-Gaussian}}

\newacronym{EPnP}{EPnP}{{Efficient Perspective-n-Point}}
\newacronym{ENU}{ENU}{east-north-up}

\newacronym[shortplural={FPGAs}, longplural={field-programmable gate arrays}]{FPGA}{FPGA}{field-programmable gate array}
\newacronym{FPS}{FPS}{frames per second}
\newacronym{FPSW}{FPS/W}{FPS per watt}

\newacronym{GPGPU}{GPGPU}{general purpose computation on a {GPU}}
\newacronym[shortplural={GPUs}, longplural={graphic processing units}]{GPU}{GPU}{graphic processing unit}
\newacronym{GPS}{GPS}{{Global Positioning System}}
\newacronym{GTA}{GTA V}{{Grand Theft Auto V}}

\newacronym{HLS}{HLS}{high-level synthesis}

\newacronym{ICP}{ICP}{{Iterative-Closest-Point}}
\newacronym{IMU}{IMU}{inertial measurement unit}
\newacronym{INS}{INS}{{Inertial Navigation System}}

\newacronym{LIDAR}{LiDAR}{{Light Detection and Ranging}}
\newacronym{L1-rel}{\LOneRel}{{relative \LOne-Norm}}
\newacronym{L1-abs}{\LOneAbs}{{absolute \LOne-Norm}}

\newacronym{KLT}{KLT}{Kanade-Lucas-Tomasi}

\newacronym{MDEs}{MDE/s}{million disparity estimations per second}
\newacronym{MGM}{MGM}{More-Global Matching}
\newacronym[shortplural={MRFs}, longplural={{Markov Random Fields}}]{MRF}{MRF}{{Markov Random Field}}
\newacronym{MVS}{MVS}{multi-view stereo}

\newacronym{NCC}{NCC}{normalized cross-correlation}

\newacronym{PCL}{PCL}{{Point Cloud Library}}

\newacronym{RANSAC}{RANSAC}{{Random Sampling Consensus}}
\newacronym{RMSE}{RMSE}{{Root Mean Square Error}}
\newacronym{ROC}{ROC}{receiver operating characteristic}
\newacronym[shortplural={ROIs}, longplural={regions of interest}]{ROI}{RoI}{region of interest}

\newacronym{SAD}{SAD}{sum of absolute differences}
\newacronym{SFM}{SfM}{{Structure-from-Motion}}
\newacronym{SGBM}{SGBM}{{Semi-Global Block Matching}}
\newacronym{SGM}{SGM}{{Semi-Global Matching}}
\newacronym{SIMD}{SIMD}{{Single-Instruction-Multiple-Data}}
\newacronym{SISD}{SISD}{{Single-Instruction-Single-Data}}
\newacronym{SLAM}{SLAM}{simultaneous localization and mapping}
\newacronym{SMDE}{SMDE}{{Self-supervised Monocular Depth Estimation}}
\newacronym[shortplural={SoCs}, longplural={systems-on-a-chip}]{SoC}{SoC}{system-on-a-chip}
\newacronym{SSE}{SSE}{{Streaming SIMD Extensions}}
\newacronym{SSIM}{SSIM}{Structural Similarity}
\newacronym[shortplural={STNs}, longplural={Spatial Transformer Networks}]{STN}{STN}{{Spatial Transformer Network}}

\newacronym{TLS}{TLS}{terrestrial laser scanning}

\newacronym[shortplural={UAVs}, longplural={unmanned aerial vehicles}]{UAV}{UAV}{unmanned aerial vehicle}

\newacronym{WTA}{WTA}{winner-takes-it-all}

%% file: contents/01_introduction.tex
%%%%%%%%%%%%%%%%%%%%%%%%%%%%%%%%%%%%%%%%%%%%%%%
\section{Introduction}
\label{sec:intro}

%%%%%%%%%%%%%%%%%%%%%%%%%%%%%%%%%%%%%%%%%%%%%%%
% KAO: Sloppy spacing ensures non-overfull lines. Can be removed if this is not an issue.
\sloppy

The image-based estimation of depth maps and geometry by \gls*{DIM} and \gls*{MVS} is one of the fundamental tasks in photogrammetry, remote sensing and computer vision, facilitating a wide range of high-level applications, such as autonomous navigation, urban planning and monitoring, simulation and 3D modeling, as well as virtual, mixed and augmented reality. %
For all these applications, usually no time constraints exist, meaning that it does not matter whether the processing time is in the order of a few minutes, a few hours or sometimes even a couple of days, as long as the final result is as accurate and complete as possible. %
Especially for the task of urban monitoring and planning, a large-scale $3$D reconstruction and modeling is desired, often facilitated by additional data, \eg data captured by means of \gls*{ALS} or \gls*{TLS}, or manual post-processing. %

However, the on-going development and increasing availability of \gls*{COTS} \glspl*{UAV} open up new possibilities and applications for image-based $3$D mapping, in both offline and online processing. %
In recent years, for example, the use of \gls*{COTS} \glspl*{UAV} by emergency forces, such as the fire brigade and medical rescue services, heave steadily increased, facilitating disaster relief or search-and-rescue missions by allowing a quick and large-scale assessment of the situation or enabling the monitoring of areas which are inaccessible for ground forces \citep{Restas2015drone, Furutani2021drones}. %   
In this, image-based techniques and photogrammetry based on aerial reconnaissance are a key element in supporting the rescue workers, provided that the environmental conditions, \eg whether and daytime, allow for a visual inspection \citep{Furutani2021drones}.
There exists a large collection of software toolboxes, such as \COLMAP~\citep{Schoenberger2016sfm, Schoenberger2016mvs}, for performing offline photogrammetric $3$D reconstruction allowing to accurately reconstruct the disaster site from aerial imagery. %
Their focus, however, is primarily on offline and accurate processing. %
This hinders the use for a rapid $3$D mapping during the image acquisition, due to the required run-time for high-accurate $3$D modeling and the consideration of all input images for processing. %
However, in order to efficiently aid first responders in their tasks, the run-time of the algorithms matters which, in turn, raises the need for efficient, fast and incremental $3$D mapping in order to support a rapid assessment.
Here, the availability of $3$D data, for example, allows to reason on damage caused by an incidence, or structural integrity of a partly collapsed building, as well as route planning through areas which are difficult to access or to account for $3$D geometry when creating an orthographic map. %
In order to accommodate these applications, we propose a novel approach for fast \acrlong*{MVS} with surface-aware \gls*{SGM}, denoted as \FASSMVS. %
The approach, 
\begin{itemize}
\item uses plane-sweep sampling to perform hierarchical dense multi-image matching,
\item utilizes and extends the widely used \acrlong*{SGM} algorithm \citep{Hirschmueller2005, Hirschmueller2008} to favor not only fronto-parallel surfaces in the computation of dense depth maps, by incorporating a surface-aware regularization based on local surface normals, 
\item efficiently computes dense depth, normal and confidence maps from image sequences, allowing to facilitate the task of incremental \gls*{UAV}-borne $3$D mapping,
\item is quantitatively evaluated on two public datasets for dense \gls*{MVS} with accurate ground truth and that is demonstrated on a use-case specific dataset.
\end{itemize}
\FASSMVS combines and extends our previous work presented in \citet{Ruf2017cross, Ruf2019efficient}, by
\begin{itemize}
\item a more detailed description of the employed algorithms,
\item extending the plane-sweep multi-image matching for the use of non-fronto-parallel plane orientations,
\item improving the surface-aware regularization of the \gls*{SGM} algorithm,
\item using a different confidence measure for estimation of the confidence map,
\item a thorough evaluation and ablation study with respect to different aspects and configurations of the approach, 
\item providing a detailed discussion with respect to support of rescue workers by aerial reconnaissance. 
\end{itemize}
Even though this approach is proposed with the above-mentioned use-case in mind, it is not restricted to airborne data and can also be used to perform an incremental and online $3$D mapping of an environment captured by a ground-based robot or sensor system. % 

In the scope of this work, different terminologies with respect to the run-time capabilities and processing rate of the considered algorithms are used. %
Here, \emph{real-time processing} is used to denote a fast and low latency processing. %
This means that the processing rate of the algorithm is high enough, in order to allow an immediate reaction based on the calculated results. %
For example, in the case of reactive collision avoidance based on depth data from a stereo camera, the calculation of the scene depth needs to be fast enough to still be able to initiate an appropriate maneuver to avoid imminent collision. %
Real-time processing is deliberately not defined by a minimum processing rate, since the available reaction time depends on various factors, such as flight speed. %
In contrast, \emph{online processing} is used to denote a fast processing of the algorithm without setting hard time constraints. %
Ideally, an algorithm for online processing should be able to keep up with the frame rate of the input data. %
At the same time, however, it is not critical if the algorithm has a high latency and if the results are only available a couple of frames after the input of corresponding reference frame. %
This, for example, is the case when performing depth estimation from two or more images that are captured by a single moving camera, \ie monocular \acrlong*{MVS}. % 
Here, enough input images with appropriate baseline have first to be collected before the actual processing can be started. %
Both real-time and online processing refer to a computation during the acquisition or receiving of the input data. %
Thus, corresponding algorithms only have a confined set of input data available during execution. %
\emph{Offline} processing, on the other hand, is done fully disconnected from the actual acquisition of the input data and, thus, it is assumed that algorithms, which are executed offline, have access to all available input data. %

\subsection{Paper Outline}

This paper is structured as follows: In \Cref{sec:related_work}, the related work on incremental image-based $3$D mapping for online processing as well as modern approaches for learning-based \gls*{DIM} and \gls*{MVS} are briefly summarized. %
In this, it is also delineated, how the presented approach differs from those presented in the related work. %
In \Cref{sec:methodology}, the overall processing pipeline of the presented approach is illustrated and outlined with a quick overview. %
This is followed by a detailed description on the implementation and methodology of the individual steps of the processing pipeline. %
The approach is quantitatively and qualitatively evaluated on two public and two private datasets. %
The datasets, the error metrics as well as the results of the conducted experiments are presented in \Cref{sec:experiments}. %
Subsequently, the findings are discussed  and put into context of the considered use-case in \Cref{sec:discussion}, before providing a summary and concluding remarks as well as a short outlook on future work in \Cref{sec:conclusion}. %

%% file: contents/02_related_work.tex
%%%%%%%%%%%%%%%%%%%%%%%%%%%%%%%%%%%%%%%%%%%%%%%
\subsection{Related Work}%
\label{sec:related_work}
%%%%%%%%%%%%%%%%%%%%%%%%%%%%%%%%%%%%%%%%%%%%%%%
% KAO: Sloppy spacing ensures non-overfull lines. Can be removed if this is not an issue.
\sloppy

Due to the ever-increasing demand for detailed $3$D models, the research in the fields of photogrammetry, remote sensing and computer vision has brought up a number of software suites and applications, that focus on the estimation of accurate and dense depth and geometry information from a large set of input images, by means of \gls*{DIM} and \gls*{MVS}. %
Prominent and widely used representatives of such applications are MVE \citep{Goesele2007multi}, PMVS \citep{Furukawa2010pmvs}, SURE \citep{Rothermel2012, Wenzel2013Sure}, COLMAP \citep{Schoenberger2016mvs} and OpenMVS\footnote{\url{http://cdcseacave.github.io/openMVS}}, to name a few. %
These approaches, however, are designed for offline processing, aiming at the accuracy and completeness of the resulting $3$D model, while assuming that all input data is available at the time of reconstruction and that no critical constraints on the computation time or hardware resources are set. %
In contrast, the aim of \FASSMVS is to extract dense depth and geometry information from image sequences, while they are acquired. 
Or at least while the image data stream is received if a direct processing is not possible due to the acquisition by a small \gls*{UAV} and its limited hardware resources for example. %
Thus, the focus lies in the incremental and online processing of the input data by \gls*{DIM} and \gls*{MVS}. %
In the following, we give a brief overview of the related work on incremental camera-based mapping for online processing in \Cref{sec:related_work_incremental}. %
Due to the advancements of deep-learning-based approaches in nearly all fields of computer vision, we also provide a short overview of the related work on learning-based approaches for \gls*{DIM} and \gls*{MVS} in \Cref{sec:related_work_deepMVS}. %

\subsubsection{Incremental Camera-Based Mapping for Online Processing}
\label{sec:related_work_incremental}

Early work on incremental and online camera-based mapping of the local environment was mainly driven by robotic and \gls*{AR} applications \citep{Klein2007ptam, Davison2007monoslam, Eade2006scalable}. %
Here, the main goal was to robustly localize the camera pose, and in turn the sensor carrier, with respect to its surrounding, in order to navigate through the environment or enhance the camera images with additional information. %
Since the focus of these so-called \gls*{SLAM} algorithms is the estimation of the camera pose and trajectory, the detailed and dense mapping of the environment was rather of secondary interest.
Thus, these approaches mainly relied on point features for the tracking and mapping rather than direct pixel matching. %
However, in order to provide a convincing \gls*{AR} experience, a dense and detailed model of the environment is essential. %
Subsequent works \citep{Newcombe2010, Newcombe2011} have proposed a dense mapping simultaneous to the acquisition of the image data and the localization of the camera, resulting in a detailed reconstruction of a small \gls*{AR} workspace. %
Since these approaches, however, aim to reconstruct rather small-scale environments, they make use of short baseline video clips for the image matching, which in turn allows to rely on dense optical flow methods to find dense pixel correspondences \citep{Newcombe2010}. %
In contrast, as input to the approach presented in this work, it is assumed to have image data captured by a \gls*{UAV}, which is typically flying several tens of meters away from the object of interest. %
Thus, the presented approach is rather aimed to densely map a large-scale environment, which in turn hinders to track pixel-wise correspondences between consecutive frames, but requires a wide-baseline image matching instead. %
However, the presented approach is not solely restricted to large-scale environments and a wide baseline image matching, as experiments with respect to the employed multi-image matching, done in previous work \citep{Ruf2017cross}, show. %

Early work on camera-based mapping and reconstruction of urban surroundings was done by \citet{Gallup2007} and \citet{Pollefeys2008}, who employed the plane-sweep algorithm for true multi-image matching, first proposed by \citet{Collins1996space}, to map and reconstruct building facades from images captured by a vehicle-mounted camera in real-time. %
In this, they rely on vanishing points, which are detected in the input images, and on data from an additional \gls*{IMU} to recover the orientations of the building facades and the ground plane relative to the camera. %
To find the optimal plane configuration for each pixel and, in turn, extract a depth map from the results of the \gls*{DIM}, \citet{Pollefeys2008} employ a Bayesian formulation with a subsequent selection of the \gls*{WTA} solution, while \citet{Gallup2007} minimize a formulated energy functional. %
In \citep{Pollefeys2008}, the estimation of the camera poses is done by using a \gls*{KLT} feature tracker. %
A big advantage in the mapping and reconstruction of urban areas is that most objects in such scenery can be approximated well by planar structures, which is why plane-sweep \gls*{DIM} is well-suited for this task. %
Other approaches for urban reconstruction from ground-based imagery, like those from \citet{Furukawa2009}, \citet{Sinha2009} and \citet{Gallup2010piece}, perform a piece-wise planar reconstruction by fitting multiple, differently oriented planes into the scene and optimizing photometric consistency. %
In this, they minimize an energy functional by a graph-cut algorithm, which takes a couple of minutes on a commodity CPU. %
Algorithms having a couple of minutes run-time to estimate a single depth map do not seem to be suitable for fast and online processing at first sight. %
However, depending on their ability to be parallelized and optimized for the execution on a GPU, they might be useful after all. %

Around the same time as the previously mentioned work was released, \citet{Hirschmueller2005, Hirschmueller2008} proposed the so-called \gls*{SGM} algorithm, which evolved into one of the most widely used approaches for both online and offline \gls*{DIM}, due its efficiency and convincing results. %
It has been deployed on both desktop \citep{Spangenberg2014, Banz2011real} and embedded \citep{Hernandez2016embedded, Zhao2020fp, Ruf2021restac} hardware and is used in a wide range of applications, such as \gls*{ADAS} \citep{Spangenberg2014}, real-time obstacle detection and collision avoidance on-board \glspl*{UAV} \citep{Barry2015fpga} and urban mapping and reconstruction from aerial imagery \citep{Rothermel2012, Wenzel2013Sure, Haala2015}. %
In their work, \citet{Sinha2014} combine the plane-sweep multi-image matching with the \gls*{SGM} algorithm to estimate dense and highly accurate disparity maps. %
In contrast to the presented approach, \citet{Sinha2014} use local slanted planes, which are extracted from feature correspondences, to create disparity hypotheses and employ the \gls*{SGM} algorithm to recover a disparity map. %
They, evaluate their approach on a high-resolution stereo benchmark and achieve significant improvement over the standard \gls*{SGM} algorithm in both run-time and accuracy. %
The improvement in terms of run-time is attributed to the fact that the local plane-sweep allows to test a locally confined part of the complete disparity range for each pixel, thus reducing the computational complexity of the optimization within the \gls*{SGM} algorithm. %
Similar improvements to overcome the problem of high computational complexity due to a large disparity range, which is inherent to oblique aerial imagery, were done by \citet{Haala2015}, by embedding the \gls*{SGM} into a hierarchical coarse-to-fine processing. %

Even though a large number of urban environments can be well abstracted by piecewise planar reconstructions, not all structures are fronto-parallel, meaning that their surface orientations are not parallel to the image plane. %
In order to account for slanted surfaces, \citet{Kuschk2013}, for example, have incorporated a second-order smoothness assumption into their energy function. %
The initial formulation of the \gls*{SGM} algorithm, however, only models a first-order smoothness term and thus favors fronto-parallel surfaces, leading to stair-casing artifacts when reconstructing slanted surfaces. %
Especially when aiming for a visually appealing reconstruction of the environment, this is to be avoided. %
While \citet{Hermann2009inclusion} and \citet{Ni2018second} propose to incorporate a second-order smoothness assumption into the formulation of the \gls*{SGM} energy function, \citet{Scharstein2018surface} propose a more simplistic and yet effective improvement to address this issue. %
More specifically, plane priors are used, which, for example, can be recovered from normal maps or point correspondences, to adjust the zero-cost transition within the path aggregation of the \gls*{SGM}, thus penalizing deviations from the surface orientation represented by the prior. %
The major advantage over the other approaches is that the pixel-wise offset for the zero-cost transition can be calculated in advance and is in its magnitude the same for opposite aggregation paths, making its use very efficient. %

\subsubsection{Learning of Dense Image Matching and Multi-View Stereo Reconstruction}
\label{sec:related_work_deepMVS}

With the advancements and success of deep-learning-based methods in other topics of computer vision and photogrammetry, such as object detection, classification or image segmentation, it was just a matter of time when the first learning-based approaches for the task of \gls*{DIM} and \gls*{MVS}, that would outperform state-of-the-art model-based approaches, would be presented. %
Early works \citep{Han2015matchnet, Zbontar2016stereo, Hartmann2017learned} use deep \glspl*{CNN} to learn similarity measures between image patches and, in turn, build up a $3$D cost volume from which a disparity or depth map is extracted by conventional methods, \eg \gls*{SGM} \citep{Hirschmueller2005, Hirschmueller2008}. %

Early approaches to perform actual \gls*{MVS} with deep learning are the so-called MVSNet \citep{Yao2018mvsnet} and DeepMVS \citep{Huang2018deepmvs}. %
Both use the plane-sweep algorithm to match the pixels of multiple input images based on learned features and similarity measures and to build up a cost volume, just like the conventional approaches. %
To regularize the computed cost volume and to extract the depth map, both use a $3$D U-Net \citep{Ronneberger2015unet}. %
$3$D \glspl*{CNN}, such as the $3$D U-Net, use a great amount of memory and are computationally not very efficient, which is why other approaches, such as the ones presented in \citep{Yao2019recurrent, Yan2020dense}, exchange the $3$D U-Net by a cascade of $2$D \glspl*{CNN}. %
Further approaches \citep{Cheng2020deep,Gu2020cascade} remedy the high memory consumption by the use of hierarchical coarse-to-fine processing, as also done in our work. %
In the construction of the cost volume there also exist other strategies, such using as gated convolution \citep{Yi2020pyramid} or reprojecting the image data into a $3$D voxel grid \citep{Ji2017surfacenet}. %

What all these approaches have in common, however, is that they are trained in a supervised manner, requiring datasets with appropriate ground truth. %
Most of them are using, among others, the DTU MVS benchmark \citep{Jensen2014dtu}, which also serves as evaluation dataset in the scope of our work. %
The availability and versatility of appropriate datasets, however, is not very high, especially with respect to real-world scenarios, which still greatly hinders the practical use of deep-learning-based \gls*{MVS} approaches. %
To overcome this problem, recent approaches, such as the ones presented in \citep{Khot2019learning, Huang2021m3vsnet} try to train models in an unsupervised, or sometimes also denoted as self-supervised, manner. %
But again, their practical use and ability for generalization still needs more studies \citep{Khot2019learning}. %
These limitations are the reasons why learning-based approaches for the task of \gls*{MVS} are not yet practical for the considered use-case, namely to reliably assist emergency forces in the incremental and online mapping of the operational area. %

In summary, the presented approach adopts a plane-sweep algorithm similar to the one presented by \citet{Pollefeys2008} to perform efficient dense multi-image matching and employs an improved implementation of the \gls*{SGM} algorithm to extract the depth map from the results of the \gls*{DIM}. %
The use of a plane-sweep algorithm for the task of \gls*{DIM} is mainly motivated by its ability to create depth hypotheses by matching an arbitrary number of input images as well as the fact that it can efficiently be optimized for the massively parallel execution on \acrshortpl*{GPU}, making particularly suitable for online processing. %
In the improved implementation of the \gls*{SGM} algorithm, we, among others, adopt the approach presented by \citet{Scharstein2018surface} to account for non-fronto-parallel surfaces by adjusting the zero-cost transition based on surface information stored inside a normal map. %
Very similar to the approach presented in this work seems to be the approach from \citet{Roth2019reduction}. %
They also rely on the improvements proposed by \citet{Scharstein2018surface} and combine the \gls*{SGM} with a plane-sweep \gls*{DIM}. %
However, their work focuses on the estimation of disparity images from ground-based stereo image pairs and was only evaluated on synthetic scenes so far. %
Moreover, we also propose to reduce the fronto-parallel bias of the \gls*{SGM} algorithm, by adjusting the zero-cost transition in the path aggregation based on the gradient of the minimum cost path.
The complete depth estimation pipeline is embedded in a hierarchical processing scheme, just as proposed by \citet{Haala2015}, in order to reduce the computational complexity induced by a large scene depth inherent to oblique aerial imagery. % 

%% file: contents/03_methodology.tex
%%%%%%%%%%%%%%%%%%%%%%%%%%%%%%%%%%%%%%%%%%%%%%%
\section{Materials and Methods}%
\label{sec:methodology}
%%%%%%%%%%%%%%%%%%%%%%%%%%%%%%%%%%%%%%%%%%%%%%%

% KAO: Sloppy spacing ensures non-overfull lines. Can be removed if this is not an issue.
\sloppy

The following sections first give an overview on the full processing pipeline for an efficient \gls*{MVS} with plane-sweep multi image matching and surface-aware \gls*{SGM} (\Cref{sec:methodology_overview}). %
This is followed by a detailed description of the dense multi-image matching with plane-sweep sampling (\Cref{sec:methodology_planesweep}) and of the proposed extensions of the \gls*{SGM} algorithm for surface-aware depth map extraction (\Cref{sec:methodology_sgm}). %
In addition to the depth map, the presented approach also computes a normal map (\Cref{sec:methodology_normal}) and a confidence map (\Cref{sec:methodology_confidence}). %
In a final post-processing step (\Cref{sec:methodology_post}), further outliers are removed by masking out untextured regions and checking for geometric consistency. %

\subsection{Processing Pipeline for Fast Multi-View Stereo using Plane-Sweep Multi-Image Matching and Surface-Aware Semi-Global Optimization}
\label{sec:methodology_overview}

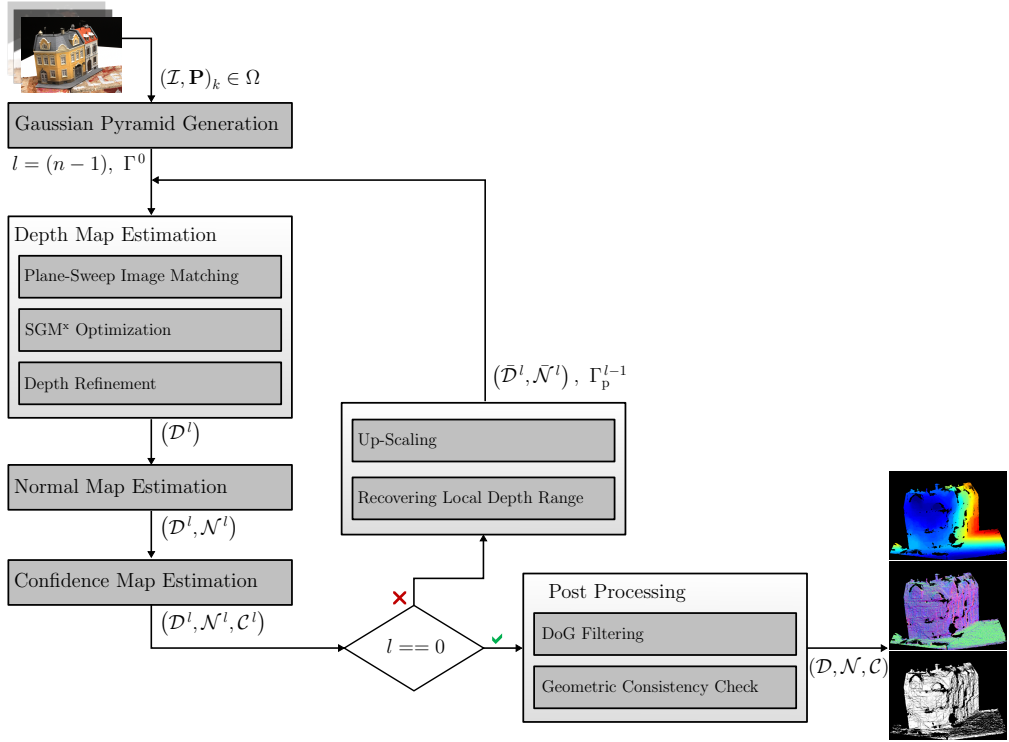
\begin{figure}[t!]%
	\centering%
	\resizebox{\columnwidth}{!}{\subimport{figures/}{methodology-edited.pdf_tex}}%
	\caption{%
		Overview of the proposed approach for incremental \gls*{MVS} with plane-sweep multi-image matching and surface-aware \gls*{SGM} optimization. %
		Given a bundle of images and corresponding camera poses $\left(\mathcal{I}, \mathbf{P}\right)_k$  of an input sequence, a hierarchical \gls*{MVS}  estimation is performed to recover a depth, normal and confidence map $\left(\mathcal{D}, \mathcal{N}, \mathcal{C}\right)$.
	}%
	\label{fig:methodology_overview}%
\end{figure}

An overview of the processing pipeline of our approach for incremental \gls*{MVS}, based on plane-sweep sampling and a surface-aware \gls*{SGM} optimization (\FASSMVS) is illustrated by \Cref{fig:methodology_overview}. %
Given an input bundle $\left(\mathcal{I}, \mathbf{P}\right)_k \in \Omega$, consisting of $k$ input images $\mathcal{I}$, extracted from an image sequence in sequential order, as well as corresponding camera poses $\mathbf{P}$, our approach computes depth, normal and confidence maps $\left(\mathcal{D}, \mathcal{N}, \mathcal{C}\right)$ for a defined reference image $\mathcal{I}_{\mathrm{ref}}$, which is typically the middle one of the input bundle $\Omega$. %
In this, we assume that the input has been calibrated, \ie that the images are free of lens distortion and that the full projection matrix $\mathbf{P}_k = \mathbf{K} \left[\mathbf{R}_k^\intercal\ \ -\mathbf{R}_k^\intercal \mathrm{C}_k \right]$ of each image is known. %
Here, $\mathrm{C}_k \in \mathbb{R}^3$ denotes the locations of the camera centers with respect to a reference coordinate system $\mathrm{O}_{\mathrm{ref}}$, while the column vectors of $\mathbf{R}^\intercal_k \in SO(3)$ hold the normalized coordinate axes of the camera coordinate system $\mathrm{O}_{\mathrm{cam}}$, as seen from $\mathrm{O}_{\mathrm{ref}}$.
The intrinsic calibration matrix $\mathbf{K}$ is equal for all cameras, since it is assumed that the images are captured from a single camera. %
Typical for an \gls*{MVS} approach, it is assumed that the input images depict the scene which is to be reconstructed from slightly different viewpoints. %

Before any processing, a Gaussian image pyramid with $n$ pyramid levels is computed for each image of the input bundle, allowing a subsequent hierarchical processing. %
The lowest pyramid levels ($l = 0$) hold the input images with their original image size. %
While moving up the pyramid, the images are first blurred with a Gaussian filter of size of $3\times 3$\px and with $\sigma = 1$, before being scaled down by a factor of $0.5$ in both image directions. %
Furthermore, the intrinsic calibration matrices are also adjusted by halving the focal length and the coordinates of the principal point, in order to account for the reduced image size. %
This results in an augmentation of the input bundle $\Omega$ by $n-1$ additional sets. %
In the following, a superscript is used to mark the results and processes at a specific pyramid level. %
The pipeline is initialized at the coarsest pyramid level $l = (n-1)$ with the smallest image size, executing three subsequent computational parts at each pyramid level, thus resulting in a coarse-to-fine processing. %

The first part of the actual processing, namely the depth estimation, computes a depth map $\mathcal{D}^l$ and is, in turn, subdivided into a plane-sweep multi-image matching creating depth hypotheses and the \xSGM optimization extracting the optimal depth from the set of hypotheses.
The latter one adopts the \gls*{SGM} approach, which was first presented by \citet{Hirschmueller2005, Hirschmueller2008}, to the plane-sweep matching, and extends it to account for non-fronto-parallel surface structures. %
\Cref{sec:methodology_planesweep} gives a detailed description on the employed plane-sweep algorithm, while the extension of the \gls*{SGM} algorithm is discussed in \Cref{sec:methodology_sgm}. %
A concluding depth refinement and median filter with a kernel size of $5\times 5$\px is used to filter small outliers in the resulting depth map. %

The second part of the hierarchical processing estimates a normal map $\mathcal{N}^l$ from the previously computed depth map $\mathcal{D}^l$. %
Apart from being an additional output of \FASSMVS, the normal map is also used as part of the \xSGM optimization to account for the surface orientation in the next hierarchical iteration. %
The normal map is regularized by an appearance-based weighted Gaussian smoothing in order to smooth out small variations while preserving discontinuities. %
\Cref{sec:methodology_normal} describes the details on how the normal map is extracted from a single depth map. %  

In the third part, a confidence map $\mathcal{C}^l$ is computed, holding pixel-wise confidence scores in the interval of $\left[0, 1\right]$ with respect to the depth estimates in $\mathcal{D}^l$. %
The final confidence scores are computed based on the surface orientation at the considered pixel. %
Details on the computation of $\mathcal{C}^l$ are given in \Cref{sec:methodology_confidence}.

Inherent to a hierarchical coarse-to-fine processing, as long as the lowest levels of the image pyramids are not yet reached, \ie while $l > 0$, the depth map $\mathcal{D}^l$ and normal map $\mathcal{N}^l$ computed at level $l$ are used to initialize the depth map estimation at the next pyramid level $l-1$. %
In this, $\mathcal{D}^l$ and $\mathcal{N}^l$ are upscaled with nearest neighbor interpolation to the image size of the next pyramid level, yielding $\bar{\mathcal{D}}^l$ and $\bar{\mathcal{N}}^l$. %
Then, $\bar{\mathcal{D}}^l$ is first used to compute the pixel-wise sampling range $\Gamma^{\,l-1}_\mathrm{p}$ of the multi-image plane-sweep algorithm at the next pyramid level. %
Here, the $\Gamma^{\,l-1}_\mathrm{p}$ is computed for each pixel $\mathrm{p}$ separately, based on the previous depth estimate $\bar{d}^{\,l}_{\,\mathrm{p}} = \bar{D}^{\,l}(\mathrm{p})$ and a predefined window with a radius of $\Delta d$ around $\bar{d}^{\,l}_{\,\mathrm{p}}$: %
\begin{equation}
\label{eq:sampling_range}
\begin{aligned}
\Gamma^{\,l-1}_\mathrm{p} &= \left[d^{\,l-1}_{\,\mathrm{p, min}}, d^{\,l-1}_{\,\mathrm{p, max}}\right], \text{with} \\
d^{\,l-1}_{\,\mathrm{p, min}} &= \bar{d}^{\,l}_{\,\mathrm{p}}-\Delta d, \\
d^{\,l-1}_{\,\mathrm{p, max}} &= \bar{d}^{\,l}_{\,\mathrm{p}}+\Delta d.
\end{aligned}
\end{equation}
In the first iteration, the sampling range is set equally for all pixels and is parameterized by minimum and maximum scene depth: $\Gamma = \left[d_{\,\mathrm{min}}, d_{\,\mathrm{max}}\right]$.
The upscaled normal map $\bar{\mathcal{N}}^l$ is used by one of the proposed \gls*{SGM} extensions to account for the surface orientation within the scene.
The final depth, normal and confidence maps are the outcome of the processing at the lowest pyramid level.
They are denoted as $\mathcal{D}$, $\mathcal{N}$ and $\mathcal{C}$ respectively, and have the same image size as the input images. %

In a final post-processing step, we use a \gls*{DoG} filter \citep{Wenzel2016dense} to unmask image regions, which do not have distinctive texture information that allows to perform a reliable matching, as well as a geometric consistency check. %
Details on the final post-processing are given in \Cref{sec:methodology_post}. %

\subsection{Real-Time Dense Multi-Image Matching with Plane-Sweep Sampling}
\label{sec:methodology_planesweep}

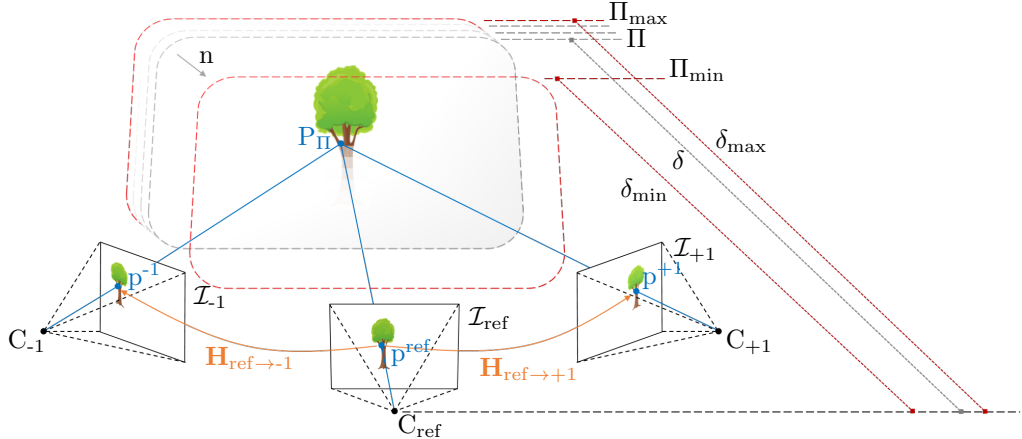
\begin{figure*}[t!]%
	\centering%	
	\def\svgwidth{\textwidth}
	\subimport{figures/}{Plane-Sweep-edited.pdf_tex}%
	\caption{%
		Overview of the algorithm for plane-sweep multi-image matching. %
		A scene is sampled by a plane $\Pi = (\mathrm{n}, \delta)$, with $\mathrm{n}$ being the normal vector of the plane and $\delta$ being the orthogonal distance of the plane from $\mathrm{C}_{\mathrm{ref}}$, that is swept along its normal vector between two bounding planes $\Pi_{\mathrm{max}}$ and $\Pi_{\mathrm{min}}$ through space. %
		For each distance $\delta$ of $\Pi$ the reference pixel $\mathrm{p}^{\mathrm{ref}}$ is projected by the plane induced homography $\mathbf{H}_{\mathrm{ref}\rightarrow k}$ into arbitrary number of viewpoints, where it is matched against the corresponding pixel in $\mathcal{I}_k$. %
	}%
	\label{fig:planesweep_overview}%
\end{figure*}

Dense image matching refers to the process of finding a dense correspondence field between pixels of two or more images from which depth hypotheses are extracted. %
In contrast to a sparse correspondence field, which is typically only established for discriminative feature points, the aim of dense image matching is to find correspondences for a majority of pixels in the reference image, usually relying on a photometric similarity measure. %
If the relative camera poses of input images are known, these correspondences can further be transformed into a depth map, which encodes the pixel-wise scene depth from the vantage point of the reference camera for which the correspondence field and depth map are computed. %
Furthermore, if intrinsic information of the reference camera as well as an absolute position with respect to a reference coordinate system are known, the depth map can be transformed into a point cloud in three-dimensional space, revealing the structure of the scene \citep{Remondino2013dense}. %

When considering the case of a calibrated stereo setup with two views, the fundamental matrix of the epipolar geometry allows for each pixel in the reference image to reduce the search space for the pixel correspondence in the matching image to the corresponding one-dimensional epipolar line, greatly reducing the complexity. %
The epipolar constraint is a key characteristic of the two-view geometry. % 
A similar characteristic exists also for the three- and four-view geometry. %
Analogously to the fundamental matrix, the trifocal and quadrifocal tensors allow to establish point-line correspondences between three and four views, respectively. %
Apart from the increasing complexity in the computation of the tensors with the number of views due to more degrees of freedom, their use, however, does not go beyond four views~\citep{Hartley2003}. %

Besides the fundamental matrix, there exists a second projective relationship between two views, which can be extended equally to an arbitrary number of views. %
Again, we consider the case of a two-camera setup with known intrinsic and extrinsic parameters. %
This time, however, an additional scene plane $\Pi= \left(\mathrm{n}, \delta\right)$ is positioned the field of view of both cameras.
Here, the scene plane $\Pi$ parameterized by its normal vector $\mathrm{n}$ and its distance $\delta$ from the first camera, \ie the reference camera.
Given this setup, the image point $\mathrm{p}^{\mathrm{ref}}$ in reference camera image can directly be mapped onto the image point $\mathrm{p}^{k}$ in the other camera image via the homography $\mathbf{H}$ induced by the plane $\Pi$ according to %
\begin{equation}
\begin{aligned}
\mathrm{p}^{k} &= \mathbf{H}\left(\Pi, \mathbf{P}_{\mathrm{ref}}, \mathbf{P}_k\right) \cdot \mathrm{p}^{\mathrm{ref}}\ ,\ \ \text{with} \\
\mathbf{H}\left(\Pi, \mathbf{P}_{\mathrm{ref}}, \mathbf{P}_k\right) &= \mathbf{K}_k \cdot \frac{\mathbf{R}-\mathrm{t}\mathrm{n}^\intercal}{\delta} \cdot \mathrm{K}_{\mathrm{ref}}^{\text{-}1} .
\end{aligned}
\label{eq:homography}
\end{equation}
Here, $\mathbf{K}_{\mathrm{ref}}$ and $\mathbf{K}_k$ denote the intrinsic matrices of both cameras, and $\left[\mathbf{R}\ \mathrm{t}\right]$ denotes the relative transformation matrix of the neighbor pose $\mathbf{P}_k$ with respect to reference pose $\mathbf{P}_{\mathrm{ref}}$. %
\Cref{eq:homography} is geometrically interpreted by casting a viewing ray through the image point $\mathrm{p}^{\mathrm{ref}}$ and intersecting it with the scene plane $\Pi$, yielding a scene point $\mathrm{P}_\Pi$, which is then projected into the second camera, resulting in the image point $\mathrm{p}^k$ \citep{Hartley2003}. %

\subsubsection{The Hierarchical Plane-Sweep Algorithm for Real-Time Multi-Image Matching}
\label{sec:methodology_planesweep_algo}

\begin{algorithm}[t!]
 \KwData{a calibrated image bundle $\Omega^l$ at the pyramid level $l$, and a set of planes $\Pi$ with a normal vector $\mathrm{n}$ and varying distances $\delta$ as well as a local depth sampling range $\Gamma^{\,l}_\mathrm{p} = [d^{\,l}_{\,\mathrm{p,min}}, d^{\,l}_{\,\mathrm{p,max}}]$. %
  }
 \KwResult{three-dimensional cost volume $\mathcal{S}$, holding the pixel-wise matching score for each pixel $\mathrm{p}^{\mathrm{ref}} \in \mathcal{I}^l_{\mathrm{ref}}$ and plane $\Pi$.}
 \BlankLine
	
 determine bounding planes $\Pi_{\mathrm{min}}$ and $\Pi_{\mathrm{max}}$ located at $\delta_{\mathrm{min}}$ and $\delta_{\mathrm{max}}$, so that the local depth range $\Gamma^{\,l}_\mathrm{p}$ is completely sampled (\cf \Cref{sec:methodology_planesweep_boundingPlanes}).
 
 \ForEach{pixel $\mathrm{p}^{\mathrm{ref}} \in \mathcal{I}^l_{\mathrm{ref}}$ {\normalfont \textbf{and}} distance $\delta \in \left[\delta_{\mathrm{min}}, \delta_{\mathrm{max}}\right]$}{ %
  
  configure scene plane $\Pi = (\mathrm{n}, \delta)$.

  determine pixels $\mathrm{p}^{\,k}$ in all matching images $\mathcal{I}^l_k \in \Omega^l \setminus \mathcal{I}^l_{\mathrm{ref}}$ : %
  $$
  	\mathrm{p}^{\,k} = \mathbf{H}\left(\Pi, \mathbf{P}^l_{\mathrm{ref}}, \mathbf{P}^l_k\right)\cdot \mathrm{p}^{\mathrm{ref}} .
  $$ %
  
  warp local image patches $\mathcal{P}^l_{k} \in \mathcal{I}^l_k$ around $\mathrm{p}^{\,k}$, with the same size as the support region of the matching cost function $C(\cdot)$, into $\mathcal{I}^l_{\mathrm{ref}}$ : %
  $$
  	\tilde{\mathcal{P}}^l_{k} = \mathbf{H}\left(\Pi, \mathbf{P}^l_{\mathrm{ref}}, \mathbf{P}^l_k\right)^{\text{-}1}\cdot \mathcal{P}^l_{k} .
  $$ %
  
  compute the matching cost $s\left(\mathrm{p}, \Pi\right)$ between reference patch $\mathcal{P}^l_{\mathrm{ref}} \in \mathcal{I}^l_{\mathrm{ref}}$ and $\tilde{\mathcal{P}}^l_{k}$ for left and right subset of cameras separately : %
  $$
  	s^{\mathrm{L}}\left(\mathrm{p}, \Pi\right) = \sum_{k < \mathrm{ref}} C\left(\mathcal{P}^l_{\mathrm{ref}}, \tilde{\mathcal{P}}^l_{k}\right) ,
  $$
  $$
 	s^{\mathrm{R}}\left(\mathrm{p}, \Pi\right) = \sum_{k > \mathrm{ref}} C\left(\mathcal{P}^l_{\mathrm{ref}}, \tilde{\mathcal{P}}^l_{k}\right) .
  $$ %
  
  store the minimum of left and right matching cost (accounting for occlusions as described by \citet{Kang2001}) into three-dimensional cost volume $\mathcal{S}$ : %
  $$
  	\mathcal{S}^l\left(\mathrm{p}, \Pi\right) = \min\lbrace s^{\mathrm{L}}\left(\mathrm{p}, \Pi\right), s^{\mathrm{R}}\left(\mathrm{p}, \Pi\right) \rbrace .
  $$ %  
 }
 \caption{Plane-sweep multi-image matching executed at a specific pyramid level $l$ of the proposed hierarchical processing scheme.}
 \label{alg:plane_sweep}
\end{algorithm}

Based on the relationship between two cameras and a scene plane, \citet{Collins1996space} proposed an algorithm for true multi-image matching.
This algorithm samples the scene-space between two bounding planes $\Pi_{\mathrm{min}}$ and $\Pi_{\mathrm{max}}$, located at $\delta_{\mathrm{min}}$ and $\delta_{\mathrm{max}}$, by sweeping a plane along its normal vector $\mathrm{n}$ through space and matching the input images according to \Cref{eq:homography} for each distance~$\delta~\in~\left[\delta_{\mathrm{min}},~\delta_{\mathrm{max}}\right]$ of the plane relative to the reference camera. %
For each position of the plane, an arbitrary number of matching images are warped by the plane-induced homography $\mathbf{H}^{\text{-}1}_{\mathrm{ref}\rightarrow k}$ into the view of the reference camera, where they are matched against the reference image.
If the scene plane is close to a three-dimensional structure, then the corresponding image regions of the warped matching images overlap with the projection in the reference image, allowing to deduce the scene depth of the corresponding object from the parameterization of the corresponding plane (\cf \Cref{fig:planesweep_overview}).
First denoted as space-sweep algorithm, it was adopted by numerous studies on multi-image matching and \gls*{MVS} \citep{Gallup2007, Pollefeys2008, Sinha2014}, eventually denoting it as plane-sweep algorithm. %
The presented approach for hierarchical multi-image matching is based on the plane-sweep algorithm presented by \citet{Pollefeys2008} and it is described in \Cref{alg:plane_sweep}.%

As part of the actual image matching, the Hamming distance of the \gls*{CT} \citep{Zabih1994} as well as a negated, truncated and scaled form of the \gls*{NCC} \citep{Scharstein2018surface, Sinha2014} are employed and evaluated as cost function $C(\cdot)$. %
And since the approach considers a bundle of input images with an equal number of matching images to either sides of the reference image, the approach presented by \citet{Kang2001} is adopted, using the minimum aggregated matching cost of the left and right subset of the matching images in order to account for occlusions. %
The resulting three-dimensional cost volume $\mathcal{S}^{\,l}$ is of size $w^{\,l} \times h^{\,l} \times |\delta^{\,l}|$, with $w^{\,l}$ and $h^{\,l}$ being the width and height of the reference image and $|\delta^{\,l}|$ being the number plane positions at which the matching is performed, all with respect to the current pyramid level $l$. %
In this, the cost volume $\mathcal{S}^{\,l}$ is implemented as a dynamic cost volume \citep{Haala2015} for all but the topmost pyramid level, since the sampling range $\Gamma^{\,l}_{\mathrm{p}}$ is determined independently for each pixel $\mathrm{p}$. %
Nonetheless, the complete set of plane distances $\delta \in [\delta_{\mathrm{min}}, \delta_{\mathrm{max}}]$, deduced from $\Gamma$, are precomputed for each pyramid level $l$ and the same for all pixels. % 
This, in turn, allows to precompute the homographic mappings for all planes $\Pi$. %
In the following, we describe how we find the bounding planes $\Pi_{\mathrm{min}}$ and $\Pi_{\mathrm{max}}$ and the corresponding distances $\delta_{\mathrm{min}}$ and $\delta_{\mathrm{max}}$ (\Cref{sec:methodology_planesweep_boundingPlanes}), as well as how we use the cross-ratio to find appropriate sampling steps within $\Gamma$ (\Cref{sec:methodology_planesweep_crossratio}). %

\subsubsection{Determining the Bounding Planes Corresponding to the Given Depth Range}
\label{sec:methodology_planesweep_boundingPlanes}

As previously described, it is assumed that two bounding planes, namely $\Pi_{\mathrm{min}}$ and $\Pi_{\mathrm{max}}$ with corresponding distances $\delta_{\mathrm{min}}$ and $\delta_{\mathrm{max}}$, in between which the scene is to be sampled, are known. %
In case of a fronto-parallel sampling strategy, \ie $\mathrm{n} = (0\ 0\ \text{-}1)^\intercal$ with respect to the local camera coordinate system, the distances $\delta_{\mathrm{min}}$ and $\delta_{\mathrm{max}}$ are equal to the minimum and maximum depth, namely $d_{\mathrm{min}}$ and $d_{\mathrm{max}}$. %
This, however, does not hold for non-fronto-parallel plane orientations. %
To find the bounding planes for slanted plane orientations, a view-frustum, corresponding to the reference camera, is first constructed for which the depth is to be estimated. %
This view-frustum is represented by a pyramid that resembles the field-of-view of the camera and that is truncated by two fronto-parallel near and far planes that are located at $d_{\mathrm{min}}$ and $d_{\mathrm{max}}$. %
Given the four corner points of the view-frustum on the near plane $\mathrm{X}^{\mathrm{near}}_i$ and the four on the far plane $\mathrm{X}^{\mathrm{far}}_i$, the minimum and maximum distance $\delta_{\mathrm{min}}$ and $\delta_{\mathrm{max}}$ are found according to:
\begin{equation}
\begin{aligned}
\delta_{\mathrm{min}} &= \min\limits_{i}(|\mathrm{n}^{\intercal}\cdot\mathrm{X}^{\mathrm{near}}_i|)\ ,\ \ \text{and} \\
\delta_{\mathrm{max}} &= \min\limits_{i}(|\mathrm{n}^{\intercal}\cdot\mathrm{X}^{\mathrm{far}}_i|). \\
\end{aligned}
\end{equation}
In order to avoid an orientation flip of the image data, all camera centers $\mathrm{C}_i$ need to lie in front of $\Pi_{\mathrm{min}}$ with respect to the sweeping direction, thus for all camera centers $\mathrm{n}^{\intercal}\cdot\mathrm{C}_i + \delta_{\mathrm{min}} > 0$ must hold. %

\subsubsection{Finding the Sampling Steps by Utilizing the Cross-Ratio}
\label{sec:methodology_planesweep_crossratio}

As stated by \Cref{eq:homography}, the sampling planes $\Pi$ of the plane-sweep algorithm are parameterized by two parameters, namely the normal vector $\mathrm{n}$, denoting the orientation and sweeping direction of the plane, and the orthogonal distance $\delta$ from the optical center of the reference camera $\mathrm{C}_{\mathrm{ref}}$. %
While $\mathrm{n}$ allows to adjust the warping of the images and, thus, the image matching to the surface orientation within the scene, the second parameter $\delta$ determines the step-size with which the scene is sampled. %
In this, a straight-forward approach would be to select the step-size in such a way that the scene is sampled with a desired resolution, \ie sweeping the planes at equidistant unit intervals through the scene space. %
However, it is not guaranteed that a thorough sampling of the scene with a small step-size results in a higher accuracy. %
If the step-size is not chosen in accordance with the camera positions of the input images and the baseline between the cameras, the matching results of two or more consecutive plane positions might not reveal enough difference and, thus, introduce ambiguities between multiple plane hypotheses. %
Furthermore, in terms of efficiency, it is important to vary the sampling rate in scene space with respect to the distance of the plane relative to the reference camera, since the perspective projection requires an increasingly smaller step-size as the plane moves closer to the camera. %

\begin{algorithm}[t!]
 \KwData{two cameras with full projection matrices $\mathbf{P}_\mathrm{ref}$ and $\mathbf{P}_k$, an image point $\mathrm{p}^{\mathrm{ref}}$ inducing largest disparity when warped from $\mathcal{I}_{\mathrm{ref}}$ to $\mathcal{I}_k$, as well as two bounding planes $\Pi_{\mathrm{min}}$ and $\Pi_{\mathrm{max}}$.} %
 \KwResult{list of orthogonal plane distances $\delta$ relative to $\mathrm{C}_{\mathrm{ref}}$, such that the maximum pixel-displacement between the warped images of two consecutive planes is less than or equal to $1$.}
 \BlankLine
 calculate the viewing ray $\mathrm{V}^{\mathrm{ref}}_{\mathrm{p}}$, going through $\mathrm{C}_\mathrm{ref}$ and $\mathrm{p}^{\mathrm{ref}}$, and intersect it with $\Pi_{\mathrm{min}}$ and $\Pi_{\mathrm{max}}$, yielding the scene points $\mathrm{P}_{\mathrm{min}}$ and $\mathrm{P}_{\mathrm{max}}$. %
 
 project the optical center $\mathrm{C}_\mathrm{ref}$, as well as $\mathrm{P}_{\mathrm{min}}$ and $\mathrm{P}_{\mathrm{max}}$ onto the image plane of the second camera, yielding the epipole $\mathrm{e}^k_{\mathrm{ref}}$ and the two image points $\mathrm{p}^k_{\mathrm{min}}$ and $\mathrm{p}^k_{\mathrm{max}}$, all lying on the epipolar line $\mathrm{l}^k_{\mathrm{p}}$.
 
 determine the unit vector $\mathrm{k} = \frac{\mathrm{p}^k_{\mathrm{min}} - \mathrm{p}^k_{\mathrm{max}}}{||\mathrm{p}^k_{\mathrm{min}} - \mathrm{p}^k_{\mathrm{max}}||}$, being the normalized direction of $\mathrm{l}^k_{\mathrm{p}}$ and pointing from $\mathrm{m}^k_{\mathrm{max}}$ to $\mathrm{m}^k_{\mathrm{min}}$.
 
 \For{$\mathrm{p}^k_i \gets \mathrm{p}^k_{\mathrm{max}}$ \KwTo $\mathrm{p}^k_{\mathrm{min}}$ \KwBy $\mathrm{p}^k_{i+1} = \mathrm{p}^k_{i} + \mathrm{k}$}{
 	given the viewing rays $\mathrm{V}^k_{\mathrm{e}_{\mathrm{ref}}}$, $\mathrm{V}^k_{\mathrm{p}_{\mathrm{min}}}$, $\mathrm{V}^k_{\mathrm{p}_i}$ and  $\mathrm{V}^k_{\mathrm{p}_{\mathrm{max}}}$ going through the optical center of $\mathrm{C}_k$ and $\mathrm{e}^k_{\mathrm{ref}}$, $\mathrm{p}^k_{\mathrm{min}}$, $\mathrm{p}^k_i $ and $\mathrm{p}^k_{\mathrm{max}}$ respectively, apply \Cref{eq:doppelverhaeltnisA} and \Cref{eq:doppelverhaeltnisB} to compute $\mathrm{P}_i \in \mathrm{V}^{\mathrm{ref}}_{\mathrm{p}}$ according to:
 	$$
	\begin{aligned}
	Q(\mathrm{V}^k_{\mathrm{e}_{\mathrm{ref}}}, \mathrm{V}^k_{\mathrm{p}_{\mathrm{min}}}, \mathrm{V}^k_{\mathrm{p}_i}, \mathrm{V}^k_{\mathrm{p}_{\mathrm{max}}})
	&= \frac{\sin(\alpha(\mathrm{V}^k_{\mathrm{e}_{\mathrm{ref}}}, \mathrm{V}^k_{\mathrm{p}_i})) \cdot \sin(\alpha(\mathrm{V}^k_{\mathrm{p}_{\mathrm{min}}},\mathrm{V}^k_{\mathrm{p}_{\mathrm{max}}}))}{\sin(\alpha(\mathrm{V}^k_{\mathrm{e}_{\mathrm{ref}}},\mathrm{V}^k_{\mathrm{p}_{\mathrm{max}}})) \cdot \sin(\alpha(\mathrm{V}^k_{\mathrm{p}_{\mathrm{min}}},\mathrm{V}^k_{\mathrm{p}_i}))} \\
	&= \frac{\Delta(\mathrm{C}_{\mathrm{ref}},\mathrm{P}_i) \cdot \Delta(\mathrm{P}_{\mathrm{min}},\mathrm{P}_{\mathrm{max}})}{\Delta(\mathrm{C}_{\mathrm{ref}},\mathrm{P}_\mathrm{{max}}) \cdot \Delta(\mathrm{P}_{\mathrm{min}},\mathrm{P}_i)} .
	\end{aligned}
	$$
 	
 	since $Q(\mathrm{C}_{\mathrm{ref}}, \mathrm{P}_{\mathrm{min}}, \mathrm{P}_i, \mathrm{P}_{\mathrm{max}}) = Q(\mathrm{C}_{\mathrm{ref}}, \delta_{\mathrm{min}}, \delta, \delta_{\mathrm{max}})$, derive $\delta$ relative to $\mathrm{C}_{\mathrm{ref}}$ according to:
	$$
	\begin{aligned}
	\frac{\delta \cdot \left(\delta_\mathrm{max}-\delta_\mathrm{min}\right)}{\delta_\mathrm{max} \cdot\left(\delta -\delta_\mathrm{min}\right)} 
	&=\frac{\sin(\alpha(\mathrm{V}^k_{\mathrm{e}_{\mathrm{ref}}}, \mathrm{V}^k_{\mathrm{p}_i})) \cdot \sin(\alpha(\mathrm{V}^k_{\mathrm{p}_{\mathrm{min}}},\mathrm{V}^k_{\mathrm{p}_{\mathrm{max}}}))}{\sin(\alpha(\mathrm{V}^k_{\mathrm{e}_{\mathrm{ref}}},\mathrm{V}^k_{\mathrm{p}_{\mathrm{max}}})) \cdot \sin(\alpha(\mathrm{V}^k_{\mathrm{p}_{\mathrm{min}}},\mathrm{V}^k_{\mathrm{p}_i}))} .
	\end{aligned}
	$$
 }
 \caption{Finding plane distances $\delta$ by utilizing the cross-ratio.}
 \label{alg:finding_plane_postions}
 \vspace{5mm}
\end{algorithm}

Thus, a common approach is to select the sampling positions of the planes according to the disparity change induced by two consecutive planes. %
In this, the pixel-wise motion between the warped images of two consecutive planes should not exceed an absolute value of $1$ \citep{Szeliski2004, Pollefeys2008}. %
To ensure that the maximum disparity change between the warped images of two consecutive planes is less or equal to $1$, \citet{Pollefeys2008} evaluate the displacements occurring at the boundaries of the most distant images for a set of planes with predefined parameters and only select those that fulfill the stated criterion for the actual image matching. %
Thus, for each plane within the predefined set, an additional test is performed, involving the warping of image points at the boundaries, in order to determine whether the plane is suitable or not. %

In contrast, with this approach, we aim to directly derive the distances of the sampling planes from correspondences in image space, implicitly leading to an inverse depth sampling in scene space. %
Given an image point $\mathrm{p}^{\mathrm{ref}} \in \mathcal{I}_{\mathrm{ref}}$ in the reference image, an intuitive approach would be to select multiple sampling points $\mathrm{p}^k \in \mathcal{I}_k$ on the corresponding epipolar line $\mathrm{l}^k_{\mathrm{p}}$ in one of the other cameras, and find the corresponding plane distances by triangulation between $\mathrm{p}^{\mathrm{ref}}$ and $\mathrm{p}^k$. %
The effort of triangulation, however, can be avoided by relying on the cross-ratio which is invariant under perspective projection (\cf \ref{sec:crossratio}). %
%The characteristics of the cross-ratio and an explanation on how it aids to determine the sampling points of the plane-sweep algorithm are provided in the following section. %

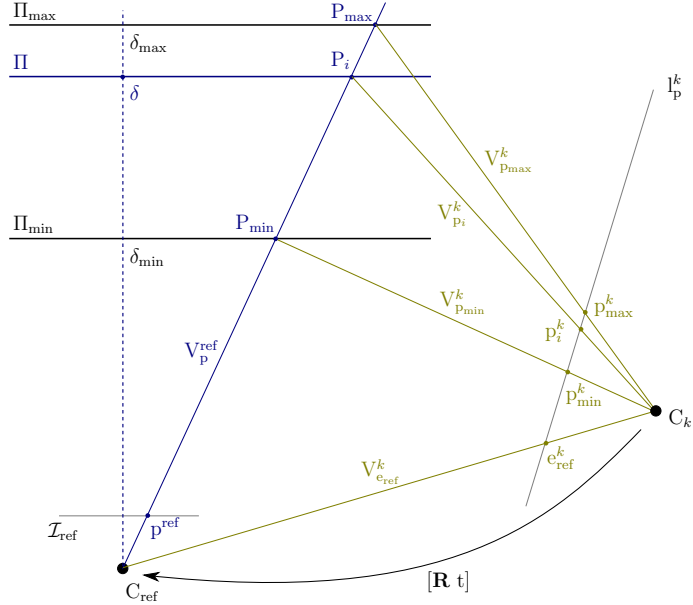
\begin{figure}[t!]%
	\centering
	\def\svgwidth{\columnwidth}
	\resizebox{0.66\columnwidth}{!}{\subimport{figures/}{ComputeDistanceOfPlane-edited.pdf_tex}}%
	\caption{Determination of the orthogonal distance parameter of the sampling planes of the plane-sweep multi-image matching by using the cross-ratio and epipolar geometry.
	Here, $\mathrm{C}_{\mathrm{ref}}$ and $\mathrm{C}_k$ represent the positions of the optical centers of the two cameras. %
	}%
	\label{fig:planeDistance}%
\end{figure}

%In the following, we will discuss how, in our approach, we use the cross-ratio to determine the plane distances as part of the plane sweep algorithm. %
Given the projection matrices $\mathbf{P}_\mathrm{ref}$ and $\mathbf{P}_k$ of two cameras, as illustrated in \Cref{fig:planeDistance}, we center the coordinate system in the optical center of the reference camera $\mathrm{C}_\mathrm{ref}$, so that $\mathbf{P}_\mathrm{ref} = \mathbf{K} \left[\mathbf{I}\ 0\right].$ %
Thus, we are aiming to find the plane distances relative to $\mathrm{C}_\mathrm{ref}$. %
In case of \gls*{MVS} with multiple cameras, for $\mathrm{C}_k$, we select the camera which will induce the largest image offset, and thus giving an upper bound on the disparity range. %
As already noted by \citet{Pollefeys2008}, this is typically the camera which is most distant from the reference camera. %

Furthermore, we again assume to have two bounding planes $\Pi_{\mathrm{max}}$ and $\Pi_{\mathrm{min}}$, which limit the sweep space of the planes. %
Just as in the work of \citet{Pollefeys2008}, it is ensured that the sweep space does not intersect the convex hull of the cameras, in order to avoid inversions in the process of image matching (\cf \Cref{sec:methodology_planesweep_boundingPlanes}). %
We choose $\mathrm{p}^{\mathrm{ref}}$ as the pixel which induces the largest disparity when warped from $\mathcal{I}_{\mathrm{ref}}$ to $\mathcal{I}_k$ via $\mathbf{H}\left(\Pi_{\mathrm{min}}, \mathbf{P}_{\mathrm{ref}}, \mathbf{P}_k\right)$, typically one of the four corners, in order to guarantee a maximum disparity change between successive planes and find a set of sampling planes, with distances $\delta$ relative to $\mathrm{C}_{\mathrm{ref}}$ according to \Cref{alg:finding_plane_postions} and illustrated by \Cref{fig:planeDistance}. %

This approach is computationally efficient and not restricted to a fronto-parallel orientation of the sampling planes as long as the optical axis of the reference camera intersects with the planes and the sweeping vector has a component that is parallel to the optical axis. % 
Furthermore, in order to accommodate for all possible setups of $\mathrm{C}_{\mathrm{ref}}$ and $\mathrm{C}_k$, it is important to use $Q(\mathrm{V}^k_{\mathrm{e}_{\mathrm{ref}}}, \mathrm{V}^k_{\mathrm{p}_{\mathrm{min}}}, \mathrm{V}^k_{\mathrm{p}_i}, \mathrm{V}^k_{\mathrm{p}_{\mathrm{max}}})$ in \Cref{alg:finding_plane_postions}, since $\mathrm{e}^k_{\mathrm{ref}}$ would flip to the side of $\mathrm{p}^k_{\mathrm{max}}$ if the focal plane of reference camera is behind $\mathrm{C}_k$. %

\subsection{Depth Map Computation with Surface-Aware Semi-Global Matching}
\label{sec:methodology_sgm}

The hierarchical plane-sweep algorithm for multi-image matching, as described in \Cref{sec:methodology_planesweep}, produces at each pyramid level a three-dimensional cost volume $\mathcal{S}^l(\mathrm{p}, \Pi)$, which holds matching costs for each pixel $\mathrm{p} \in \mathcal{I}_{\mathrm{ref}}$ induced by a given plane $\Pi$ located at distance $\delta$ orthogonal to the optical center $\mathrm{C}_{\mathrm{ref}}$ of the reference camera. 
In the second stage of the depth estimation within \FASSMVS, the cost volume is regularized by a semi-global optimization scheme, yielding a dense depth map $\mathcal{D}^l$. %
Building upon the original \gls*{SGM} approach \citep{Hirschmueller2005, Hirschmueller2008}, we propose three different optimization schemes (\xSGM). %
Apart from a straight-forward adaptation of the \gls*{SGM} approach to the plane-sweep sampling, we also adopt the approach of \citet{Scharstein2018surface} to also favor slanted surfaces by considering surface information available in the form of surface normals. %
Furthermore, we investigate a third extension, which penalizes deviations from the gradient of the minimum cost path within the \gls*{SGM} optimization scheme. %
In the following, we first recap the \gls*{SGM} algorithm in \Cref{sec:methodology_sgm_algo}, before describing the proposed extension in detail in \Cref{sec:methodology_sgm_x}. %

\subsubsection{Semi-Global Matching}
\label{sec:methodology_sgm_algo}

\glsreset{SGM}
\glsreset{DIM}

In early studies on \gls*{DIM}, \citet{Scharstein2002} have grouped the stereo algorithms into three groups based on their optimization strategy, \ie the approach on how to extract the final disparity or depth map from the depth hypotheses produced by the image matching. %
The first group contains all algorithms, which employ a local optimization scheme. %
In this, a locally confined support region is used to find the pixel-wise optimal depth, \ie the \gls*{WTA} solution, from the cost volume. %
While local algorithms are computationally very efficient, they only model an implicit smoothness assumption within the local window. %
Global methods on the other hand, which make up the second category of the taxonomy presented by \citet{Scharstein2002}, formulate explicit smoothness assumptions as part of an energy function, which is to be minimized in a global optimization scheme. %
This, in turn, however, makes such algorithms computationally more expensive, while at the same time producing more accurate results than local methods. %

The third category, to which the \gls*{SGM} algorithm of \citet{Hirschmueller2005, Hirschmueller2008} belongs to, is a subgroup to the global methods. %
It contains algorithms, which explicitly formulate the stereo problem in a global energy function, yet employ dynamic programming to approximate the computation of the optimal solution. %
In dynamic programming, the original optimization problem is broken down into a number of smaller, less complex problems, which can be processed and solved independently of each other.
The solution of the initial problem is then the sum of the solutions of the sub-problems. %
This makes such algorithms computationally very efficient, while at the same time resulting in a globally consistent and accurate solution.

In his work, \citet{Hirschmueller2005, Hirschmueller2008} proposed the \gls*{SGM} algorithm for the task of disparity estimation as part of the stereo normal case. %
In this, the stereo problem is formulated as a two-dimensional \gls*{MRF} with the energy function $E(\mathcal{U})$, which is to be minimized in order to find the disparity map $\mathcal{U}$:
\begin{equation}
\label{eq:sgm-energy}
\begin{split}
	E(\mathcal{U}) = \sum_{\mathrm{p}} \Big(\mathcal{S}(\mathrm{p}, u_{\mathrm{p}}) & + \sum_{\mathrm{q} \in \mathcal{W}_{\mathrm{p}}} \varphi_1 \cdot [|u_{\mathrm{p}} - u_{\mathrm{q}}| = 1] \\
	 & + \sum_{\mathrm{q} \in \mathcal{W}_{\mathrm{p}}} \varphi_2 \cdot [|u_{\mathrm{p}} - u_{\mathrm{q}}| > 1]\Big).
\end{split}
\end{equation}
here, $[\cdot]$ denotes the Iverson bracket, penalizing an absolute deviation of $1$ between disparity $u$ of pixel $\mathrm{p}$ and a neighboring pixel $\mathrm{q} \in \mathcal{W}_{\mathrm{p}}$ by the penalty $\varphi_1$, and a deviation greater than $1$ by $\varphi_2$, with $\varphi_1 < \varphi_2$. %

The minimization of \Cref{eq:sgm-energy}, however, is efficiently approximated by the utilization of dynamic programming. %
For each pixel, the matching costs inside the cost volume $\mathcal{S}$ are aggregated along numerous concentric one-dimensional paths with direction $\mathrm{r}$, which can be processed fully independently of each other. %
In this, the matching costs for a pixel $\mathrm{p}$ and a disparity $u \in \Gamma = [u_{\mathrm{min}}, u_{\mathrm{max}}]$ are recursively aggregated along the path according to: %
\begin{equation}
\label{eq:sgm-path}
\begin{aligned}
	L_\mathrm{r}(\mathrm{p}, u) = \mathcal{S}(\mathrm{p}, u) + \min\limits_{u'}\Big(L_\mathrm{r}(\mathrm{p-r}, u')+V(u,u')\Big).
\end{aligned}
\end{equation}
The unary data term of the energy functional, which holds the corresponding matching costs for pixel $\mathrm{p}$ and disparity $u$ inside the cost volume $\mathcal{S}$ is denoted as $\mathcal{S}(\mathrm{p}, u)$. %
The smoothness term, which penalizes deviations in the disparity $u$ of the currently considered pixel $\mathrm{p}$ and the disparity $u'$ of the previous pixel along the path is formulated by: %
\begin{equation}
\label{eq:sgm-smooth}
\begin{aligned}
V(u,u') &= 
	\begin{cases}
			0 &,\ \text{if}\ u = u' \\
    	\varphi_1 &,\ \text{if}\ \left|u-u'\right| = 1 \\
        \varphi_2 &,\ \text{if}\ \left|u-u'\right| > 1.
    \end{cases}
\end{aligned}
\end{equation}
After all paths for pixel $\mathrm{p}$ have been processed, the aggregated costs of the individual paths are summed up and stored inside the aggregated cost~volume~$\bar{\mathcal{S}}$: %
\begin{equation}
\label{eq:sgm-aggr}
\begin{aligned}
\bar{\mathcal{S}}(\mathrm{p}, u) &= \sum\limits_r{L_\mathrm{r}(\mathrm{p}, u)} .
\end{aligned}
\end{equation}
From this, the final disparity map is extracted by computing the pixel-wise \gls*{WTA} solution:
\begin{equation}
\label{eq:sgm-wta}
\begin{aligned}
\mathcal{U}(\mathrm{p}) = \argmin\limits_u{\bar{\mathcal{S}}(\mathrm{p}, u)} .
\end{aligned}
\end{equation}

\subsubsection{Extension of Semi-Global Matching to Plane-Sweep Matching and Different Surface Orientations}
\label{sec:methodology_sgm_x}

In the following, we describe in detail, how we have adopted the proposed \gls*{SGM} algorithm to the creation of depth hypotheses with the help of plane-sweep multi-image matching, as well as the extensions we have employed to account for non-fronto-parallel surfaces.
The proposed extensions \xSGM only affect the aggregation of the matching costs along the concentric paths. %
Thus, the extraction of the depth map $\mathcal{D}$ is done analogously to \Cref{eq:sgm-aggr} and \Cref{eq:sgm-wta} of the \gls*{SGM} algorithm, with the disparity being substituted by depth. %
If a fronto-parallel plane orientation is considered during the plane-sweep, the depth can be directly extracted from the plane parameterization. %
However, for non-fronto-parallel orientations, the depth map $\mathcal{D}$ is computed by a pixel-wise intersection of the viewing rays with the corresponding \gls*{WTA} solutions. %

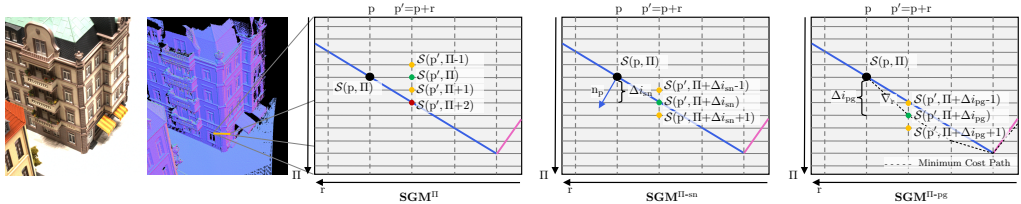
\begin{figure*}[t!]
	\centering%
	\resizebox{\textwidth}{!}{\subimport{figures/}{xsgm-edited.pdf_tex}}%
	\caption{ %
		Illustration of the three presented \xSGM path aggregations along one path direction $\mathrm{r}$. %
		\textbf{Column~1:} Reference image and normal map of a building. Illustrated area marked with yellow line. %
		\textbf{Column~2:} \piSGM path aggregation. %
		The blue and pink lines represent the blue and pink surface orientations on the building facade. %
		Aggregating the path costs for pixel $\mathrm{p}$ at plane $\Pi$, \piSGM will incorporate the previous costs at the same plane position (green) without additional penalty. %
		The previous path costs at $\Pi\,\pm 1$ (yellow) will be penalized with $\varphi_1$. %
		The previous path costs located at $\Pi\text{+}2$ (red), which is actually located on the corresponding surface, will be penalized with the highest penalty $\varphi_2$.
		\textbf{Column~3:} \snSGM uses the normal vector $\mathrm{n_p}$, encoding the surface orientation at pixel $\mathrm{p}$, and computes a discrete index jump $\Delta i_{\mathrm{sn}}$, which ideally adjusts the zero-cost transition, causing the previous path costs at $\Pi\text{+}2$ to not be penalized. %
		\textbf{Column~4:} Similar to \snSGM, \pgSGM adjusts the zero-cost transition. However, the discrete index jump $\Delta i_{\mathrm{pg}}$ is derived from the running gradient $\nabla \mathrm{r}$ of the minimum cost path. %
		}	
	\label{fig:xsgm} 
\end{figure*}

\paragraph{Resolving Plane Hypotheses with Semi-Global Matching}

Since the plane-sweep algorithm does not compute hypotheses on disparities, but rather pixel-wise plane distances relative to the reference camera and thus depth, the first \gls*{SGM} extension we propose is a straight-forward adaption of the standard \gls*{SGM} algorithm to a multi-view plane-sweep sampling. %
In this, the formulation of the \gls*{SGM} path aggregation is modified to 
\begin{equation}
\label{eq:xsgm}
\begin{aligned}
	L_\mathrm{r}(\mathrm{p}, \Pi) = \mathcal{S}(\mathrm{p}, \Pi) + \min\limits_{\delta'}\Big(L_\mathrm{r}(\mathrm{p-r}, \Pi')+V_{\Pi}(\Pi,\Pi')\Big),
\end{aligned}
\end{equation}
where $\Pi$ denotes the sampling plane at distance $\delta$. %
The smoothness term $V_{\Pi}$ now penalizes the selection of different planes between adjacent pixels along the path $L_\mathrm{r}$, instead of disparities. %
It is formulated as:  
\begin{equation}
\label{eq:sgm-smooth-fp}
\begin{aligned}
V_{\Pi}(\Pi,\Pi') &= 
	\begin{cases}
			0 &,\ \text{if}\ I(\Pi) = I(\Pi') \\
    	\varphi_1 &,\ \text{if}\ \left|I(\Pi) - I(\Pi')\right| = 1 \\
    \varphi_2 &,\ \text{if}\ \left|I(\Pi) - I(\Pi')\right| > 1,
    \end{cases}
\end{aligned}
\end{equation}
with $I(\cdot)$ being a function that returns the index of $\Pi$ within the set of sampling planes (\cf \Cref{fig:xsgm}, Column 2).
We denote this extension as \emph{plane-wise} \gls*{SGM} (\piSGM). %
In our previous publication, \citep{Ruf2019efficient} we have referred to this extension as \emph{fronto-parallel} \gls*{SGM} (\fpSGM), since we have only considered a fronto-parallel sweeping direction so far. %
However, the extension is not restricted to a fronto-parallel plane orientation in the plane-sweep sampling and is also evaluated with slanted planes in the scope of this work. %
Given a pixel-wise \gls*{WTA} plane parameterization, the corresponding depth is extracted by intersecting the viewing ray through pixel $\mathrm{p} = (p_x\ p_y)^\intercal$ with the corresponding plane: %
\begin{equation}
\begin{aligned}
	d_{\mathrm{p}} = \frac{- \delta}{\mathrm{n}^\intercal \cdot \mathbf{K}^{\text{-}1} \cdot (p_x\ p_y\ 1)^\intercal} .
\end{aligned}
\end{equation}

\paragraph{Incorporating Surface Normals to Adjust the Zero-Cost Transition}

The smoothness term of the initial \gls*{SGM} algorithm is formulated with discrete disparity differences (\cf \Cref{eq:sgm-smooth}), penalizing discrete disparity jumps between neighboring pixels. %
In its optimization scheme, it does not consider any subpixel disparity and thus favors fronto-parallel surface structures, leading to stair-casing artifacts if no post-processing is employed \citep{Scharstein2018surface}. %
The same holds for our first extension, \piSGM. %
Even though the plane-sweep sampling also supports non-fronto-parallel plane orientations, the smoothness term of \piSGM (\cf \Cref{eq:sgm-smooth-fp}) does not, and strongly penalizes index jumps in the sampling planes of more than 1. %
While this is desired if the plane-orientation coincides with the surface orientation, it will still lead to stair-casing artifacts if the surface and plane orientations do not align. %
In order to overcome the favoring of fronto-parallel structures and adjust the smoothness term of \gls*{SGM} to surfaces that are slanted with respect to the sampling direction, \citet{Scharstein2018surface} suggest to add an offset to the smoothness term.
This offset can be extracted from additional information on the surface orientation, \eg surface normals, that will make the zero-cost transition coincide with the surface orientation. %
We have adopted this approach as part of our second extension and thus call it \emph{surface normal} \gls*{SGM} (\snSGM). %

In our hierarchical approach, we extract the normal vectors from the normal map $\mathcal{N}^{l+1}$, which was estimated in the previous level of the pyramid (\cf \Cref{fig:methodology_overview}). %
The pixel-wise normal vectors $\mathrm{n}_\mathrm{p} = \mathcal{N}^{l+1}(\mathrm{p})$ indicate the surface orientation at the scene point $\mathrm{P}$, which is computed by intersecting the viewing ray through $\mathrm{p}$ with the plane $\Pi$. %
From this, the discrete index jump ${\Delta i}_{\mathrm{sn}}$ through the set of sampling planes can be calculated which is caused by the tangent plane to $\mathrm{n}_\mathrm{p}$. %
Since the plane-sweep sampling is not restricted to fronto-parallel plane orientations, the index jump ${\Delta i}_{\mathrm{sn}}$ needs to be computed based on the difference between the tangent plane at $\mathrm{P}_{\Pi}$ and the orientation of the sampling planes in the direction $\mathrm{r}$ of the currently considered aggregation path.
With ${\Delta i}_{\mathrm{sn}}$, the smoothness term used by our extension \snSGM is adjusted according to
\begin{equation}
\label{eq:sgm-smooth-sn}
\begin{aligned}
V_{\Pi\text{-}\mathrm{sn}}(\Pi,\Pi') = V_{\Pi}(\Pi + {\Delta i}_{\mathrm{sn}},\Pi') .
\end{aligned}
\end{equation}
This allows to align the zero-cost transition of the \gls*{SGM} path aggregation to the surface orientation of the scene (\cf \Cref{fig:xsgm}, Column 3). %
The pixel-wise discrete index jumps can be computed once for each pixel $\mathrm{p}$ and each path direction $\mathrm{r}$, as also noted by \citet{Scharstein2018surface}, providing little computational overhead. %

\paragraph{Penalizing Deviations from the Gradient of the Minimum Cost Path}

Instead of relying on additional information, \eg normal vectors, the third of our proposed extension computes the running gradient $\nabla \mathrm{r}$ in scene space, which corresponds to the minimal path costs, in order to adjust the zero-cost transition in the aggregation of the path costs for non-fronto-parallel surface orientations. %
Hence, it is denoted as \emph{path gradient} \gls*{SGM} (\pgSGM). %

The gradient vector $\nabla \mathrm{r} = \mathrm{P} - \mathrm{P'}$ in scene space is dynamically computed while traversing along the path $\mathrm{r}$. %
In this, $\mathrm{P}$ again denotes the scene point that is found by intersecting the viewing ray through $\mathrm{p}$ with $\Pi$, while $\mathrm{P'}$ denotes the scene point, which is parameterized by $\mathrm{p'}$ and the plane $\hat{\Pi}'$. %
Here, $\mathrm{p'} = \mathrm{p+r}$ represents the predecessor of $\mathrm{p}$ along the path $\mathrm{r}$ and $\hat{\Pi}'$ denotes the plane at distance $\hat{\delta} = \argmin_{\delta}{L_\mathrm{r}(\mathrm{p'}, \Pi)}$ associated with the previous minimal path costs. %

From this, a discrete index jump ${\Delta i}_{\mathrm{pg}}$ is computed, which is again used to account for possibly slanted surfaces in scene space by adjusting the zero-cost transition of the smoothness term according to %
\begin{equation}
\label{eq:sgm-smooth-pg}
\begin{aligned}
V_{\Pi\text{-}\mathrm{pg}}(\Pi,\Pi') = V_{\Pi}(\Pi + {\Delta i}_{\mathrm{pg}},\Pi') .
\end{aligned}
\end{equation}
This implicitly penalizes deviations from the running gradient between two scene points corresponding to two consecutive pixels on the aggregation path $\mathrm{r}$ (\cf \Cref{fig:xsgm}, Column 4). %

\subsubsection{Adaptive Smoothness Penalties Within Semi-Global Matching}
\label{sec:adaptive_p2}

In the original publication of the \gls*{SGM} algorithm, \citet{Hirschmueller2005, Hirschmueller2008} suggests to use an adaptive adjustment of the second penalty $\varphi_2$ according to the image gradient along path $\mathrm{r}$. %
This should enforce a preservation of depth discontinuities at object boundaries. %
In this work, the adaptive adjustment of $\varphi_2$ is based on the absolute intensity difference ($\Delta \mathcal{I}_{\mathrm{pq}}$) between two neighboring pixels $\mathrm{p}$ and $\mathrm{q}$. %
It is formulated by 
\begin{equation}
\label{eq:p2-grad}
\begin{aligned}
\varphi_2 = \varphi_1 \cdot \left(1+\alpha \cdot \exp\left(-\frac{\Delta \mathcal{I}_{\mathrm{pq}}}{\beta}\right)\right)
\end{aligned}
\end{equation}
with $\Delta \mathcal{I}_{\mathrm{pq}} = |\mathcal{I}(\mathrm{p}) - \mathcal{I}(\mathrm{q})|$. %
Moreover, $\alpha=8$ and $\beta=10$ are set according to \citet{Scharstein2018surface}. %
Since the presented approach uses multiple matching images, the \gls*{SGM} penalties are multiplied with the number of input images inside the left and right subsets with respect to $\mathcal{I}_{\mathrm{ref}}$, since the matching costs are summed up within these image sets. %

%Proposed in \citep{Ruf2018deep}, the second strategy uses a binary line image $\mathcal{I}_{\mathrm{ref}}^{\mathrm{line}}$, corresponding to the reference image, which can be computed from a line segment detector such as \citet{Gioi2010lsd}. %
%Given $\mathcal{I}_{\mathrm{ref}}^{\mathrm{line}}$, we set $\varphi_2^{\mathrm{line}}$ to $\varphi_1$ at each pixel marking a detected line segment: %
%\begin{equation}
%\label{eq:p2-line}
%\begin{aligned}
%\varphi_2^{\mathrm{line}} &= 
%	\begin{cases}
%    	\varphi_1 &,\ \text{if}\ I^{line}_\mathrm{ref}(\mathrm{p}) = 1 \\
%    \varphi_2 &,\ \text{otherwise}.
%    \end{cases}
%\end{aligned}
%\end{equation}
%%
%As noted in \citep{Ruf2018deep}, we argue that this will enforce strong discontinuities at object boundaries, since at pixels with corresponding line segments, distant plane hypotheses are penalized the same way as neighboring hypotheses. %

\subsubsection{Depth Refinement}

In a final optimization step in the process of depth map estimation, the pixel-wise estimates are refined to lie in between the actual sampling steps of the plane-sweep sampling. %
Here, the simple and yet effective approach to fit a parabola through the \gls*{WTA} solution and its two neighbors is employed. %
This is done analogously to implementing a disparity refinement based on curve fitting for the stereo normal case \citep{Scharstein2002}.
But since the sampling points are not equidistantly spaced between each other, the depth differences between the \gls*{WTA} solution and its two neighbors needs to be considered in the optimization and the finding of the curves' minimum. %

\subsection{Extraction of Surface Normals from Depth Maps}
\label{sec:methodology_normal}

From the estimated depth map $\mathcal{D}$ our approach computes a normal map $\mathcal{N}$, holding the local surface orientations in the form of three-dimensional normal vectors. %
The surface normal vectors $\mathrm{n_p}$ are calculated for each pixel $\mathrm{p}$ by reprojecting $\mathcal{D}$ into a three-dimensional point cloud, based on the intrinsic camera parameters and the corresponding depth estimate. %
Then, the cross-product $\mathrm{n_p} = \mathrm{h_p}\times \mathrm{v_p}$ is computed, with $\mathrm{h_p}$ being the difference vector between the scene points of two neighboring pixels to $\mathrm{p}$ in horizontal direction, and $\mathrm{v_p}$ being the difference vector in vertical direction. %

Solely using the cross-product to compute the surface orientation does not incorporate any local smoothness assumption, which is why we apply an a-posteriori smoothing to the normal map. %
In this, an appearance-based weighted Gaussian smoothing in a local two-dimensional window $\mathcal{W}_\mathrm{p}$ around $\mathrm{p}$ is employed, which adjusts the smoothing strength depending on the intensity difference between $\mathrm{q} \in \mathcal{W}_\mathrm{p}$ and $\mathrm{p}$:
\begin{equation}
\label{eq:normal}
\begin{aligned}
\mathcal{N}(\mathrm{p}) = \frac{\bar{\mathrm{n}}_\mathrm{p}}{\left|\bar{\mathrm{n}}_\mathrm{p}\right|},
\end{aligned}
\end{equation}
with
\begin{equation}
\label{eq:normal-smooth}
\bar{\mathrm{n}}_\mathrm{p} = \mathrm{n_p} + \sum\limits_{\mathrm{q}\in\mathcal{W}_\mathrm{p}}\mathrm{n_q} \cdot \frac{1}{\sqrt{2\pi\sigma^2}} \cdot \exp\left(-\frac{\left(\mathrm{q}-\mathrm{p}\right)^2}{2\sigma^2}-\frac{\Delta \mathcal{I}_{\mathrm{pq}}}{\beta}\right).
\end{equation}
Similar as in \Cref{eq:p2-grad}, $\beta$ is set to $10$, while $\sigma$ is fixed to the radius of~$\mathcal{W}_\mathrm{p}$.

\subsection{Estimation of Confidence Measures based on Surface Orientation}
\label{sec:methodology_confidence}

Apart from the depth map $\mathcal{D}$ and the normal map $\mathcal{N}$, the presented approach also computes confidence measures with respect to the depth estimates in the range of $\left[0,\ 1\right]$ and stores them inside a confidence map $\mathcal{C}$. %
Such confidence measures allow a subsequent reasoning on the certainty of the corresponding estimates and, in turn, improve further processing. %
Thus, confidence maps are helpful byproducts for subsequent steps, such as depth map fusion or scene interpretation. %
Furthermore, they allow to gain more insight on the effects of different configurations of the presented approach, based on \gls{ROC} curve analysis (\cf \Cref{fig:sgmx_roc}). %
In our previous work \citep{Ruf2019efficient}, the computation of the pixel-wise confidence values was based on the data stored inside the aggregated cost volume $\bar{\mathcal{S}}$ computed by the \gls*{SGM} optimization. %
In particular, a confidence value based on the results of the path aggregation, adopted from the observation of \citet{Drory2014semi}, as well as based on the uniqueness of the \gls*{WTA} solution was used. %
Further, experiments and evaluations relying on the analysis of a \gls*{ROC} curve have shown that these confidence measures do not provide suitable prediction on the certainty of the depth estimates. %
Thus, in this paper, the computation of the pixel-wise confidence measures relies on the geometric characteristic of the estimated depth map and is deduced from the normal vectors stored inside the normal map $\mathcal{N}$ and the plane orientations of the plane-sweep sampling.

In particular, the geometric confidence measure is based on the enclosed angles between the local surface orientation stored inside the normal map $\mathrm{n_p} = \mathcal{N}(\mathrm{p})$, the orientation of the sampling plane $\mathrm{n}_{\Pi}$ and the reverted viewing direction $\mathrm{v}$. %
This is adopted from the geometric weighting factor proposed by \citet{Kolev2014}. %
They argue that a depth estimate is more accurate if the surface orientation of the observed geometry is fronto-parallel to the image plane of the camera, and less accurate if the camera is observing slanted surfaces. %
This correlation is modeled by the scalar product between the surface orientation and the reverted viewing direction. %
Furthermore, since the image warping, as part of the image matching, can be aligned to the surface orientation by adjusting the normal vector of the plane-sweep algorithm, the plane orientation is also considered. %
Thus, the geometry-based weighting factor is computed according to: 
\begin{equation}
\label{eq:weight-normal}
\begin{aligned}
\mathcal{C}(\mathrm{p}) &= \begin{cases}
					\frac{\langle\mathrm{n}_{\mathrm{p}}, \mathrm{n}_\Pi\rangle\langle\mathrm{n}_{\Pi}, \mathrm{v}\rangle\ -\ \cos \rho}{1\ -\ \cos \rho} &,\ \text{if}\ \lbrace \sphericalangle (\mathrm{n}_{\mathrm{p}}, \mathrm{n}_\Pi) \land \sphericalangle (\mathrm{n}_{\Pi}, \mathrm{v}) \rbrace \leq \rho \\
					0 &,\ \text{otherwise}\ .
				   \end{cases}
\end{aligned}
\end{equation}%
All of the above vectors are assumed to be normalized and given with respect to the local coordinate system of the camera, thus $\mathrm{v} = (0\ 0\ \text{-}1)^\intercal$. %
Just as in the work of \citet{Kolev2014}, a critical angle $\rho = 60^{\circ}$ is used to mark the measurements, for which the enclosed angles exceed this threshold, as unreliable. %
The additional consideration of $\mathrm{n}_{\Pi}$ in \Cref{eq:weight-normal} implicitly models the indirect matching of the input images via the plane-induced homography. %

\subsection{Post-Processing and Depth Map Filtering}
\label{sec:methodology_post}
\glsreset{DoG}

In a final post-processing step, remaining outliers are removed from the depth, normal and confidence maps by applying a \gls*{DoG} filtering (\Cref{sec:methodology_post_dog}) and a geometric consistency check (\Cref{sec:methodology_post_geoConsistency}). %

\subsubsection{Difference-of-Gaussian Filtering}
\label{sec:methodology_post_dog}

\begin{algorithm}[!b]
 \KwData{unfiltered depth, normal and confidence maps ($\mathcal{D}$, $\mathcal{N}$ and $\mathcal{C}$) as well as corresponding reference image $\mathcal{I}_{\mathrm{ref}}$.} %
 \KwResult{filtered $\mathcal{D}$, $\mathcal{N}$ and $\mathcal{C}$, in which all estimates corresponding to weakly textured areas in $\mathcal{I}_{\mathrm{ref}}$ are removed.}
 \BlankLine
 use a Gaussian filter with a kernel of $7\times 7$\px to smooth the reference frame $\mathcal{I}{_\mathrm{ref}}$, yielding $\mathcal{I}^{\mathrm{smooth}}_{\mathrm{ref}}$. %
 
 compute the \gls*{DoG} image depicting local image gradients, according to: $\mathcal{I}^{\mathrm{DoG}}_{\mathrm{ref}} = \mathcal{I}_{\mathrm{ref}} - \mathcal{I}^{\mathrm{smooth}}_{\mathrm{ref}}$. %
 
 apply a binary threshold to compute the \gls*{DoG} mask $\mathcal{M}^{\mathrm{DoG}}$, marking all image areas in which the intensity change is greater than $0.5$. %
 
 remove activation areas smaller than $7$\px in $\mathcal{M}^{\mathrm{DoG}}$ by applying a speckle filter. %
 
 dilate $\mathcal{M}^{\mathrm{DoG}}$ with a kernel size of $3\times 3$\px to fill small holes in activation areas. %
 
 remove deactivation areas smaller than $21$\px by applying a speckle filter to the inverted \gls*{DoG} mask $\mathcal{M}^{\mathrm{DoG}\text{-}\mathrm{inv}} = 1 - \mathcal{M}^{\mathrm{DoG}}$. %
 
 invalidate pixels in $\mathcal{D}$, $\mathcal{N}$ and $\mathcal{C}$ for which $\mathcal{M}^{\mathrm{DoG}}=1$. %
 
 \caption{Overview on the \acrlong*{DoG} filter to invalidate all image areas belonging to weakly textured areas.}
 \label{alg:dog-filter}
\end{algorithm}

As proposed by \citet{Wenzel2016dense}, the \gls*{DoG} filter allows to remove estimates from $\mathcal{D}$, $\mathcal{N}$ and $\mathcal{C}$ by masking out pixels in image regions that only provide little textural information (\eg unsharp or overexposed areas).
Here, it is assumed that in such regions the image matching is ambiguous and that it leads to less accurate results. %
The \gls*{DoG} filter is used to detect  weakly textured areas inside the reference image $\mathcal{I}_{\mathrm{ref}}$ and build up a binary image mask, which, in turn, is used to remove the estimates from the corresponding maps. %
\Cref{alg:dog-filter} provides an overview on the implementation of the employed \gls*{DoG} filter, which is similar to the one proposed in \citep{Wenzel2016dense}. %

\subsubsection{Geometric Consistency based on Mutual Reprojection Error}
\label{sec:methodology_post_geoConsistency}

In the stereo normal case, in which the depth is constructed from only two views, a simple and yet very effective approach to enforce geometric consistency and perform occlusion detection is the so-called left-right consistency check. %
In this, the pixel-wise disparities stored inside the disparity image of the reference image are compared to the approximated disparities inside the disparity map of the matching image. %
If they differ by a certain threshold, the corresponding estimates inside the disparity map of the reference image are invalidated. %
This can also be formulated as the mutual reprojection error, in which each pixel $\mathrm{p}^{\mathrm{ref}}$ of $\mathcal{D}_{\mathrm{ref}}$, having a depth estimate $d^{\,\mathrm{ref}}_{\mathrm{p}}$, is projected into the view of a second depth map $\mathcal{D}^k$, according to $d^{\,\mathrm{ref}}_{\mathrm{p}}$ and the corresponding projection matrices $\mathbf{P}_{\mathrm{ref}}$ and $\mathbf{P}_k$, yielding the image point $\mathrm{p}^{k}$. %
Given $\mathrm{p}^k$ and the corresponding depth $d^{\,k}_{\mathrm{p}}$ from $\mathcal{D}_k$, the image point $\mathrm{p}^k$ is projected back into the view of $\mathcal{D}_{\mathrm{ref}}$, resulting in $\tilde{\mathrm{p}}^{\mathrm{ref}}$.
Finally, if the Euclidean distance between $\mathrm{p}^{\mathrm{ref}}$ and $\tilde{\mathrm{p}}^{\mathrm{ref}}$ exceeds a given threshold $\eta_{\mathrm{r}}$ the estimate at $\mathrm{p}^{\mathrm{ref}}$ is invalidated. %
\citet{Schoenberger2016mvs} formulate this reprojection error for pixel $\mathrm{p}$ in a reference view and a neighboring map $k$ as $\epsilon^k_{\mathrm{r}}(\mathrm{p}) = \Vert \mathrm{p} - \mathbf{H}^k_{\mathrm{p}} \cdot \mathbf{H}_{\mathrm{p}} \cdot \mathrm{p} \Vert$, with $\mathbf{H}_{\mathrm{p}}$ being the forward projection into the view $k$ according to $d^{\,\mathrm{ref}}_{\mathrm{p}}$, and $\mathbf{H}^k_{\mathrm{p}}$ being the corresponding backward projection according to $d^{\,k}_{\mathrm{p}}$. %

We have adopted this approach to perform a final geometry-based filtering between a number of depth maps within a sliding window. %
This is not part of the actual hierarchical processing pipeline, but rather a separate post-processing step, since it requires the results of other image bundles of the input sequence. %
If possible, the center-most depth map of the sliding window $\Psi$ is chosen as a reference view for which the filtering is performed. %
At the beginning or the end of the sequence, where the sliding window would exceed the boundaries, the window is shifted to either side of the reference view so that it is always inside the boundaries of the sequence and that no depth map is filtered multiple times. %
Apart from thresholding the reprojection error $\epsilon^k_{\mathrm{r}}$, another criterion is introduced to evaluate the geometric consistency, namely the number of neighboring views for which the reprojection error is within the threshold, \ie the number of hits: $ \epsilon_{\textrm{h}}(\mathrm{p}) = \sum_{k} [\epsilon^k_{\mathrm{r}}(\mathrm{p}) < \eta_{\mathrm{r}}]$, with $[\cdot]$ being the Iverson bracket.
\Cref{alg:geometric-filter} gives an overview on geometric consistency and the according filtering of the depth, normal and confidence maps. %
In the scope of this work, the size of the sliding window is set to $|\Psi| = 5$, the threshold for the reprojection error to $\eta_{\mathrm{r}} = 10$ and the consistency threshold to $\eta_{\mathrm{h}} = 3$. %

\begin{algorithm}[h!]
 \KwData{depth, normal and confidence maps ($\mathcal{D}_k$, $\mathcal{N}_k$ and $\mathcal{C}_k$) within a sliding window $\Psi$ of the input sequence as well as corresponding projection matrices $\mathbf{P}_k$.} %
 \KwResult{filtered $\mathcal{D}_{\mathrm{ref}}$, $\mathcal{N}_{\mathrm{ref}}$ and $\mathcal{C}_{\mathrm{ref}}$ of reference view, in which all estimates that are not geometrically consistent are removed.}% 
 \BlankLine
 select $\mathcal{D}_{\mathrm{ref}}$, $\mathcal{N}_{\mathrm{ref}}$ and $\mathcal{C}_{\mathrm{ref}}$ corresponding to the center-most view within the sliding window $\Psi$.
 
 \ForEach{pixel $\mathrm{p}^{\mathrm{ref}} \in \mathcal{D}_{\mathrm{ref}}$ {\normalfont \textbf{and}} neighboring view $k \in \Psi$}{ % 
 
 calculate number of hits for which reprojection error is below threshold:
 $$
 \epsilon_{\textrm{h}}(\mathrm{p}) = \sum\limits_{k} [\epsilon^k_{\mathrm{r}}(\mathrm{p}) < \eta_{\mathrm{r}}],\ \text{with}\ \ \epsilon^k_{\mathrm{r}}(\mathrm{p}) = \Vert \mathrm{p} - \mathbf{H}^k_{\mathrm{p}} \cdot \mathbf{H}_{\mathrm{p}} \cdot \mathrm{p} \Vert .
 $$
 
 if $\epsilon_{\textrm{h}} < \eta_{\mathrm{h}}$, invalidate pixel $\mathrm{p}$ in $\mathcal{D}_{\mathrm{ref}}$, $\mathcal{N}_{\mathrm{ref}}$ and $\mathcal{C}_{\mathrm{ref}}$, by setting it to $0$.
  
 }
 \caption{Overview on the geometric consistency filter.}
 \label{alg:geometric-filter}
\end{algorithm}

%% file: figures/methodology-edited.pdf_tex
%% Accompanies image file 'methodology.pdf' (pdf, eps, ps)
\begingroup%
  \makeatletter%
  \providecommand\color[2][]{%
    \errmessage{(Inkscape) Color is used for the text in Inkscape, but the package 'color.sty' is not loaded}%
    \renewcommand\color[2][]{}%
  }%
  \providecommand\transparent[1]{%
    \errmessage{(Inkscape) Transparency is used (non-zero) for the text in Inkscape, but the package 'transparent.sty' is not loaded}%
    \renewcommand\transparent[1]{}%
  }%
  \providecommand\rotatebox[2]{#2}%
  \newcommand*\fsize{\dimexpr\f@size pt\relax}%
  \newcommand*\lineheight[1]{\fontsize{\fsize}{#1\fsize}\selectfont}%
  \ifx\svgwidth\undefined%
    \setlength{\unitlength}{725.16443928bp}%
    \ifx\svgscale\undefined%
      \relax%
    \else%
      \setlength{\unitlength}{\unitlength * \real{\svgscale}}%
    \fi%
  \else%
    \setlength{\unitlength}{\svgwidth}%
  \fi%
  \global\let\svgwidth\undefined%
  \global\let\svgscale\undefined%
  \makeatother%
  \begin{picture}(1,0.73325727)%
    \lineheight{1}%
    \setlength\tabcolsep{0pt}%
    \begin{large}
    \put(0,0){\includegraphics[width=\unitlength,page=1]{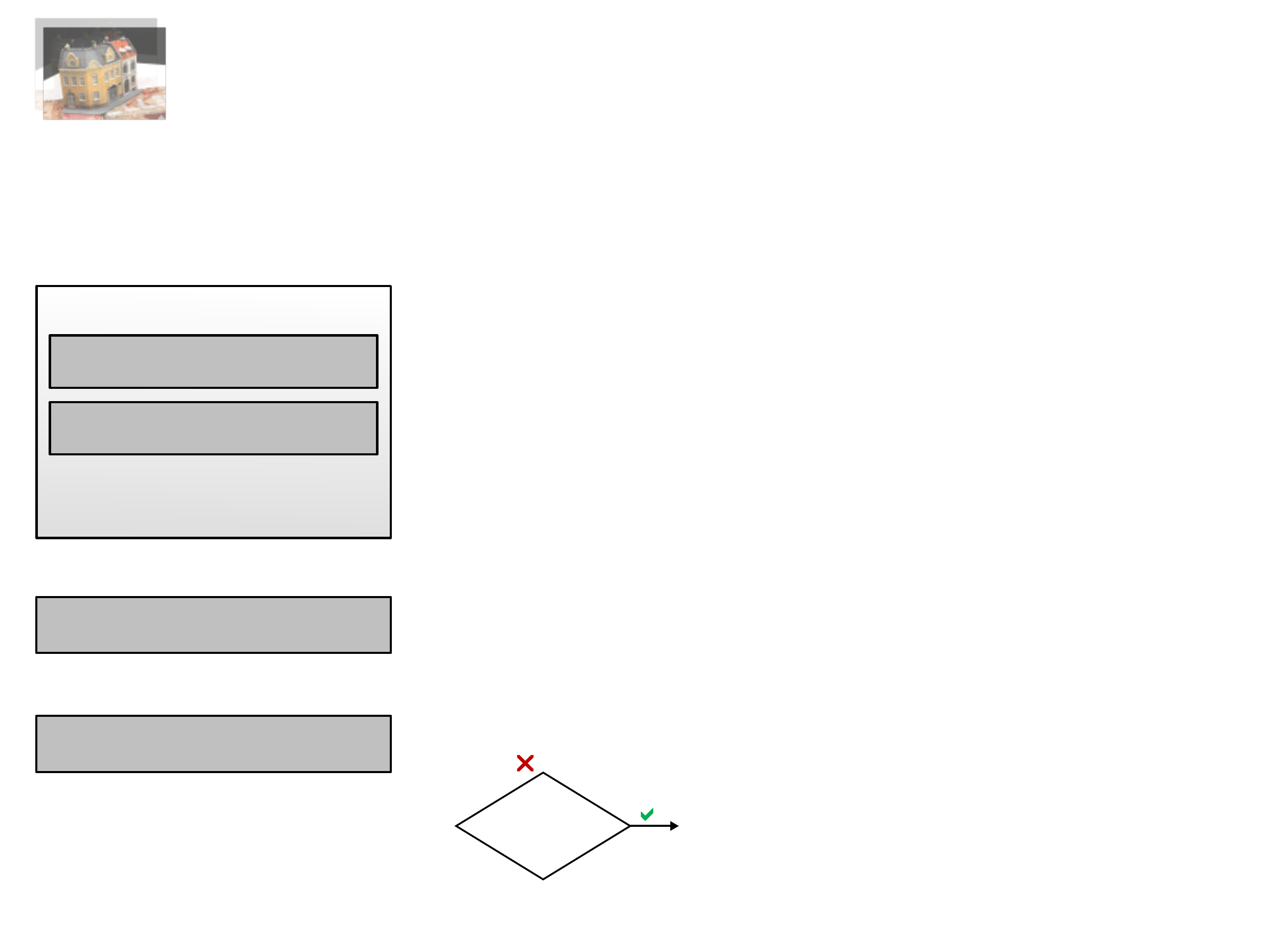}}%
    \put(0.74131127,-0.85438845){\color[rgb]{0,0,0}\makebox(0,0)[lt]{\begin{minipage}{0.27581356\unitlength}\raggedright  \end{minipage}}}%
    \put(0.175,0.64){\color[rgb]{0,0,0}\makebox(0,0)[lt]{\lineheight{0}\smash{\begin{tabular}[t]{l}$\left(\mathcal{I}, \mathbf{P}\right)_k \in \Omega$\end{tabular}}}}%
    \put(0,0){\includegraphics[width=\unitlength,page=2]{methodology.pdf}}%
    \put(0.03461611,0.59493526){\color[rgb]{0,0,0}\makebox(0,0)[lt]{\lineheight{0}\smash{\begin{tabular}[t]{l}Gaussian Pyramid Generation\end{tabular}}}}%
    \put(0.03410707,0.48726519){\color[rgb]{0,0,0}\makebox(0,0)[lt]{\lineheight{0}\smash{\begin{tabular}[t]{l}Depth Map Estimation\end{tabular}}}}%
    \put(0.04353796,0.44797571){\color[rgb]{0,0,0}\makebox(0,0)[lt]{\lineheight{0}\smash{\begin{tabular}[t]{l}\normalsize Plane-Sweep Image Matching\end{tabular}}}}%
    \put(0.04468533,0.39626108){\color[rgb]{0,0,0}\makebox(0,0)[lt]{\lineheight{0}\smash{\begin{tabular}[t]{l}\normalsize \xSGM Optimization\end{tabular}}}}%
    \put(0.03392123,0.24300235){\color[rgb]{0,0,0}\makebox(0,0)[lt]{\lineheight{0}\smash{\begin{tabular}[t]{l}Normal Map Estimation\end{tabular}}}}%
    \put(0.032,0.55586504){\color[rgb]{0,0,0}\makebox(0,0)[lt]{\lineheight{0}\smash{\begin{tabular}[t]{l}$l = (n-1),\ \Gamma^{\,0}$\end{tabular}}}}%
    \put(0.17477562,0.29474198){\color[rgb]{0,0,0}\makebox(0,0)[lt]{\lineheight{0}\smash{\begin{tabular}[t]{l}$\left(\mathcal{D}^{\,l}\right)$\end{tabular}}}}%
    \put(0.17477562,0.11072186){\color[rgb]{0,0,0}\makebox(0,0)[lt]{\lineheight{0}\smash{\begin{tabular}[t]{l}$\left(\mathcal{D}^{\,l}, \mathcal{N}^{\,l}, \mathcal{C}^{\,l}\right)$\end{tabular}}}}%
    \put(0.804,0.06748047){\color[rgb]{0,0,0}\makebox(0,0)[lt]{\lineheight{0}\smash{\begin{tabular}[t]{l}$\left(\mathcal{D}, \mathcal{N}, \mathcal{C}\right)$\end{tabular}}}}%
    \put(0.395,0.087){\color[rgb]{0,0,0}\makebox(0,0)[lt]{\lineheight{0}\smash{\begin{tabular}[t]{l}$l == 0$\end{tabular}}}}%
    \put(0.49697613,0.35){\color[rgb]{0,0,0}\makebox(0,0)[lt]{\lineheight{0}\smash{\begin{tabular}[t]{l}$\left(\bar{\mathcal{D}}^{\,l}, \bar{\mathcal{N}}^{\,l}\right),\ \Gamma^{\,l-1}_\mathrm{p}$\end{tabular}}}}%
    \put(0,0){\includegraphics[width=\unitlength,page=3]{methodology.pdf}}%
    \put(0.04353796,0.34396009){\color[rgb]{0,0,0}\makebox(0,0)[lt]{\lineheight{0}\smash{\begin{tabular}[t]{l}\normalsize Depth Refinement\end{tabular}}}}%
    \put(0.03392123,0.15121665){\color[rgb]{0,0,0}\makebox(0,0)[lt]{\lineheight{0}\smash{\begin{tabular}[t]{l}Confidence Map Estimation\end{tabular}}}}%
    \put(0.17477562,0.20295629){\color[rgb]{0,0,0}\makebox(0,0)[lt]{\lineheight{0}\smash{\begin{tabular}[t]{l}$\left(\mathcal{D}^{\,l}, \mathcal{N}^{\,l}\right)$\end{tabular}}}}%
    \put(0,0){\includegraphics[width=\unitlength,page=4]{methodology.pdf}}%
    \put(0.3673762,0.28900467){\color[rgb]{0,0,0}\makebox(0,0)[lt]{\lineheight{0}\smash{\begin{tabular}[t]{l}\normalsize Up-Scaling\end{tabular}}}}%
    \put(0.36719036,0.2328455){\color[rgb]{0,0,0}\makebox(0,0)[lt]{\lineheight{0}\smash{\begin{tabular}[t]{l}\normalsize Recovering Local Depth Range\end{tabular}}}}%
    \put(0,0){\includegraphics[width=\unitlength,page=5]{methodology.pdf}}%
    \put(0.5522029,0.14046104){\color[rgb]{0,0,0}\makebox(0,0)[lt]{\lineheight{0}\smash{\begin{tabular}[t]{l}Post Processing\end{tabular}}}}%
    \put(0.5456338,0.10117156){\color[rgb]{0,0,0}\makebox(0,0)[lt]{\lineheight{0}\smash{\begin{tabular}[t]{l}\normalsize DoG Filtering\end{tabular}}}}%
    \put(0.5456338,0.04945688){\color[rgb]{0,0,0}\makebox(0,0)[lt]{\lineheight{0}\smash{\begin{tabular}[t]{l}\normalsize Geometric Consistency Check\end{tabular}}}}%
    \put(0,0){\includegraphics[width=\unitlength,page=6]{methodology.pdf}}%
    \end{large}
  \end{picture}%
\endgroup%

%% file: figures/Plane-Sweep-edited.pdf_tex
%% Creator: Inkscape 0.91_64bit, www.inkscape.org
%% Accompanies image file 'Plane-Sweep.pdf' (pdf, eps, ps)
\begingroup%
  \makeatletter%
  \providecommand\color[2][]{%
    \errmessage{(Inkscape) Color is used for the text in Inkscape, but the package 'color.sty' is not loaded}%
    \renewcommand\color[2][]{}%
  }%
  \providecommand\rotatebox[2]{#2}%
  \ifx\svgwidth\undefined%
    \setlength{\unitlength}{1621.3770108bp}%
    \ifx\svgscale\undefined%
      \relax%
    \else%
      \setlength{\unitlength}{\unitlength * \real{\svgscale}}%
    \fi%
  \else%
    \setlength{\unitlength}{\svgwidth}%
  \fi%
  \global\let\svgwidth\undefined%
  \global\let\svgscale\undefined%
  \makeatother%
  \begin{picture}(1,0.41219333)%
    \begin{footnotesize}
    \put(0.60111017,0.40135908){\color[rgb]{0,0,0}\makebox(0,0)[lb]{\smash{$\Pi_{\mathrm{max}}$}}}%
    \put(0.65959806,0.34546311){\color[rgb]{0,0,0}\makebox(0,0)[lb]{\smash{$\Pi_{\mathrm{min}}$}}}%
    \put(0.61815355,0.37509576){\color[rgb]{0,0,0}\makebox(0,0)[lb]{\smash{$\Pi$}}}%
    \put(0,0){\includegraphics[width=\unitlength,page=1]{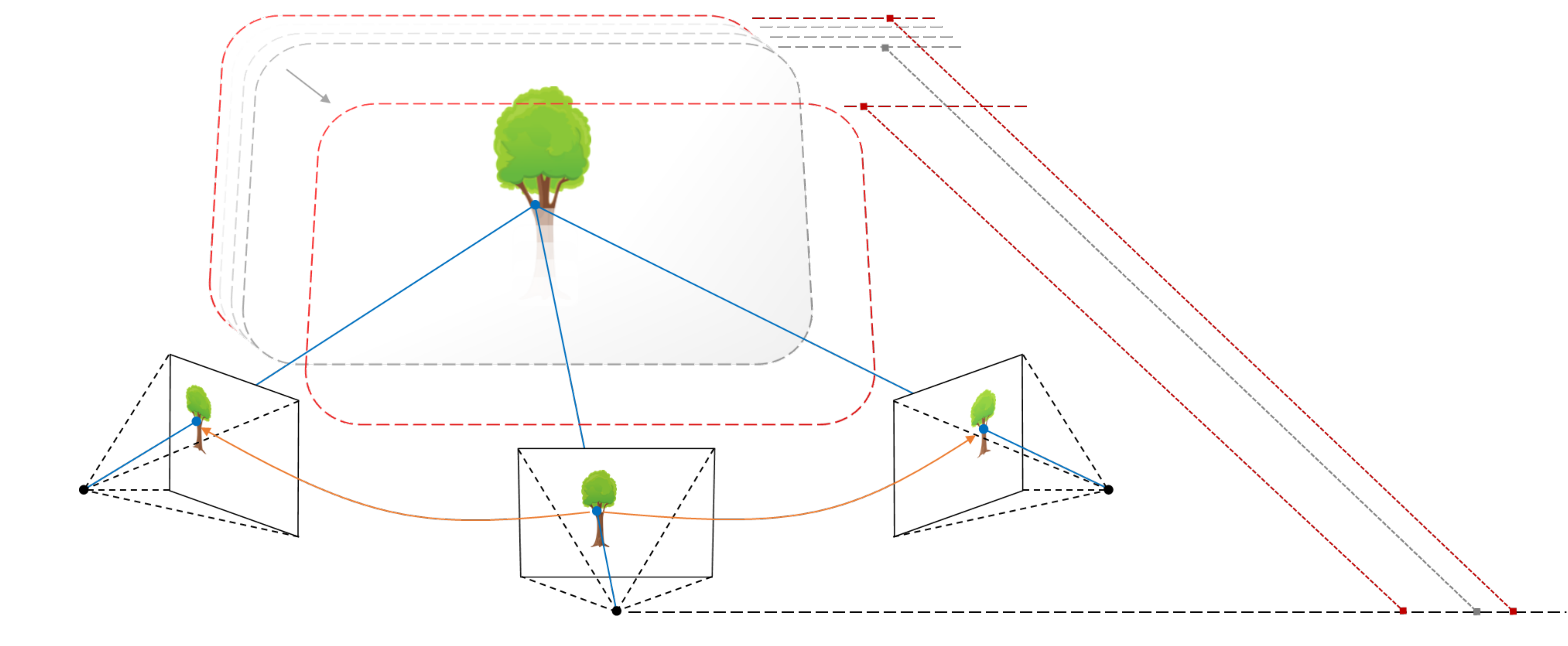}}%
    \put(0.65885051,0.2288335){\color[rgb]{0,0,0}\makebox(0,0)[rb]{\smash{$\delta_{\mathrm{min}}$}}}%
    \put(0.67470487,0.25653887){\color[rgb]{0,0,0}\makebox(0,0)[rb]{\smash{$\delta$}}}%
    \put(0.70452569,0.27668156){\color[rgb]{0,0,0}\makebox(0,0)[lb]{\smash{$\delta_{\mathrm{max}}$}}}%
    \put(0.20372989,0.35951905){\color[rgb]{0,0,0}\makebox(0,0)[lb]{\smash{$\mathrm{n}$}}}%
    \put(0.33309553,0.28008578){\color[rgb]{0,0.43921569,0.75294118}\makebox(0,0)[rb]{\smash{$\mathrm{P}_{\Pi}$}}}%
    \put(0.05511085,0.08276546){\color[rgb]{0,0,0}\makebox(0,0)[rb]{\smash{$\mathrm{C}_{\text{-}1}$}}}%
    \put(0.39600243,0.00006831){\color[rgb]{0,0,0}\makebox(0,0)[lb]{\smash{$\mathrm{C}_{\mathrm{ref}}$}}}%
    \put(0.71442766,0.08276546){\color[rgb]{0,0,0}\makebox(0,0)[lb]{\smash{$\mathrm{C}_{\text{+}1}$}}}%
    \put(0.46571651,0.10526519){\color[rgb]{0,0,0}\makebox(0,0)[lb]{\smash{$\mathcal{I}_{\mathrm{ref}}$}}}%
    \put(0.19872989,0.12526519){\color[rgb]{0,0,0}\makebox(0,0)[lb]{\smash{$\mathcal{I}_{\text{-}1}$}}}%
    \put(0.66470487,0.17226519){\color[rgb]{0,0,0}\makebox(0,0)[lb]{\smash{$\mathcal{I}_{\text{+}1}$}}}%
    \put(0.47571651,0.05585967){\color[rgb]{0.92941176,0.49019608,0.18823529}\makebox(0,0)[lb]{\smash{$\mathbf{H}_{\mathrm{ref}\rightarrow \text{+}1}$}}}%
    \put(0.29630143,0.06585967){\color[rgb]{0.92941176,0.49019608,0.18823529}\makebox(0,0)[rb]{\smash{$\mathbf{H}_{\mathrm{ref}\rightarrow \text{-}1}$}}}%
    \put(0.38960026,0.06973466){\color[rgb]{0,0.43921569,0.75294118}\makebox(0,0)[lb]{\smash{$\mathrm{p}^{\mathrm{ref}}$}}}%
    \put(0.63320574,0.14526519){\color[rgb]{0,0.43921569,0.75294118}\makebox(0,0)[lb]{\smash{$\mathrm{p}^{\text{+}1}$}}}%
    \put(0.13451145,0.1441993){\color[rgb]{0,0.43921569,0.75294118}\makebox(0,0)[lb]{\smash{$\mathrm{p}^{\text{-}1}$}}}%
    \end{footnotesize}
  \end{picture}%
\endgroup%

%% file: figures/ComputeDistanceOfPlane-edited.pdf_tex
%% Accompanies image file 'ComputeDistanceOfPlane.pdf' (pdf, eps, ps)
\begingroup%
  \makeatletter%
  \providecommand\color[2][]{%
    \errmessage{(Inkscape) Color is used for the text in Inkscape, but the package 'color.sty' is not loaded}%
    \renewcommand\color[2][]{}%
  }%
  \providecommand\transparent[1]{%
    \errmessage{(Inkscape) Transparency is used (non-zero) for the text in Inkscape, but the package 'transparent.sty' is not loaded}%
    \renewcommand\transparent[1]{}%
  }%
  \providecommand\rotatebox[2]{#2}%
  \newcommand*\fsize{\dimexpr\f@size pt\relax}%
  \newcommand*\lineheight[1]{\fontsize{\fsize}{#1\fsize}\selectfont}%
  \ifx\svgwidth\undefined%
    \setlength{\unitlength}{531.11915588bp}%
    \ifx\svgscale\undefined%
      \relax%
    \else%
      \setlength{\unitlength}{\unitlength * \real{\svgscale}}%
    \fi%
  \else%
    \setlength{\unitlength}{\svgwidth}%
  \fi%
  \global\let\svgwidth\undefined%
  \global\let\svgscale\undefined%
  \makeatother%
  \begin{picture}(1,0.88598295)%
    \lineheight{1}%
    \setlength\tabcolsep{0pt}%
    \put(0,0){\includegraphics[width=\unitlength,page=1]{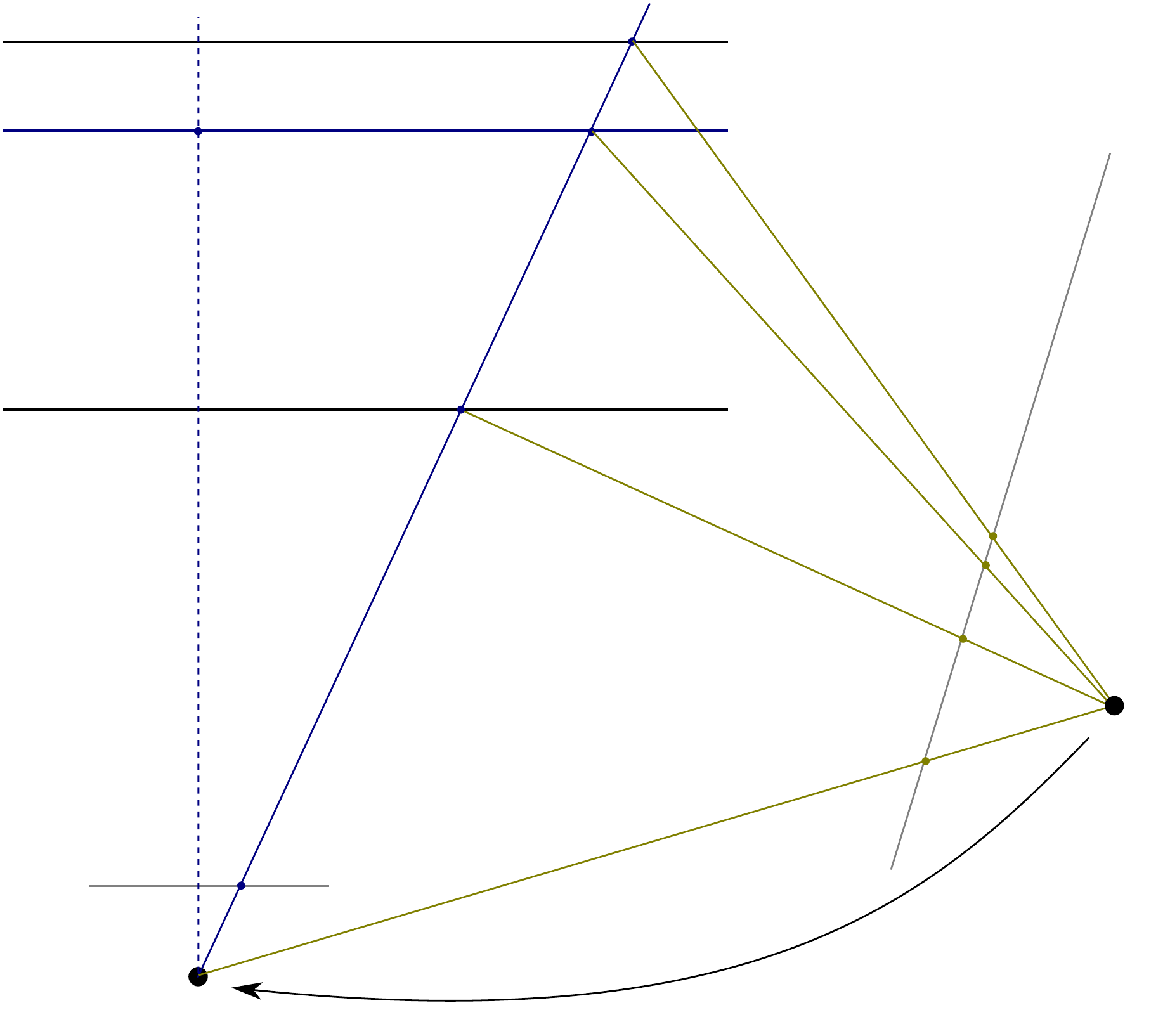}}%
    \put(0.61605705,0.024){\color[rgb]{0,0,0}\makebox(0,0)[lt]{\lineheight{1.25}\smash{\begin{tabular}[t]{l}$\left[\mathbf{R}\ \mathrm{t}\right]$\end{tabular}}}}%
    \put(0.9716927,0.26280413){\color[rgb]{0,0,0}\makebox(0,0)[lt]{\lineheight{1.25}\smash{\begin{tabular}[t]{l}$\mathrm{C}_k$\end{tabular}}}}%
    \put(0.17373716,0.00794043){\color[rgb]{0,0,0}\makebox(0,0)[lt]{\lineheight{1.25}\smash{\begin{tabular}[t]{l}$\mathrm{C}_{\mathrm{ref}}$\end{tabular}}}}%
    \put(0.06040944,0.09711628){\color[rgb]{0,0,0}\makebox(0,0)[lt]{\lineheight{1.25}\smash{\begin{tabular}[t]{l}$\mathcal{I}_{\mathrm{ref}}$\end{tabular}}}}%
    \put(0.00837801,0.54865382){\color[rgb]{0,0,0}\makebox(0,0)[lt]{\lineheight{1.25}\smash{\begin{tabular}[t]{l}$\Pi_{\mathrm{min}}$\end{tabular}}}}%
    \put(0.00837801,0.8618188){\color[rgb]{0,0,0}\makebox(0,0)[lt]{\lineheight{1.25}\smash{\begin{tabular}[t]{l}$\Pi_{\mathrm{max}}$\end{tabular}}}}%
    \put(0.96978262,0.75211018){\color[rgb]{0,0,0}\makebox(0,0)[lt]{\lineheight{1.25}\smash{\begin{tabular}[t]{l}$\mathrm{l}^k_{\mathrm{p}}$\end{tabular}}}}%
    \put(0.00837801,0.78589572){\color[rgb]{0,0,0.50196078}\makebox(0,0)[lt]{\lineheight{1.25}\smash{\begin{tabular}[t]{l}$\Pi$\end{tabular}}}}%
    \put(0.18,0.74){\color[rgb]{0,0,0.50196078}\makebox(0,0)[lt]{\lineheight{1.25}\smash{\begin{tabular}[t]{l}$\delta$\end{tabular}}}}%
    \put(0.18,0.5){\color[rgb]{0,0,0}\makebox(0,0)[lt]{\lineheight{1.25}\smash{\begin{tabular}[t]{l}$\delta_{\mathrm{min}}$\end{tabular}}}}%
    \put(0.18,0.815){\color[rgb]{0,0,0}\makebox(0,0)[lt]{\lineheight{1.25}\smash{\begin{tabular}[t]{l}$\delta_{\mathrm{max}}$\end{tabular}}}}%
    \put(0.21,0.10){\color[rgb]{0,0,0.50196078}\makebox(0,0)[lt]{\lineheight{1.25}\smash{\begin{tabular}[t]{l}$\mathrm{p}^{\mathrm{ref}}$\end{tabular}}}}%
    \put(0.335,0.55){\color[rgb]{0,0,0.50196078}\makebox(0,0)[lt]{\lineheight{1.25}\smash{\begin{tabular}[t]{l}$\mathrm{P}_{\mathrm{min}}$\end{tabular}}}}%
    \put(0.475,0.79){\color[rgb]{0,0,0.50196078}\makebox(0,0)[lt]{\lineheight{1.25}\smash{\begin{tabular}[t]{l}$\mathrm{P}_i$\end{tabular}}}}%
    \put(0.475,0.86){\color[rgb]{0,0,0.50196078}\makebox(0,0)[lt]{\lineheight{1.25}\smash{\begin{tabular}[t]{l}$\mathrm{P}_{\mathrm{max}}$\end{tabular}}}}%
    \put(0.79338253,0.205){\color[rgb]{0.50196078,0.50196078,0}\makebox(0,0)[lt]{\lineheight{1.25}\smash{\begin{tabular}[t]{l}$\mathrm{e}^k_{\mathrm{ref}}$\end{tabular}}}}%
    \put(0.82042729,0.298){\color[rgb]{0.50196078,0.50196078,0}\makebox(0,0)[lt]{\lineheight{1.25}\smash{\begin{tabular}[t]{l}$\mathrm{p}^k_{\mathrm{min}}$\end{tabular}}}}%
    \put(0.79,0.39132463){\color[rgb]{0.50196078,0.50196078,0}\makebox(0,0)[lt]{\lineheight{1.25}\smash{\begin{tabular}[t]{l}$\mathrm{p}^k_{i}$\end{tabular}}}}%
    \put(0.86202903,0.42831923){\color[rgb]{0.50196078,0.50196078,0}\makebox(0,0)[lt]{\lineheight{1.25}\smash{\begin{tabular}[t]{l}$\mathrm{p}^k_{\mathrm{max}}$\end{tabular}}}}%
    \put(0.26,0.36144768){\color[rgb]{0,0,0.50196078}\makebox(0,0)[lt]{\lineheight{1.25}\smash{\begin{tabular}[t]{l}$\mathrm{V}^{\mathrm{ref}}_{\mathrm{p}}$\end{tabular}}}}%
    \put(0.52,0.18662127){\color[rgb]{0.50196078,0.50196078,0}\makebox(0,0)[lt]{\lineheight{1.25}\smash{\begin{tabular}[t]{l}$\mathrm{V}^k_{\mathrm{e}_{\mathrm{ref}}}$\end{tabular}}}}%
    \put(0.6355073,0.43637345){\color[rgb]{0.50196078,0.50196078,0}\makebox(0,0)[lt]{\lineheight{1.25}\smash{\begin{tabular}[t]{l}$\mathrm{V}^k_{\mathrm{p}_{\mathrm{min}}}$\end{tabular}}}}%
    \put(0.6321207,0.565){\color[rgb]{0.50196078,0.50196078,0}\makebox(0,0)[lt]{\lineheight{1.25}\smash{\begin{tabular}[t]{l}$\mathrm{V}^k_{\mathrm{p}_{i}}$\end{tabular}}}}%
    \put(0.70084382,0.64461523){\color[rgb]{0.50196078,0.50196078,0}\makebox(0,0)[lt]{\lineheight{1.25}\smash{\begin{tabular}[t]{l}$\mathrm{V}^k_{\mathrm{p}_{\mathrm{max}}}$\end{tabular}}}}%
  \end{picture}%
\endgroup%

%% file: figures/xsgm-edited.pdf_tex
%% Accompanies image file 'xsgm.pdf' (pdf, eps, ps)
\begingroup%
  \makeatletter%
  \providecommand\color[2][]{%
    \errmessage{(Inkscape) Color is used for the text in Inkscape, but the package 'color.sty' is not loaded}%
    \renewcommand\color[2][]{}%
  }%
  \providecommand\transparent[1]{%
    \errmessage{(Inkscape) Transparency is used (non-zero) for the text in Inkscape, but the package 'transparent.sty' is not loaded}%
    \renewcommand\transparent[1]{}%
  }%
  \providecommand\rotatebox[2]{#2}%
  \newcommand*\fsize{\dimexpr\f@size pt\relax}%
  \newcommand*\lineheight[1]{\fontsize{\fsize}{#1\fsize}\selectfont}%
  \ifx\svgwidth\undefined%
    \setlength{\unitlength}{850.32000732bp}%
    \ifx\svgscale\undefined%
      \relax%
    \else%
      \setlength{\unitlength}{\unitlength * \real{\svgscale}}%
    \fi%
  \else%
    \setlength{\unitlength}{\svgwidth}%
  \fi%
  \global\let\svgwidth\undefined%
  \global\let\svgscale\undefined%
  \makeatother%
  \begin{picture}(1,0.2001129)%
  	\begin{footnotesize}
    \lineheight{1}%
    \setlength\tabcolsep{0pt}%
    \put(0,0){\includegraphics[width=\unitlength,page=1]{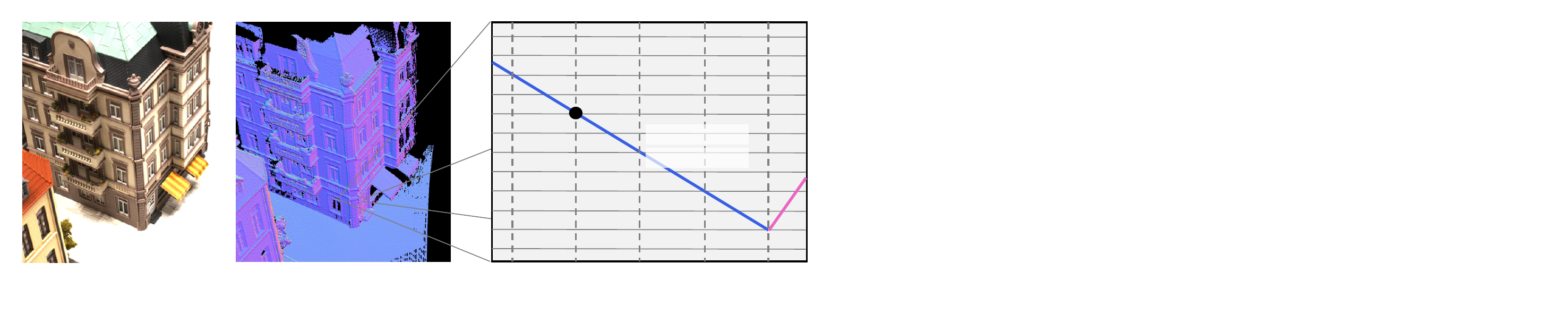}}%
    \put(0,0){\includegraphics[width=\unitlength,page=2]{xsgm.pdf}}%
    \put(0,0){\includegraphics[width=\unitlength,page=3]{xsgm.pdf}}%
    \put(0,0){\includegraphics[width=\unitlength,page=4]{xsgm.pdf}}%
    \put(0,0){\includegraphics[width=\unitlength,page=5]{xsgm.pdf}}%
    \put(0,0){\includegraphics[width=\unitlength,page=6]{xsgm.pdf}}%
    \put(0,0){\includegraphics[width=\unitlength,page=7]{xsgm.pdf}}%
    \put(0,0){\includegraphics[width=\unitlength,page=8]{xsgm.pdf}}%
    \put(0,0){\includegraphics[width=\unitlength,page=9]{xsgm.pdf}}%
    \put(0,0){\includegraphics[width=\unitlength,page=10]{xsgm.pdf}}%
    \put(0,0){\includegraphics[width=\unitlength,page=11]{xsgm.pdf}}%
    \put(0,0){\includegraphics[width=\unitlength,page=12]{xsgm.pdf}}%
    \put(0,0){\includegraphics[width=\unitlength,page=13]{xsgm.pdf}}%
    \put(0,0){\includegraphics[width=\unitlength,page=14]{xsgm.pdf}}%
    \put(0,0){\includegraphics[width=\unitlength,page=15]{xsgm.pdf}}%
    \put(0,0){\includegraphics[width=\unitlength,page=16]{xsgm.pdf}}%
    \put(0,0){\includegraphics[width=\unitlength,page=17]{xsgm.pdf}}%
    \put(0,0){\includegraphics[width=\unitlength,page=18]{xsgm.pdf}}%
    \put(0,0){\includegraphics[width=\unitlength,page=19]{xsgm.pdf}}%
    \put(0,0){\includegraphics[width=\unitlength,page=20]{xsgm.pdf}}%
    \put(0.36556824,0.19093988){\color[rgb]{0,0,0}\makebox(0,0)[lt]{\lineheight{1.25}\smash{\begin{tabular}[t]{l}$\mathrm{p}$\end{tabular}}}}%
    \put(0.39113509,0.19093988){\color[rgb]{0,0,0}\makebox(0,0)[lt]{\lineheight{1.25}\smash{\begin{tabular}[t]{l}$\mathrm{p}'\text{=}\mathrm{p}\text{+}\mathrm{r}$\end{tabular}}}}%
    \put(0.31930331,0.01563647){\color[rgb]{0,0,0}\makebox(0,0)[lt]{\lineheight{1.25}\smash{\begin{tabular}[t]{l}$\mathrm{r}$\end{tabular}}}}%
    \put(0.29188897,0.02900283){\color[rgb]{0,0,0}\makebox(0,0)[lt]{\lineheight{1.25}\smash{\begin{tabular}[t]{l}$\Pi$\end{tabular}}}}%
    \put(0.33318702,0.11399238){\color[rgb]{0,0,0}\makebox(0,0)[lt]{\lineheight{1.25}\smash{\begin{tabular}[t]{l}$\mathcal{S}(\mathrm{p}, \Pi)$\end{tabular}}}}%
    \put(0.60550144,0.19093988){\color[rgb]{0,0,0}\makebox(0,0)[lt]{\lineheight{1.25}\smash{\begin{tabular}[t]{l}$\mathrm{p}$\end{tabular}}}}%
    \put(0.63108004,0.19093988){\color[rgb]{0,0,0}\makebox(0,0)[lt]{\lineheight{1.25}\smash{\begin{tabular}[t]{l}$\mathrm{p}'\text{=}\mathrm{p}\text{+}\mathrm{r}$\end{tabular}}}}%
    \put(0.55924826,0.01563647){\color[rgb]{0,0,0}\makebox(0,0)[lt]{\lineheight{1.25}\smash{\begin{tabular}[t]{l}$\mathrm{r}$\end{tabular}}}}%
    \put(0.53183393,0.02900283){\color[rgb]{0,0,0}\makebox(0,0)[lt]{\lineheight{1.25}\smash{\begin{tabular}[t]{l}$\Pi$\end{tabular}}}}%
    \put(0.58222709,0.11233418){\color[rgb]{0,0,0}\makebox(0,0)[lt]{\lineheight{1.25}\smash{\begin{tabular}[t]{l}$\mathrm{n}_{\mathrm{p}}$\end{tabular}}}}%
    \put(0.84730452,0.19093988){\color[rgb]{0,0,0}\makebox(0,0)[lt]{\lineheight{1.25}\smash{\begin{tabular}[t]{l}$\mathrm{p}$\end{tabular}}}}%
    \put(0.87290664,0.19093988){\color[rgb]{0,0,0}\makebox(0,0)[lt]{\lineheight{1.25}\smash{\begin{tabular}[t]{l}$\mathrm{p}'\text{=}\mathrm{p}\text{+}\mathrm{r}$\end{tabular}}}}%
    \put(0.80107486,0.01563647){\color[rgb]{0,0,0}\makebox(0,0)[lt]{\lineheight{1.25}\smash{\begin{tabular}[t]{l}$\mathrm{r}$\end{tabular}}}}%
    \put(0.77366053,0.029000283){\color[rgb]{0,0,0}\makebox(0,0)[lt]{\lineheight{1.25}\smash{\begin{tabular}[t]{l}$\Pi$\end{tabular}}}}%
    \put(0.86296727,0.1031336){\color[rgb]{0,0,0}\makebox(0,0)[lt]{\lineheight{1.25}\smash{\begin{tabular}[t]{l}$\nabla_{\mathrm{r}}$\end{tabular}}}}%
    \put(0.81484427,0.1061812){\color[rgb]{0,0,0}\makebox(0,0)[lt]{\lineheight{1.25}\smash{\begin{tabular}[t]{l}$\Delta i_{\mathrm{pg}}$\end{tabular}}}}%
    \put(0.86666193,0.04180074){\color[rgb]{0,0,0}\makebox(0,0)[lt]{\lineheight{1.25}\smash{\begin{tabular}[t]{l}\tiny{- - - - - }\scriptsize{ Minimum Cost Path}\end{tabular}}}}%
%    \put(0.87103676,0.04180074){\color[rgb]{0,0,0}\makebox(0,0)[lt]{\lineheight{1.25}\smash{\begin{tabular}[t]{l}-\end{tabular}}}}%
%    \put(0.87541158,0.04180074){\color[rgb]{0,0,0}\makebox(0,0)[lt]{\lineheight{1.25}\smash{\begin{tabular}[t]{l}-\end{tabular}}}}%
%    \put(0.87992753,0.04180074){\color[rgb]{0,0,0}\makebox(0,0)[lt]{\lineheight{1.25}\smash{\begin{tabular}[t]{l}-\end{tabular}}}}%
%    \put(0.88430235,0.04180074){\color[rgb]{0,0,0}\makebox(0,0)[lt]{\lineheight{1.25}\smash{\begin{tabular}[t]{l}-\end{tabular}}}}%
%    \put(0.88867717,0.04180074){\color[rgb]{0,0,0}\makebox(0,0)[lt]{\lineheight{1.25}\smash{\begin{tabular}[t]{l}-\end{tabular}}}}%
    \put(0.39607176,0.00739487){\color[rgb]{0,0,0}\makebox(0,0)[lt]{\lineheight{1.25}\smash{\begin{tabular}[t]{l}\textbf{\piSGM}\end{tabular}}}}%
    \put(0.63526583,0.00739487){\color[rgb]{0,0,0}\makebox(0,0)[lt]{\lineheight{1.25}\smash{\begin{tabular}[t]{l}\textbf{\snSGM}\end{tabular}}}}%
    \put(0.87642166,0.00739487){\color[rgb]{0,0,0}\makebox(0,0)[lt]{\lineheight{1.25}\smash{\begin{tabular}[t]{l}\textbf{\pgSGM}\end{tabular}}}}%
    \put(0.41313909,0.14164924){\color[rgb]{0,0,0}\makebox(0,0)[lt]{\lineheight{1.25}\smash{\begin{tabular}[t]{l}$\mathcal{S}(\mathrm{p}', \Pi\text{-}1)$\end{tabular}}}}%
    \put(0.41313909,0.12570964){\color[rgb]{0,0,0}\makebox(0,0)[lt]{\lineheight{1.25}\smash{\begin{tabular}[t]{l}$\mathcal{S}(\mathrm{p}', \Pi)$\end{tabular}}}}%
    \put(0.4125798,0.11200718){\color[rgb]{0,0,0}\makebox(0,0)[lt]{\lineheight{1.25}\smash{\begin{tabular}[t]{l}$\mathcal{S}(\mathrm{p}', \Pi\text{+}1)$\end{tabular}}}}%
    \put(0.4125798,0.09662687){\color[rgb]{0,0,0}\makebox(0,0)[lt]{\lineheight{1.25}\smash{\begin{tabular}[t]{l}$\mathcal{S}(\mathrm{p}', \Pi\text{+}2)$\end{tabular}}}}%
    \put(0.61056637,0.1394121){\color[rgb]{0,0,0}\makebox(0,0)[lt]{\lineheight{1.25}\smash{\begin{tabular}[t]{l}$\mathcal{S}(\mathrm{p}, \Pi)$\end{tabular}}}}%
    \put(0.618676,0.11088862){\color[rgb]{0,0,0}\makebox(0,0)[lt]{\lineheight{1.25}\smash{\begin{tabular}[t]{l}$\Delta i_{\mathrm{sn}}$\end{tabular}}}}%
    \put(0.65292394,0.10054186){\color[rgb]{0,0,0}\makebox(0,0)[lt]{\lineheight{1.25}\smash{\begin{tabular}[t]{l}$\mathcal{S}(\mathrm{p}', \Pi\text{+}\Delta i_{\mathrm{sn}})$\end{tabular}}}}%
    \put(0.65264431,0.11620181){\color[rgb]{0,0,0}\makebox(0,0)[lt]{\lineheight{1.25}\smash{\begin{tabular}[t]{l}$\mathcal{S}(\mathrm{p}', \Pi\text{+}\Delta i_{\mathrm{sn}}\text{-}1)$\end{tabular}}}}%
    \put(0.65292394,0.0848819){\color[rgb]{0,0,0}\makebox(0,0)[lt]{\lineheight{1.25}\smash{\begin{tabular}[t]{l}$\mathcal{S}(\mathrm{p}', \Pi\text{+}\Delta i_{\mathrm{sn}}\text{+}1)$\end{tabular}}}}%
    \put(0.85245676,0.13969175){\color[rgb]{0,0,0}\makebox(0,0)[lt]{\lineheight{1.25}\smash{\begin{tabular}[t]{l}$\mathcal{S}(\mathrm{p}, \Pi)$\end{tabular}}}}%
    \put(0.89515468,0.10305864){\color[rgb]{0,0,0}\makebox(0,0)[lt]{\lineheight{1.25}\smash{\begin{tabular}[t]{l}$\mathcal{S}(\mathrm{p}', \Pi\text{+}\Delta i_{\mathrm{pg}}\text{-}1)$\end{tabular}}}}%
    \put(0.8951547,0.08683939){\color[rgb]{0,0,0}\makebox(0,0)[lt]{\lineheight{1.25}\smash{\begin{tabular}[t]{l}$\mathcal{S}(\mathrm{p}', \Pi\text{+}\Delta i_{\mathrm{pg}})$\end{tabular}}}}%
    \put(0.8951547,0.0708998){\color[rgb]{0,0,0}\makebox(0,0)[lt]{\lineheight{1.25}\smash{\begin{tabular}[t]{l}$\mathcal{S}(\mathrm{p}', \Pi\text{+}\Delta i_{\mathrm{pg}}\text{+}1)$\end{tabular}}}}%
    %\put(0.89263904,0.04293559){\color[rgb]{0,0,0}\makebox(0,0)[lt]{\lineheight{1.25}\smash{\begin{tabular}[t]{l}Minimum Cost Path $L_{\mathrm{r}}(\mathrm{p}, \Pi)$\end{tabular}}}}%
  	\end{footnotesize}
  \end{picture}%
\endgroup%

%% file: contents/04_experiments.tex
%%%%%%%%%%%%%%%%%%%%%%%%%%%%%%%%%%%%%%%%%%%%%%%
\section{Results}
\label{sec:experiments}
%%%%%%%%%%%%%%%%%%%%%%%%%%%%%%%%%%%%%%%%%%%%%%%
% KAO: Sloppy spacing ensures non-overfull lines. Can be removed if this is not an issue.
\sloppy

The following sections present the results of experiments conducted in the scope of this work. %
In these, the effects of different configurations of the presented approach are evaluated and analyzed with respect to individual aspects, such as accuracy, efficiency and application-specific usability. %
First, the datasets and evaluation metrics used to investigate the potential of the presented approach are introduced in \Cref{sec:experiments_datasets} and \Cref{sec:experiments_metrics}, respectively. %
Then, the need for a hierarchical processing is evaluated and the optimal number of input images, \ie the optimal size of the input bundle, is empirically found in \Cref{sec:experiments_plane-sweep}. %
This is followed by a short comparison in \Cref{sec:experiments_sim-measure}, where the focus is set on the two similarity metrics and cost functions that are implemented in the scope of this work. %
In \Cref{sec:experiments_surface-aware}, the ability of the three \gls*{SGM} extensions to reconstruct non-fronto-parallel surface structures is evaluated and compared to the effects of using a non-fronto-parallel plane orientation within plane-sweep sampling. %
An evaluation of the improvements gained by post-filtering is presented in \Cref{sec:experiments_post-filtering}, before comparing the results of the best configurations to those achieved by approaches for offline \gls*{MVS}, \eg \COLMAP \citep{Schoenberger2016mvs}, in \Cref{sec:experiments_comparisonToColmap}. %
Finally, before discussing the presented results, the outcomes of use-case-specific experiments are shown and qualitatively illustrated in \Cref{sec:experiments_qualitative}. %

The complete processing pipeline of the presented approach, except the generation of the Gaussian image pyramids and the parameterization of the plane-sweep algorithm, is implemented in CUDA and is thus optimized for massively parallel computing by \gls*{GPGPU}, which, in turn, is embedded in a C++ application.
All experiments, and thus all timing measurements, were conducted using a NVIDIA Titan X GPU and an Intel XEON CPU E5-2650 running with 2.20\GHz. %
Even though the CPU is designed for a server architecture, only a small part of our approach is run on the CPU, and thus its superiority over commodity desktop hardware is insignificant. %

\subsection{Evaluation Datasets}
\label{sec:experiments_datasets}

The presented approach is quantitatively evaluated on two public datasets, which also provide an appropriate ground truth, namely the DTU Robot \gls*{MVS} dataset \citep{Jensen2014dtu, Aanaes2016Large} and the dataset from the 3DOMcity Benchmark \citep{Osdemir2019multi}. %
In order to provide appropriate data for a quantitative evaluation, these two datasets rely on images of scale-modeled buildings and an urban scenery from which an accurate ground truth is acquired.
For a qualitative evaluation and a discussion regarding the usability of the presented approach for online dense image matching and 3D reconstruction, two privately captured datasets of real-world scenes are used, henceforth referred to as the TMB and the FB dataset. %
In the following, the characteristics of these datasets are briefly introduced.
In particular, which portions of the data sets are used and what kind of ground truth is available for the evaluation. %

\begin{figure*}[t!]%
    \centering%
    \includegraphics[width=\textwidth]{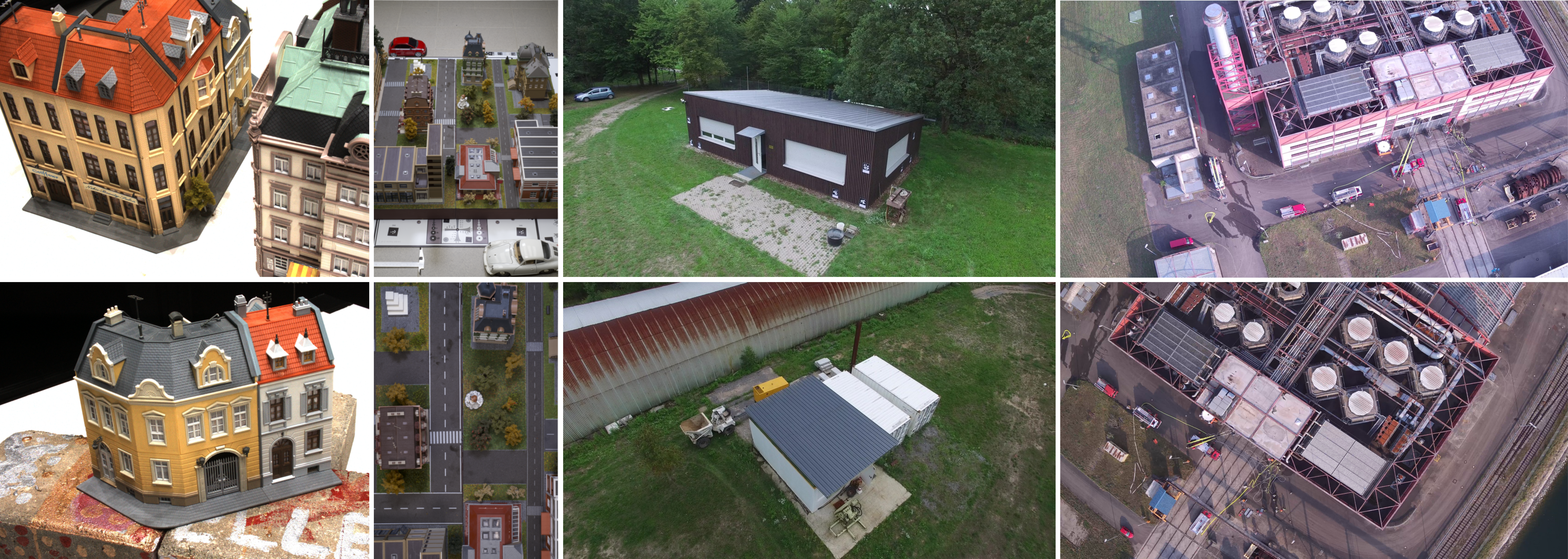} %
    \caption{ %
    		Overview of the four datasets used for performance evaluation in the scope of this paper. %
    		\textbf{Column~1:} Two building models from the DTU Robot MVS dataset. %
    		\textbf{Column~2:} Example images in oblique and nadir view from the 3DOMcity Benchmark dataset. %
    		\textbf{Column~3:} Excerpt of the privately acquired TMB dataset. %
    		\textbf{Column~4:} Use-case-specific dataset acquired during an exercise of the local fire brigade. %
    	} %
    \label{fig:datasets}%
\end{figure*}%

\paragraph{DTU Robot MVS Dataset}

The DTU Robot MVS dataset (\Cref{fig:datasets}, Column~1) is comprised of 124 different table top scenes. %
For each scene, input images are provided, which were captured under eight different lighting conditions from 49 locations. %
These locations are the same for all scenes, since the camera is mounted on an industrial robot arm, which could repeatedly be moved to the set camera poses. %
Furthermore, for each scene, a ground truth is provided in the form of a detailed point cloud captured by a structured light scanner. %

For the quantitative evaluation of the performance of the presented approach, 21 scans of different building models were selected, since these scenes are closest to the targeted use-case, namely the online reconstruction of urban scenes and man-made structures. %
In this, the already undistorted images with an image resolution of $1600\times1200$\px, together with the corresponding intrinsic and extrinsic camera data, were used as input data to the approach. %
The benchmark provides an evaluation routine, and a corresponding score table, for the assessment of a reconstructed point cloud with respect to the ground truth captured by the structured light scanner.
However, the focus of this work lies in the estimation of depth and normal maps only, without the subsequent fusion into a three-dimensional point cloud. %
Thus, in order to evaluate the accuracy of the estimated depth and normal maps, corresponding ground truth data is extracted from the structured light scans by rendering them from the view points of the corresponding input images, given the intrinsic and extrinsic camera data. %

\paragraph{3DOMcity Benchmark Dataset}

Within the DTU benchmark, the camera is moved along a circular trajectory, resembling an orbital flight of a \gls*{UAV}, with the camera focusing on the object of interest.
This kind of camera movement or flight trajectory is typical for the case when an object is to be fully reconstructed with a maximum precision \citep{Wenzel2013image}. %
However, depending on the aircraft and the surrounding constraints, such a flight is not always feasible or desired. %
Typical for the mapping of a larger area, a grid flight is performed, where the aircraft is flying linearly over the area of interest, with the camera orientation being fixed with respect to the sensor carrier. %

Such a configuration is simulated by the data provided as part of the 3DOMcity Benchmark \citep{Osdemir2019multi} (\Cref{fig:datasets}, Column~2). %
In this, images of a scale-modeled urban scene, comprised of differently sized and shaped buildings, as well as streets and vegetation, are captured with a DSLR camera moved along a rigid bar in parallel lines over the model. %
The distorted and undistorted images are provided with a maximum resolution of $6016\times4016$\px from oblique and nadir vantage points, with a forward and sideways overlap of $85/75\percent$ in case of the oblique views and $80/65\percent$ between the nadir views. %
While the benchmark internally uses a point cloud of two exemplary building models, captured by a multi-stripe triangulation-based laser scanner, to assess the accuracy of submitted \gls*{DIM} point clouds, it publicly provides a reference point cloud for the task of classification, which is computed by a semi-global \gls*{DIM} algorithm \citep{Osdemir2019multi}. %

To use the data of the 3DOMcity Benchmark for the quantitative evaluation of the performance of the presented approach, the already undistorted images are first down-scaled to a size of $1798\times1200$\px, preserving the initial aspect ratio, before estimating the intrinsic camera parameters with the help of COLMAP \citep{Schoenberger2016sfm}. %
The extrinsic camera data is extracted from the reference that is provided as part of the benchmark. %
For the accuracy assessment of the depth maps, the reference data is computed by rendering the reference \gls*{DIM} point cloud of the whole model, provided for the task of point cloud classification, from the viewpoints of the input images, just as in the case of the dataset of the DTU benchmark. %

\paragraph{Real-World Use-Case-Specific TMB and FB Dataset}

The strength and aim of the datasets from the DTU and 3DOMcity benchmarks are their small scale and the associated ability to record or compute accurate reference data, which in turn facilitates a quantitative evaluation of the accuracy of the assessed algorithms. %
These datasets, however, are recorded under controlled environments and do not fully accommodate for the use-case aimed at with the presented approach, namely the online \gls*{MVS} from monocular video data captured by commodity \glspl*{UAV}, flying at altitudes below $100$\m. %
In order to account for the named use-case and perform a qualitative evaluation on real-world data, appropriate test data was collected into a private dataset. %

This dataset is two-fold. %
The first part, namely the TMB dataset (\Cref{fig:datasets}, Column~3), consists of four sequences captured by a DJI Phantom $3$ professional, flying around a free-standing house and  containers at altitudes between $8$\m to $15$\m. %
For the second part, which is denoted as the fire brigade (FB) dataset (\Cref{fig:datasets}, Column~4), images were acquired during a fire brigade exercise around a big industrial building. %
The data was captured using a DJI Matrice 200 with a Zenmuse XT2 sensor flying linearly over the area on which the exercise was performed.
%While the \gls*{UAV} used to capture the TMB dataset was piloted manually, resulting in flight trajectory that is not strictly orbital with varying velocity, the flight of the DJI Mavic, capturing the IOSB dataset, was planned and executed by software, resulting in a steadier trajectory with a constant velocity.
For all sequences, the images were captured at a frame-rate of about $1$\fps and down-sampled to an image size of $1920\times1080$\px. %
However, due to the varying velocities, the distances between the frames are not always the same and thus images that are not appropriate as input to the presented approach, \eg by providing too little offset, are discarded. %
%To pre-process the image data in order for it to be used as input to the presented approach, \ie computing the corresponding camera projection matrices and undistorting the input image, the \gls*{SFM} pipeline of COLMAP \citep{Schoenberger2016sfm} was used. %
As reference data, detailed point clouds of each sequence were computed by means of \gls*{SFM} and \gls*{MVS} using \COLMAP \citep{Schoenberger2016sfm, Schoenberger2016mvs}. %
In order to have a metric reference, given the GNSS metadata provided by the \glspl*{UAV} for each input image, the reference data was transformed into a local \gls*{ENU} frame. %
The undistorted images as well as the intrinsic and extrinsic camera data produced by \COLMAP serve as input to the presented approach for the evaluation with respect to the TMB and FB dataset. %

\subsection{Error Measures}
\label{sec:experiments_metrics}

For a direct quantification of the error between the estimated depth map $\mathcal{D}_{\mathrm{est}}$ and the corresponding ground truth $\mathcal{D}_{\mathrm{gt}}$, absolute and relative \LOne measures are used:
\begin{equation}
\begin{aligned}
\text{L1-abs}(\mathcal{D}_{\mathrm{est}}, \mathcal{D}_{\mathrm{gt}}) = \frac{1}{|\mathcal{V}|}\sum\limits_{\mathrm{p}\in\mathcal{V}}{|\mathcal{D}_{\mathrm{est}}(\mathrm{p}) - \mathcal{D}_{\mathrm{gt}}(\mathrm{p})|}\ ,\ \ \text{and}\ 
\label{eq:l1abs-measure}
\end{aligned}
\end{equation}
\begin{equation}
\label{eq:l1rel-measure}
\begin{aligned}
\text{L1-rel}(\mathcal{D}_{\mathrm{est}}, \mathcal{D}_{\mathrm{gt}}) = \frac{1}{|\mathcal{V}|}\sum\limits_{\mathrm{p}\in\mathcal{V}}{\frac{|\mathcal{D}_{\mathrm{est}}(\mathrm{p}) - \mathcal{D}_{\mathrm{gt}}(\mathrm{p})|}{\mathcal{D}_{\mathrm{gt}}(\mathrm{p})}}\ . \ \ \ \ \ \ \ \
\end{aligned}
\end{equation}
In this, $\mathcal{V}$ denotes the set of pixels for which both $\mathcal{D}_{\mathrm{est}}$ and $\mathcal{D}_{\mathrm{gt}}$ have valid depth measurements. %
While \LOneAbs provides an absolute and, in turn, interpretable insight on the mean error of the estimated depth map, it is rather unsuitable for comparing the results across multiple datasets with different depth ranges.
This is because the error of depth measurements typically increases with increasing depth, which leads to a higher absolute error for datasets with a larger scene depth. %
In order to compensate for this effect, the relative \LOneRel measure normalizes the absolute difference by the depth stored at the corresponding ground truth pixel. %
This reduces the effect that erroneous pixels in distant areas of the scene have on the error score, while at the same time increasing the weight of the pixels that are close to the camera. %

The two error measures introduced above provide a simple strategy to evaluate the error of the estimates. %
However, they do not allow to reason about the completeness and density of the estimated depth map. %
Thus, since the focus of this work is on dense \gls*{MVS}, it is also of great interest to know how many pixels of $\mathcal{D}_{\mathrm{est}}$ are actually filled with correct estimates. %
To do so, two tightly coupled error scores, namely the accuracy ($\text{Acc}_{\theta}$) and completeness ($\text{Cpl}_{\theta}$), are also used. %
These scores are typically used to evaluate classification tasks, but have also been used to evaluate range measurements in recent years  \citep{Schoeps2017eth3d, Knapitsch2017tanks}. 
On the one hand, the accuracy $\text{Acc}_{\theta}$ indicates the amount of pixels within the estimated depth map $\mathcal{D}_{\mathrm{est}}$, for which the corresponding depth value is within a given threshold $\theta$ to the ground truth: 
\begin{equation}
	\label{eq:acc-measure}
    \text{Acc}_{\theta}(\mathcal{D}_{\mathrm{est}}, \mathcal{D}_{\mathrm{gt}}) =\frac{1}{|\mathcal{E}|}\sum\limits_{\mathrm{p}\in \mathcal{V}}\left[  \max\left(\frac{ \mathcal{D}_{\mathrm{est}}(\mathrm{p}) }{\mathcal{D}_{\mathrm{gt}}(\mathrm{p})}, \frac{\mathcal{D}_{\mathrm{gt}}(\mathrm{p})}{\mathcal{D}_{\mathrm{est}}(\mathrm{p})}\right) < \theta \right]\ .
\end{equation}
The completeness $\text{Cpl}_{\theta}$, on the other hand, indicates the fraction of the ground truth pixels, for which estimates exist, which are within the given distance threshold to the reference:
\begin{equation}
	\label{eq:cpl-measure}
    \text{Cpl}_{\theta}(\mathcal{D}_{\mathrm{est}}, \mathcal{D}_{\mathrm{gt}}) =\frac{1}{|\mathcal{G}|}\sum\limits_{\mathrm{p}\in \mathcal{V}}\left[  \max\left(\frac{ \mathcal{D}_{\mathrm{est}}(\mathrm{p}) }{\mathcal{D}_{\mathrm{gt}}(\mathrm{p})}, \frac{\mathcal{D}_{\mathrm{gt}}(\mathrm{p})}{\mathcal{D}_{\mathrm{est}}(\mathrm{p})}\right) < \theta \right]\ . 
\end{equation}
Again, $\mathcal{V}$ holds the set of pixels for which both $\mathcal{D}_{\mathrm{est}}$ and $\mathcal{D}_{\mathrm{gt}}$ have valid depth measurements. %
Similarly, $\mathcal{E}$ denotes the pixel set with valid estimates, while $\mathcal{G}$ holds the pixels with valid ground truth values. %
In both \Cref{eq:acc-measure} and \Cref{eq:cpl-measure}, the operator $[\cdot]$ refers to the Iverson bracket. %
The threshold $\theta$ is given as the percentage of the corresponding ground truth value. %
For example, $\text{Acc}_{1.25}$ and $\text{Cpl}_{1.25}$ hold the fraction of pixels with respect to the $\mathcal{D}_{\mathrm{est}}$ and $\mathcal{D}_{\mathrm{gt}}$, for which the difference between the estimate and ground truth is smaller than $25$\percent of the corresponding ground truth depth. %
These two measures can further be summarized into a combined score, namely the $\text{F}_{\theta}$-score, which is the harmonic mean between the $\text{Acc}_{\theta}$ and $\text{Cpl}_{\theta}$: 
\begin{equation}
	\label{eq:fscore-measure}
    \text{F}_{\theta}(\mathcal{D}_{\mathrm{est}}, \mathcal{D}_{\mathrm{gt}}) =2\cdot \frac{\text{Acc}_{\theta} \cdot \text{Cpl}_{\theta}}{\text{Acc}_{\theta} + \text{Cpl}_{\theta}}\ .
\end{equation}
Thus, a high $\text{F}_{\theta}$-score indicates a good trade-off between the achieved accuracy of the depth map and its completeness with respect to the ground truth. %

%Finally, to evaluate the the correctness of the estimated normal maps, the \gls*{AAE} between $\mathcal{N}_{\mathrm{est}}$ and $\mathcal{N}_{\mathrm{gt}}$ is calculated according to:
%%
%\begin{equation}
%\begin{aligned}
%\text{AAE}(\mathcal{N}_{\mathrm{est}}, \mathcal{N}_{\mathrm{gt}}) = \frac{1}{|\mathcal{V}|}\sum\limits_{\mathrm{p}\in\mathcal{V}}{\cos^{-1}\left\langle\mathcal{N}_{\mathrm{est}}(\mathrm{p}), \mathcal{N}_{\mathrm{gt}}(\mathrm{p})\right\rangle}\ .\ 
%\label{eq:aae-measure}
%\end{aligned}
%\end{equation}
%%
%In this, the pixel-wise scalar product between the normalized vectors of the estimate and the ground truth is calculated, from which the angle between the two vectors is computed. %
%Again, the mean error is calculated by normalizing by the size of $\mathcal{V}$, holding the pixels for which both an estimate and ground truth exists. %

\subsection{Need for Hierarchical Processing and Finding the Optimal Number of Input Images}
\label{sec:experiments_plane-sweep}

In the first experiment, the need and importance of the hierarchical processing scheme within the presented approach as well as the best configuration on the size of the input bundle are evaluated. %
In this, a couple of aspects are considered in order to find the best configuration for the succeeding experiments. %
The objective is to find the appropriate number $n$ of Gaussian pyramid levels and the size $|\Omega|$ of the input bundle, providing a good trade-off between %
\begin{itemize} %
\item the error of the resulting depth maps, measured by L1-abs and L1-rel, %
\item the sampling density of the scene and the entailed resources needed for the computation, %
\item as well as the resulting processing run-time. %
\end{itemize} %
For this experiment, a fronto-parallel plane orientation is used as part of the plane-sweep image matching and the \gls*{NCC} with a support region of $5\times5$\px is set as similarity measure. %
The optimization of the cost volume and the extraction of the optimal depth map is done by employing the \piSGM scheme, which is the adoption of the standard \gls*{SGM} optimization to the use of plane-sweep image matching (\cf \Cref{sec:methodology_sgm_x}). %
The smoothness penalty within the \gls*{SGM} optimization is set to $\varphi_1 = 100$, while the adaptive $\varphi_2$ penalty as described in \Cref{sec:adaptive_p2} is used. %
This, together with the $5\times5$ sized \gls*{NCC} as matching cost, was chosen in accordance with the work of \citet{Scharstein2018surface}.
%The additionally computed normal and confidence maps are not evaluated as part of this experiment. %
To find the appropriate height of the Gaussian pyramids, the size of the input bundle, \ie the number of input images, is first set to $|\Omega| = 3$. %

\begin{table}[t] 
\caption{Mean \LOne errors achieved on the DTU and 3DOMcity dataset for a different number $n$ of Gaussian pyramid levels as part of the hierarchical processing scheme. The error metrics used are the absolute L1-abs, measured in mm, as well as the relative L1-rel measure. Both are averaged over all evaluated depth maps within each dataset.}
\label{tab:pyrLvls-accuracy}
\footnotesize
\centering
\begin{tabular}{ccccccc}
\toprule
\textbf{}					& \textbf{}				& $\boldsymbol{n = 1}$ 		& $\boldsymbol{n = 2}$ 		& $\boldsymbol{n = 3}$ 		& $\boldsymbol{n = 4}$ 		& $\boldsymbol{n = 5}$ \\
\midrule
\multirow{4}{*}{DTU}		& L1-abs				& 26.394 					& 26.221 					& \underline{23.473}		& 25.045  						& 29.676 \\
							& \scriptsize{(in \mm)} & \scriptsize{$\pm$24.262}	& \scriptsize{$\pm$23.835}	& \scriptsize{$\pm$19.656} 	& \scriptsize{$\pm$19.298} & \scriptsize{$\pm$19.436} \\
							& L1-rel				& 0.036						& 0.036					& \underline{0.032}			& 0.034							& 0.041 \\
							&						& \scriptsize{$\pm$0.032}	& \scriptsize{$\pm$0.032}	& \scriptsize{$\pm$0.026}	& \scriptsize{$\pm$0.026}	& \scriptsize{$\pm$0.026} \\
\midrule
\multirow{4}{*}{3DOMcity}	& L1-abs				& \underline{12.789}		& 14.936					& 21.801					& 32.458					& 47.422 \\
							&\scriptsize{(in \mm)}	& \scriptsize{$\pm$6.916}	& \scriptsize{$\pm$6.754}	& \scriptsize{$\pm$8.010}	& \scriptsize{$\pm$9.408}	& \scriptsize{$\pm$22.292} \\
							& L1-rel				& \underline{0.010}			& 0.012					& 0.017					& 0.026					& 0.037 \\
							& 						& \scriptsize{$\pm$0.006}	& \scriptsize{$\pm$0.006}	& \scriptsize{$\pm$0.007}	& \scriptsize{$\pm$0.009}	& \scriptsize{$\pm$0.014} \\
\bottomrule
\end{tabular}
\end{table}

\Cref{tab:pyrLvls-accuracy} lists the mean errors of the estimated depth maps when evaluated with different number of pyramid levels on the datasets of both the DTU and 3DOMcity benchmark. %
In this, the absolute and relative L1 measures are used, averaged over all depth maps within each dataset. %
It is to be expected, that the omission of any hierarchical processing, \ie the use of only one pyramid level and thus no coarse-to-fine processing, would lead to the smallest error between the estimate and ground truth. %
However, the results reveal that in case of the DTU dataset the smallest mean error, even if it is only slightly smaller, is achieved when setting $n = 3$, while the best result in case of the 3DOMcity dataset is achieved at $n = 1$. %

As described in \Cref{sec:methodology_planesweep_crossratio}, the plane distances within the plane-sweep sampling, and thus the sampling points, are selected in such a way that two consecutive planes induce a maximum disparity difference of $1$ pixel. %
Depending on the capturing setup, \ie the relative poses between the images and their obliqueness and, in turn, the range of the scene depth, this can lead to a very high number of sampling points and with it to a large memory consumption, as the dimensions of the three-dimensional cost volume need to be set accordingly. %
Thus, in order to not exceed the memory limit, the maximum number of sampling points for the highest pyramid level is restricted to $256$ in the implementation of the approach. %
In case of the camera setup of the DTU dataset and the configuration of this experiment, \ie having a bundle size of $|\Omega| = 3$, a pyramid height of $3$ is the smallest height at which the number of sampling points at the highest level does not reach or exceed the set limit, as \Cref{tab:pyrLvls-runtime} shows. %
Comparing \Cref{tab:pyrLvls-accuracy} and \Cref{tab:pyrLvls-runtime} further reveals that on both datasets the best results are achieved, when the computation is initialized at the highest pyramid level with a maximum of $128$ sampling planes.

\begin{table}[t] 
\caption{Processing run-time measured for different configurations of the pyramid height on the DTU and 3DOMcity dataset. In addition, the maximum number of sampling planes with which the scene was sampled at the highest pyramid level is stated.}
\label{tab:pyrLvls-runtime}
\footnotesize
\centering
\begin{tabular}{ccccccc}
\toprule
 \textbf{} 				& \textbf{}				& $\boldsymbol{n = 1}$ 	& $\boldsymbol{n = 2}$ 	& $\boldsymbol{n = 3}$ 	& $\boldsymbol{n = 4}$ 		& $\boldsymbol{n = 5}$ \\
\midrule
 \multirow{3}{*}{DTU}	&  Run-time  				& 2365 					& 1315 					& 386					& 220  				& 187 \\
 						& \scriptsize{(in \ms)} 	& \scriptsize{$\pm$15}	& \scriptsize{$\pm$10}	& \scriptsize{$\pm$2} 	& \scriptsize{$\pm$2}	& \scriptsize{$\pm$1} \\
 						& max. \# planes			& 256					& 256					& 128					& 64							& 32 \\
\midrule
 \multirow{3}{*}{3DOMcity}	&  Run-time  				& 613 					& 431 					& 225					& 196  				& 192 \\
 							& \scriptsize{(in \ms)} 	& \scriptsize{$\pm$3}	& \scriptsize{$\pm$3}	& \scriptsize{$\pm$1} 	& \scriptsize{$\pm$1}	& \scriptsize{$\pm$1} \\
 							& max. \# planes			& 128					& 64						& 32					& 16							& 8 \\
\bottomrule
\end{tabular}
\end{table}

Another criterion which is used to deduce the best configuration on the height of the Gaussian pyramid is the run-time needed to estimate a single depth map. %
\Cref{tab:pyrLvls-runtime} additionally lists the corresponding measurements taken, \ie the number of milliseconds it takes to estimate a single depth map given a certain number of pyramid levels, as well as the number of planes used for sampling the scene space at the highest pyramid level. %
The measurements again show, that up to $n = 3$ in case of the DTU dataset, the number of sampling planes at the highest pyramid level is equal to the limit of 256 and that with a smaller amount of sampling points the run-time decreased drastically. %
Furthermore, the significant drop of one second in run-time between using a pyramid height of 2 and 3 suggests that the decreasing use of processing resources on the GPU increases the processing speed and that going from $n = 2$ to $n = 3$ makes a significant improvement in its efficiency. %
Since the use of a higher number of pyramid levels does not only reduce the amount of sampling points, but also the image size at the highest pyramid level and with it the amount of pixels that need to be matched. %
Thus, depending on the camera setup, a hierarchical processing is very important in order to ensure a high sampling density of the scene space, while at the same time efficiently utilizing the processing hardware and, in turn, alleviating high processing speeds. %
In case of the DTU dataset, this experiment shows that the best number of pyramid levels to be used is $n = 3$, which will thus be set for the successive experiments. %
In case of the 3DOMcity dataset, \Cref{tab:pyrLvls-accuracy} suggests that the best configuration is to use the original image size. %
A hierarchical processing scheme is needed, however, in order to use \snSGM, the extension of the \gls*{SGM} algorithm to consider local surface orientations in order to account for slanted surfaces. %
Thus, in case of the 3DOMcity dataset, the successive experiments will be executed with $n = 2$, which induces only a slightly higher mean error compared to the best configuration. % 

\begin{table}[t!] 
\caption{Mean errors achieved on the DTU and $3$DOMcity dataset for different input bundle sizes~$|\Omega|$, \ie number of images. In addition, the differences in run-time, with respect to the measurements of the first part (\ie $|\Omega|=3$), are stated.}
\label{tab:bundlesize}
\footnotesize
\centering
\begin{tabular}{ccccc}
\toprule
 \textbf{} 					& \textbf{}				& $\boldsymbol{|\Omega| = 3}$ 	& $\boldsymbol{|\Omega| = 5}$ 	& $\boldsymbol{|\Omega| = 7}$  	\\
\midrule
 \multirow{6}{*}{DTU}		& \LOneAbs				& 23.473						& \underline{19.832}			& 21.843						\\
 							& \scriptsize{(in \mm)}	& \scriptsize{$\pm$19.656}		& \scriptsize{$\pm$16.225}		& \scriptsize{$\pm$21.605} 		\\
 							& \LOneRel				& 0.032							& \underline{0.027}				& 0.031							\\
 							& \scriptsize{(in \mm)}	& \scriptsize{$\pm$0.026}		& \scriptsize{$\pm$0.021}		& \scriptsize{$\pm$0.031}	    \\
 							& $\Delta$\,Run-time		& 								& +271 							& +302							\\
 							& \scriptsize{(in \ms)}	& 								& 								& 								\\
\midrule
 \multirow{6}{*}{3DOMcity}	& \LOneAbs				& 14.936						& \underline{14.615}			& 16.514						\\
 							& \scriptsize{(in \mm)}	& \scriptsize{$\pm$6.754}		& \scriptsize{$\pm$6.254}		& \scriptsize{$\pm$7.569}		\\
 							& \LOneRel				& \underline{0.012}				& 0.012							& 0.014							\\
 							& \scriptsize{(in \mm)}	& \scriptsize{$\pm$0.006}		& \scriptsize{$\pm$0.007}		& \scriptsize{$\pm$0.009}		\\
 							& $\Delta$\,Run-time 	&   							& +360 							& +410							\\
 							& \scriptsize{(in \ms)} & 								& 								& 					 			\\
 							
\bottomrule
\end{tabular}
\end{table}

In the second part of this experiment, the effects of a different number of input images and, in turn, the optimal size $|\Omega|$ of the input bundle are evaluated. %
Here, the settings for the plane-sweep image matching and the subsequent \gls*{SGM} optimization are kept the same as before. %
The height of the Gaussian pyramids is fixed to $n = 3$ in case of the DTU dataset and $n = 2$ in case of the data from 3DOMcity dataset.
\Cref{tab:bundlesize} lists the mean errors achieved on both datasets with different number of input images, as well as the difference in run-time with respect to the best configuration of the first part of the experiment. %
The results reveal that the best accuracies are achieved, when five input images are used for image matching, even though, in case of the 3DOMcity dataset, it is only a marginal improvement. %
As expected, the utilization of more input images in the process of image matching also leads to an increase in run-time, since more pixels are matched. %
At the same time, however, there is more time available to keep up with the image acquisition as discussed in \Cref{sec:discussion_runtime}. %
In conclusion, in the subsequent experiments, the size of the input bundle is set to $|\Omega| = 5$, while the height of the Gaussian image pyramids is set to $n = 3$ and $n = 2$ in case of the DTU and 3DOMcity dataset, respectively. 

\subsection{Effects of Different Similarity Measures in the Process of Dense Multi-Image Matching}
\label{sec:experiments_sim-measure}
\glsreset{CT}
\glsreset{NCC}

As part of the plane-sweep multi-image matching, this approach comprises two different similarity measures and cost functions: The Hamming distance of the \gls*{CT} as well as a truncated and scaled form of the \gls*{NCC}. %
While the \gls*{CT} is computationally less expensive than the \gls*{NCC} and is thus more suitable for real-time or online processing, it is less discriminative, which might result in a more ambiguous set of matched pixel correspondences. %
When working with a stereo normal case, in which the input images suffer only from little perspective distortion due to homographic transformations, the \gls*{CT} outperforms the \gls*{NCC} in both run-time and accuracy \citep{Ruf2021restac}.
However, as the results in \Cref{tab:simmeasure} show, the perspective distortion, resulting from the warping of images from converging cameras by means of the plane-induced homography within the plane-sweep algorithm, leads to a significant increase in error when using the \gls*{CT} as similarity measure instead of the \gls*{NCC}. %

\begin{table}[!b] 
\caption{Mean errors achieved on the DTU and 3DOMcity dataset when using different similarity measures and cost functions with different support regions.}
\label{tab:simmeasure}
\footnotesize
\centering
\begin{tabular}{cccccc}
\toprule
 \textbf{} 					& \textbf{}				& $\boldsymbol{\text{CT}_{5\times5}}$	& $\boldsymbol{\text{CT}_{9\times7}}$	& $\boldsymbol{\text{NCC}_{5\times5}}$	& $\boldsymbol{\text{NCC}_{9\times9}}$ 	\\
\midrule
 \multirow{4}{*}{DTU}		& \LOneAbs				& 42.136								& 42.305								& 19.832												& \underline{19.667} \\
 							& \scriptsize{(in \mm)}	& \scriptsize{$\pm$37.958}				& \scriptsize{$\pm$36.394}				& \scriptsize{$\pm$16.225} 							& \scriptsize{$\pm$16.453} \\
 							& \LOneRel				& 0.056									& 0.057									& 0.027													& \underline{0.027} \\
 							& \scriptsize{(in \mm)}	& \scriptsize{$\pm$0.048}				& \scriptsize{$\pm$0.046}				& \scriptsize{$\pm$0.021}	    						& \scriptsize{$\pm$0.021} \\
\midrule
 \multirow{4}{*}{3DOMcity}	& \LOneAbs				& 22.128								& 26.005								& 14.615												& \underline{13.789} \\
 							& \scriptsize{(in \mm)}	& \scriptsize{$\pm$14.218}				& \scriptsize{$\pm$14.106}				& \scriptsize{$\pm$6.254}								& \scriptsize{$\pm$5.962} \\
 							& \LOneRel				& 0.019									& 0.022									& 0.012													& \underline{0.011} \\
 							& \scriptsize{(in \mm)}	& \scriptsize{$\pm$0.014}				& \scriptsize{$\pm$0.014}				& \scriptsize{$\pm$0.007}								& \scriptsize{$\pm$0.006} \\
 							
\bottomrule
\end{tabular}
\end{table}

Apart from the two different similarity measures, the effects of different support regions are also evaluated in the scope of this experiment. %
In this, for each similarity measure, the two most commonly used configurations were tested, with a support region of a size of $5\times5$\px being a good trade-off between uniqueness and computational complexity, while, in case of the \gls*{CT}, a support region of a size of $9\times7$\px is the biggest size for which the bit-string still fits into a single $64$-bit integer. %
The configuration of the plane-sweep algorithm and the \gls*{SGM} optimization is set in accordance with the values from the first experiment (\cf \Cref{sec:experiments_plane-sweep}). %
In terms of the \gls*{SGM} penalties, $\varphi_1$ is set to $100$ for both $\text{NCC}_{5\times5}$ and $\text{NCC}_{9\times9}$, since the maximum matching cost of the \gls*{NCC} is normalized to $255$, independent of the support region. %
For $\text{CT}_{5\times5}$ and $\text{CT}_{9\times7}$, however, $\varphi_1$ is set to $9$ and $24$, respectively, which is equivalent to the configuration for \gls*{NCC}, when considering the ratio between $\varphi_1$ and the maximum matching cost. %

Even though the \gls*{NCC} with a support region of $9\times9$\px achieves the best results, $\text{NCC}_{5\times5}$ is selected for further experiments, since the rise in error is only little but its computational complexity is less and, in turn, its throughput higher than that of $\text{NCC}_{9\times9}$ as measured by \citet{Ruf2021restac}. %

\subsection{Ability to Reconstruct Non-Fronto-Parallel Surface Structures}
\label{sec:experiments_surface-aware}

\begin{table}[!b] 
\caption{Quantitative comparison of the results achieved by different implementations and adaptations of the \gls*{SGM} algorithm in combination with a fronto-parallel sweeping direction in order to also account for non-fronto-parallel surfaces.}
\label{tab:surface-aware_sgmx}
\footnotesize
\centering
\begin{tabular}{ccccc}
\toprule
 \textbf{} 					& \textbf{}				& \textbf{\piSGM}				& \textbf{\snSGM}					& \textbf{\pgSGM	}		 		\\
\midrule
 \multirow{4}{*}{DTU}		& L1-abs					& 19.832							& 19.768								& \underline{19.684}			 	\\
 							& \scriptsize{(in \mm)}	& \scriptsize{$\pm$16.225}		& \scriptsize{$\pm$16.192}			& \scriptsize{$\pm$16.154}	 	\\
 							& L1-rel					& 0.027							& 0.027								& \underline{0.027}				\\
 							& \scriptsize{(in \mm)}	& \scriptsize{$\pm$0.021}		& \scriptsize{$\pm$0.021}			& \scriptsize{$\pm$0.021}	    	\\
\midrule
 \multirow{4}{*}{3DOMCity}	& L1-abs					& \underline{14.615}				& 14.673								& 15.074							\\
 							& \scriptsize{(in \mm)}	& \scriptsize{$\pm$6.254}		& \scriptsize{$\pm$6.229}			& \scriptsize{$\pm$6.133}		\\
 							& L1-rel					& \underline{0.012}				& 0.012								& 0.012							\\
 							& \scriptsize{(in \mm)}	& \scriptsize{$\pm$0.007}		& \scriptsize{$\pm$0.007}			& \scriptsize{$\pm$0.006}    	\\
 							
\bottomrule
\end{tabular}
\end{table}

As described in \Cref{sec:methodology_sgm_x}, apart from the straight-forward combination of \gls*{SGM} with the plane-sweep sampling (\piSGM), this work comprises two further extensions of the \gls*{SGM} algorithm that allow to account for non-fronto-parallel surface structures: namely the incorporation of surface normals to adjust the zero-cost transition in the \gls*{SGM} path aggregation (\snSGM) and the penalization of deviations from the gradient of the minimum cost path (\pgSGM). %
In addition, by selecting an appropriate normal vector, and with it a corresponding sweeping direction, it is also possible to adjust the sampling plane orientations of the plane-sweep sampling to the scene structures. %
In the following, the results achieved by \snSGM and \pgSGM, in combination with a fronto-parallel sampling plane orientation, are first evaluated and compared to those achieved by \piSGM, before the effects of different non-fronto-parallel plane orientations are examined. %
The configuration of the other hyper-parameters is as described and evaluated above: the size of the input bundle is set to $|\Omega| = 5$, the \gls*{NCC} with a support region of $5\times 5$\px is used. %
The pyramid height is set to $n = 3$ and $n = 2$ for the DTU and 3DOMcity dataset, respectively. %

\begin{figure*}[p]
     \centering
     \def\svgwidth{\textwidth}
     \subimport{figures/}{sgmx_qual_comparison_DTU-edited.pdf_tex}
     \caption{Qualitative comparison of the results achieved by the three different SGM implementations on the DTU dataset. %
     \textbf{Row~1:} Reference data from the dataset, \ie the ground truth depth and normal map, as well as the reference image for which the data us computed. %
     \textbf{Rows~2~-~4:} Data, \ie depth, normal and confidence maps, computed by \piSGM, \snSGM and \pgSGM, respectively. %
     As well as difference maps holding the pixel-wise absolute difference between the estimated depth map and the ground truth. The color encoding reaches from dark blue (low error) via green to yellow (high error). %
     The estimated maps are masked according to the ground truth. %
     }
     \label{fig:sgmx_DTU_result}
\end{figure*}
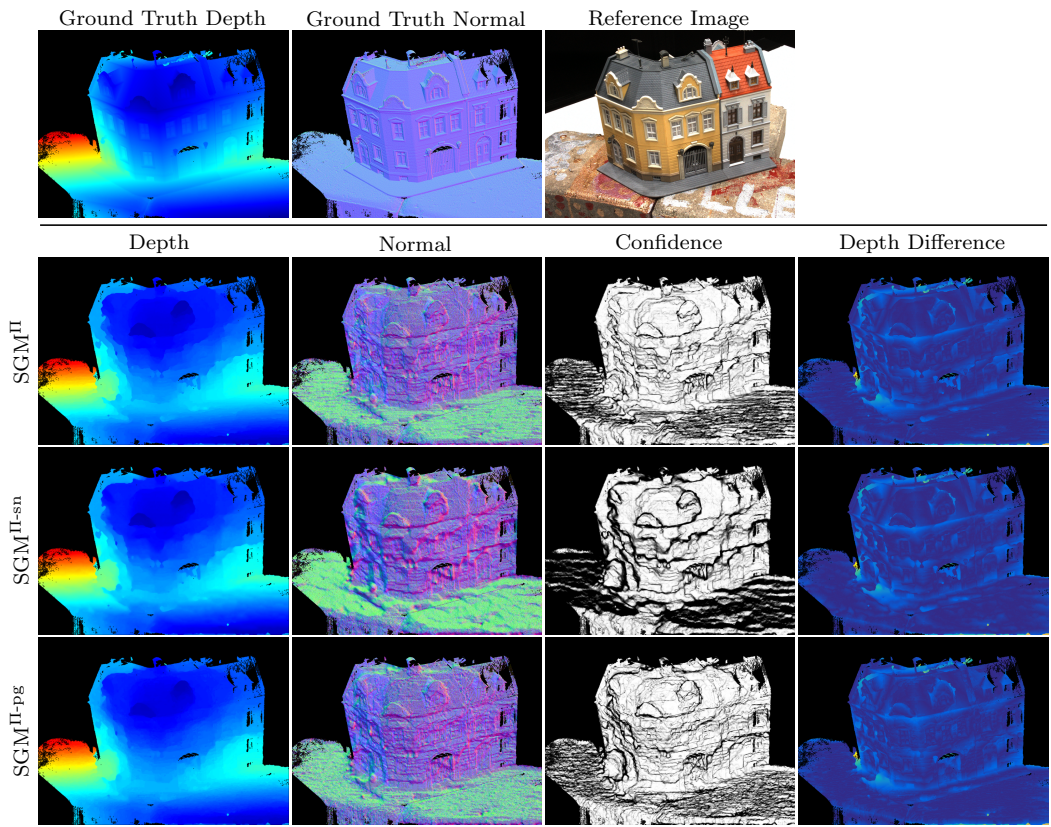

\begin{figure*}[p]
     \centering
     \def\svgwidth{\textwidth}
     \subimport{figures/}{sgmx_qual_comparison_3DOMcity-edited.pdf_tex}
     \caption{Qualitative comparison of the results achieved by the three different SGM implementations on the 3DOMcity dataset. %
     \textbf{Row~1:} Reference data from the dataset, \ie the ground truth depth and normal map, as well as the reference image for which the data us computed. %
     \textbf{Rows~2~-~4:} Data, \ie depth, normal and confidence maps, computed by \piSGM, \snSGM and \pgSGM, respectively. %
     As well as difference maps holding the pixel-wise absolute difference between the estimated depth map and the ground truth. %The color encoding reaches from dark blue (low error) via green to yellow (high error). %
     The estimated maps are masked according to the ground truth. %
     For visualization in this figure, the resulting images have been rotated counterclockwise by $90^{\circ}$. %
     Thus, the color encoding of the normal maps differs from that used in the other figures. %
     Here, red represents an upwards orientation, while green represents an orientation to the left. %
     }
     \label{fig:sgmx_3DOMcity_result}
\end{figure*}
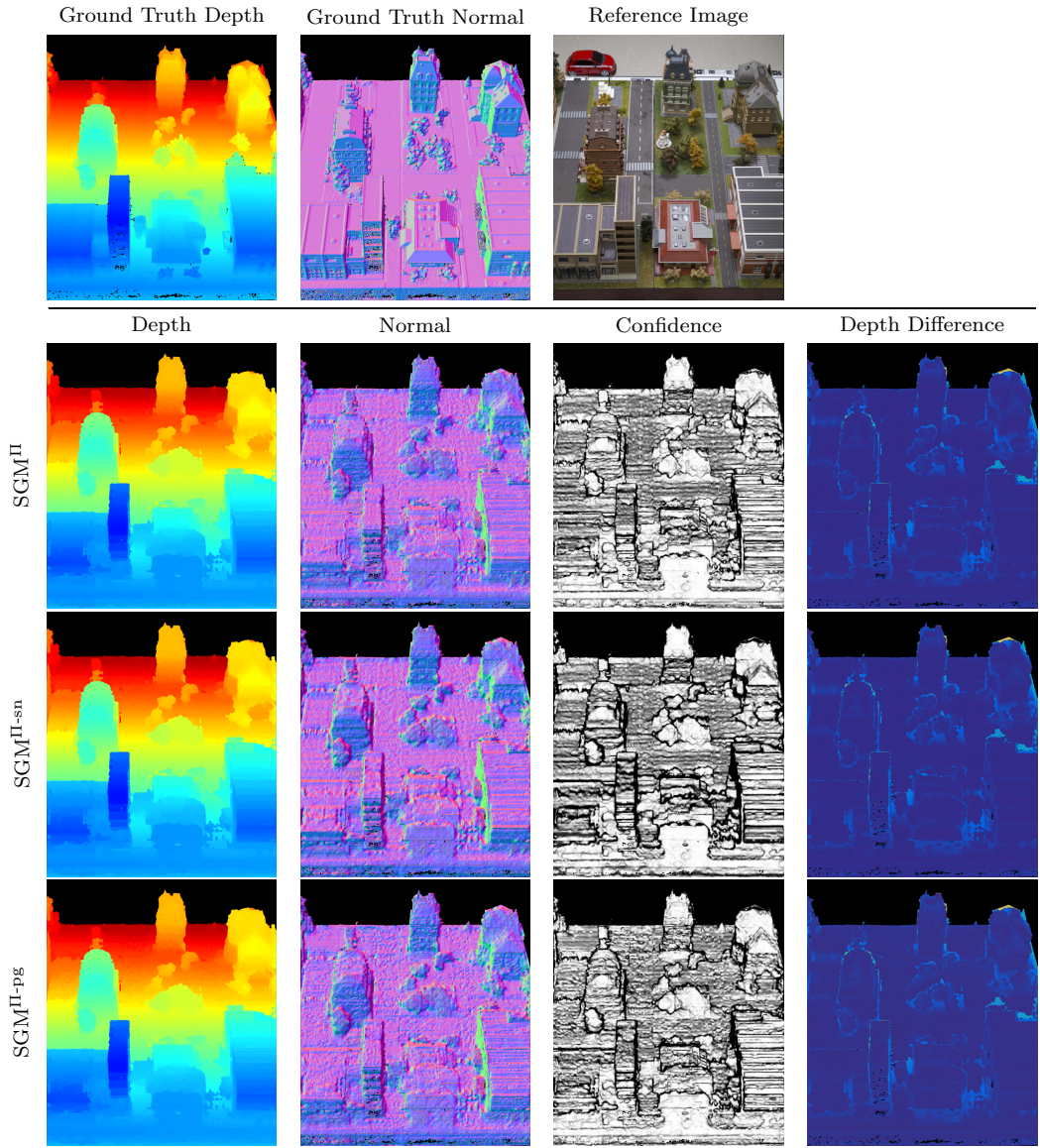

The quantitative results displayed in \Cref{tab:surface-aware_sgmx} only reveal minor differences in the \LOne error between the different implementations of the \gls*{SGM} optimization. %
While, in case of the DTU dataset, the best results are achieved by the \pgSGM implementation, on the 3DOMcity dataset, the standard adaptation of the \gls*{SGM} optimization to the plane-sweep sampling, \ie \piSGM, reaches the lowest error. %
The relative \LOne error does not reveal any difference. %
This is due to the fact that the individual \LOneRel scores only start to differ from the forth decimal place on-wards. %
Nonetheless, the ranking with respect to \LOneRel score corresponds to that of the \LOneAbs score. %
In a qualitative comparison, \Cref{fig:sgmx_DTU_result} reveals that \snSGM leads to a seemingly smoother depth and normal map (\eg on the ground plane), while at the same time, however, loosing small details and amplifying unwanted depth discontinuities in some areas, such as the building facade. %
When closely comparing the normal maps between \piSGM and \pgSGM, slightly smaller stair-casing artifacts can be noticed in case of \pgSGM, which also supports the slightly lower error in \Cref{tab:surface-aware_sgmx}. %
A qualitative comparison between the results on the 3DOMcity dataset in \Cref{fig:sgmx_3DOMcity_result}, however, does not reveal any noticeable differences between the different implementations. %
The reason for the small \LOneAbs error achieved by \piSGM on the 3DOMcity dataset is assumed to be due to the fact, that the 3DOMcity dataset also contains a subset of nadir images, in which only a small number of slanted surfaces are existing and the fronto-parallel orientation of the sampling planes coincides with most of the scene structure. %
\clearpage

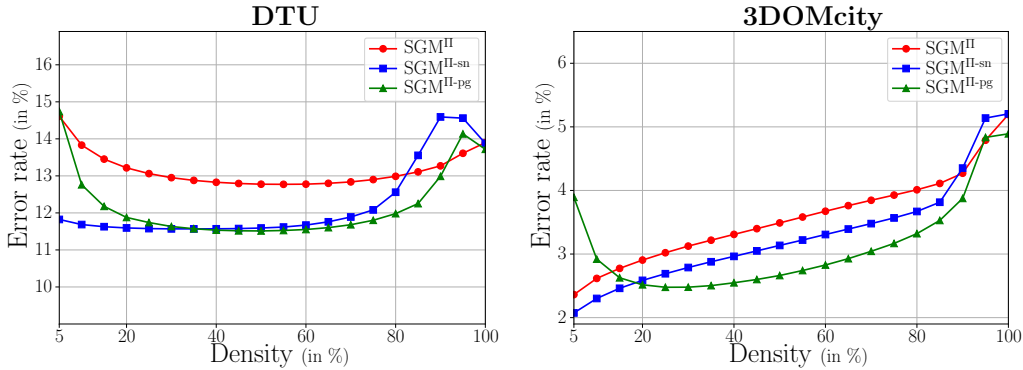
\begin{figure}[t!]%
	\centering%
	\resizebox{\columnwidth}{!}{\subimport{figures/}{ROC_ACC_Density_combined-edited.pdf_tex}} %
	\caption{ 
		ROC curves illustrating the error rate achieved by the three different \gls*{SGM} implementations as a function of increasing density of the estimated depth map.
	}%
	\label{fig:sgmx_roc}%
\end{figure}

To further quantify the strength and weaknesses of the three different \gls*{SGM} aggregation strategies, three \gls*{ROC} curves, one for each extension, are plotted for each dataset in \Cref{fig:sgmx_roc}. %
In this, the curves illustrate the error rate achieved by the corresponding \gls*{SGM} extension as a function of increasing density of the estimated depth map. %
The density of the depth map is varied by sampling the number of pixels in steps of $5$\percent based on their ordered confidence stored in $\mathcal{C}$, going from a high to a low confident estimate. %
The mean error rate is quantified by $1 \text{-} \text{Acc}_{1.05}$ (\cf \Cref{eq:acc-measure}) and states the number of sampled pixels in $\mathcal{D}$ whose absolute difference to the ground truth exceeds $5$\percent of the ground truth value. % 
Thus, at a low density of $\mathcal{D}$, \ie a high confidence threshold, the error rate should ideally be at its minimum and then increase with increasing density, reaching the overall error of $\mathcal{D}$ at a density of $100$\percent. %
The plots start off at a density of $5$\percent, since the error rate at a density of $0$\percent is undefined.
The curves in \Cref{fig:sgmx_roc} further support the superiority of the surface-aware \gls*{SGM} extensions over the standard \gls*{SGM} adaptation to plane-sweep sampling, as for both datasets the curves of \snSGM and \pgSGM are below that of \piSGM, illustrating smaller error rates. %
The fact that most of the \gls*{ROC} curves start of on a high error rate at a density of $5$\percent and then drop down before increasing again, suggests that the estimated confidence values do not represent the certainty of the depth estimates appropriately. %
The reasons for this are manifold and this is further discussed in \Cref{sec:discussion_postFilteringConfidence}. %

\begin{table}[t!] 
\caption{Run-time and error measurements of the three \gls*{SGM} extensions with $8$ and $4$ aggregation paths each, conducted on the DTU dataset with an image size of $1600\times1200$\px.}
\label{tab:surface-aware_sgmx_runtime}
\footnotesize
\centering
\begin{tabular}{ccccc}
\toprule
 \textbf{} 								& \textbf{}				& \textbf{\piSGM}			& \textbf{\snSGM}				& \textbf{\pgSGM}	\\
\midrule
 \multirow{6}{*}{$8$-Path \gls*{SGM}}		& \LOneAbs				& 19.832					& 19.768						& \underline{19.684}		 	\\
 										& \scriptsize{(in \mm)}	& \scriptsize{$\pm$16.225}	& \scriptsize{$\pm$16.192}		& \scriptsize{$\pm$16.154}	 	\\
 										& \LOneRel				& 0.027						& 0.027							& \underline{0.027}				\\
 										& \scriptsize{(in \mm)}	& \scriptsize{$\pm$0.021}	& \scriptsize{$\pm$0.021}		& \scriptsize{$\pm$0.021}	    	\\
 										& Run-time				& \underline{640}						& 895							& 2079 				\\
 										& \scriptsize{(in \ms)}	& \scriptsize{$\pm$3}		& \scriptsize{$\pm$2}			& \scriptsize{$\pm$3}	    	\\
\midrule
 \multirow{6}{*}{$4$-Path \gls*{SGM}}		& \LOneAbs				& 21.072					& 21.091						& \underline{20.908}		 	\\
 										& \scriptsize{(in \mm)}	& \scriptsize{$\pm$16.634}	& \scriptsize{$\pm$16.632}		& \scriptsize{$\pm$16.599}	 	\\
 										& \LOneRel				& 0.029						& 0.029							& \underline{0.029}				\\
 										& \scriptsize{(in \mm)}	& \scriptsize{$\pm$0.022}	& \scriptsize{$\pm$0.022}		& \scriptsize{$\pm$0.022}	    	\\
 										& Run-time				& \underline{413}						& 546							& 1132 				\\
 										& \scriptsize{(in \ms)}	& \scriptsize{$\pm$2}		& \scriptsize{$\pm$2}			& \scriptsize{$\pm$3}	    	\\							
\bottomrule
\end{tabular}
\end{table}

Since the presented approach aims for incremental and online processing, meaning that the computation should ideally keep up with the input stream, the run-time of each \gls*{SGM} extension is also of interest. %
In \Cref{tab:surface-aware_sgmx_runtime}, the run-time and error of the complete approach, with above mentioned parameterization and with respect to each of the three \gls*{SGM} implementations is listed. %
The measurements were conducted on the dataset of the DTU benchmark. %
In addition to the standard use of $8$ aggregation paths, which achieves the lowest error, the run-time and error when using only $4$ aggregation paths is also listed. %
In the latter case, the diagonal aggregation paths are omitted within the \gls*{SGM} aggregation.
This is motivated by the fact that a number of studies \citep{Banz2010realtime, Hernandez2016embedded, Ruf2021restac} show that a reduction in the number of aggregation paths from $8$ to $4$ greatly decreases the computation time of the \gls*{SGM} aggregation, while only marginally increasing its error, which is also supported by the numbers in \Cref{tab:surface-aware_sgmx_runtime}.
Furthermore, the measurements reveal that especially the \pgSGM extension introduced a great computational complexity compared to \piSGM and \snSGM. %
However, the reduction of the aggregation paths has a great impact on the run-time, reducing it by up to $45$\percent, but only marginally affecting the error. %
Whether the listed run-time is sufficient for online processing is further discussed in \Cref{sec:discussion_runtime}. %

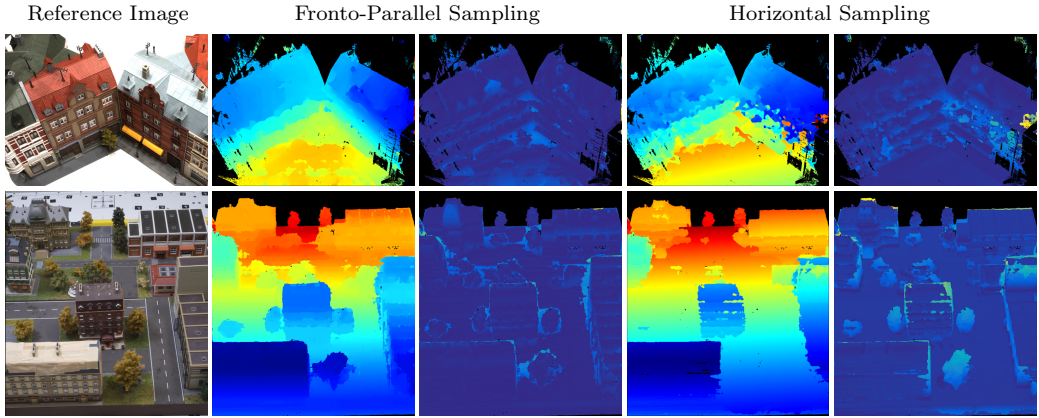
\begin{figure*}[t!]
     \centering
     \def\svgwidth{\textwidth}
     \subimport{figures/}{diffPlane_qual_comparison-edited.pdf_tex}
     \caption{Qualitative comparison between the use of a fronto-parallel and non-fronto-parallel sampling direction in combination with \piSGM. %
     \textbf{Columns 2 \& 4:} Corresponding estimated depth map. %
     \textbf{Columns 3 \& 5:} Difference map holding the pixel-wise absolute difference between the estimated depth map and the ground truth. The color encoding reaches from dark blue (low error) via green to yellow (high error). %
     The estimated depth maps and the difference maps are masked according to the ground truth. %
     }
     \label{fig:diffPlane_result}
\end{figure*}

In the second part of this experiment, the use of non-fronto-parallel plane orientations within the plane-sweep sampling is investigated. %
In this, an additional horizontal and vertical orientation, both with respect to the reference coordinate system of the scene, were selected and compared to the fronto-parallel sampling direction. %
For this, the DTU dataset is split into two subsets, one for the horizontal sampling, in which the camera is looking in a more downwards direction, as well as one for the vertical sampling, where the camera pitch is smaller. %
As reference, for both subsets, results with a fronto-parallel sampling were computed separately. %
The quantitative results reveal a major increase in error when non-fronto-parallel plane orientations are used for sampling, as also illustrated by the excerpt in \Cref{fig:diffPlane_result}. %
However, the \Cref{fig:diffPlane_result} also reveals that in areas where the surface structure coincides with the sampling direction, \eg the ground plane, the depth map is very smooth and consistent. %

\subsection{Improvements Gained by Post-Filtering and Geometric Consistency}
\label{sec:experiments_post-filtering}
\glsreset{DoG}

In a final ablative experiment, the effects of the implemented post-filtering methods, to remove remaining outliers and supposedly wrong estimates by means of \gls*{DoG} filtering (\cf \Cref{sec:methodology_post_dog}) and enforcing geometric consistency (\cf \Cref{sec:methodology_post_geoConsistency}) are studied. %
While the latter one relies on the actual estimates, the \gls*{DoG} filter is based on the assumption that image regions with low texture might lead to ambiguities in the image matching and, in turn, wrong estimates, masking out corresponding regions. %
This, however, might result in the false removal of good or even correct estimates. %

\begin{figure*}[t!]%
	\centering%	
	\resizebox{\textwidth}{!}{\subimport{figures/}{postfilteringAccCpl_combined-edited.pdf_tex}} %
	\caption{ 
		Accuracy-completeness curves of different post-filtering strategies, \ie \gls*{DoG} filtering, geometric consistency filtering, as well as a combination of both, executed in combination with the three different \gls*{SGM} extensions and a fronto-parallel sampling. %
		In the construction, the threshold $\theta$ is varied within the list of $1.25, 1.20, 1.15, 1.10, 1.05\ \text{and}\ 1.01$. %
		With a decreasing threshold, the accuracy and completeness rates drop. %
	}%
	\label{fig:postFiltering_AccCpl}%
\end{figure*}
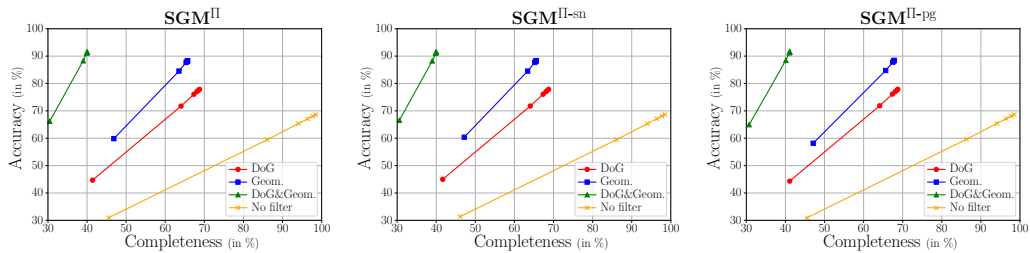

Instead of using the absolute and relative \LOne metric to quantitatively assess the results achieved when employing post-filtering, the effects are evaluated using the accuracy $\text{Acc}_{\theta}$ (\cf \Cref{eq:acc-measure}) and completeness $\text{Cpl}_{\theta}$ measure (\cf \Cref{eq:cpl-measure}). %
This is because they indirectly include information on the density of the resulting depth maps, which should ideally be as high as possible. %
Since the individual sequences of the 3DOMcity dataset are made up of too little images in order to perform a geometric consistency check with the parameterization mentioned in \Cref{sec:methodology_post_geoConsistency}, this experiment is only conducted on the dataset of the DTU benchmark. %
\Cref{fig:postFiltering_AccCpl} depicts the results of different post-filtering strategies, \ie \gls*{DoG} filtering, geometric consistency filtering, as well as a combination of both, executed in combination with the three different \gls*{SGM} extensions and a fronto-parallel sampling. %
In addition, the accuracy-completeness curve resulting from the corresponding configurations without post-filtering is also displayed as reference. %
In the construction of the curves, the threshold $\theta$ (\ie the threshold used to calculate the accuracy and completeness rates) is varied within the list of $1.25, 1.20, 1.15, 1.10, 1.05\ \text{and}\ 1.01$. %
Note that with a decreasing threshold, the accuracy and completeness rates drop. %
Thus, the highest values are achieved with $\theta = 1.25$.

Most evidently, \Cref{fig:postFiltering_AccCpl} again reveals that there is not much difference in the overall error between the three \gls*{SGM} implementations. %
However, the accuracy-completeness curves clearly show the differences that the individual post-filtering strategies make. %
Unsurprisingly, the reference configuration where no filtering is employed reaches the highest completeness, as no estimates are removed from the predicted depth map, which, in turn, also leads to the lowest accuracy. %
The use of a \gls*{DoG} filtering clearly improves this, as it is likely to remove quite a number of wrong estimates originating from low-textured areas. %
However, as expected, the \gls*{DoG} filter presumably also removes a number of correct estimates, as the use of the filtering based on geometric consistency achieves similar completeness, while at the same time reaching a higher accuracy. %
Especially when considering the scores for $\theta = 1.01$, \ie the lower left end of each curve, the use of a geometric consistency filter achieves an increase in completeness of approximately $5$\percent, while exceeding the accuracy of the use of a \gls*{DoG} filter by more than $10$\percent. %
A distinct recommendation on which filter to use, however, cannot be concluded, since both filtering strategies have their strengths and weaknesses, especially with respect to online processing as discussed in \Cref{sec:discussion_runtime}. %
A combination of both filters is not motivated. %
Even though the accuracy slightly increases, the completeness drops partly by over $20$\percent. %
Moreover, this effect can also be achieved by decreasing the threshold of the reprojection error $\eta_{\mathrm{r}}$ in the geometric consistency check, which probably will increase the accuracy even more. %

Lastly, to directly compare the different \gls*{SGM} extensions in combination with the geometry-based filtering, which achieves the best results, the corresponding $\text{F}$-scores for each evaluated $\theta$ (\cf \Cref{eq:fscore-measure}) are listed in \Cref{tab:postFiltering_Fscore}. %
Just as the results displayed in \Cref{tab:surface-aware_sgmx}, the $\text{F}$-scores reveal the superiority of \pgSGM over the other two implementations, since for all $\theta$, except one, \pgSGM reaches the highest $\text{F}$-score. %

\begin{table}[!t] 
\caption{$\text{F}$-scores achieved by the three \gls*{SGM} extensions in combination with the post-filtering based on geometric consistency.}
\label{tab:postFiltering_Fscore}
\footnotesize
\centering
\begin{tabular}{ccccccc}
\toprule
 \textbf{}	& $\boldsymbol{\text{\textbf{F}}_{1.25}}$	& $\boldsymbol{\text{\textbf{F}}_{1.20}}$	& $\boldsymbol{\text{\textbf{F}}_{1.15}}$ & $\boldsymbol{\text{\textbf{F}}_{1.10}}$	& $\boldsymbol{\text{\textbf{F}}_{1.05}}$	& $\boldsymbol{\text{\textbf{F}}_{1.01}}$	\\
 			& \scriptsize{(in \%)}						& \scriptsize{(in \%)}						& \scriptsize{(in \%)}					  & \scriptsize{(in \%)}					& \scriptsize{(in \%)}						& \scriptsize{(in \%)} \\						
\midrule	

\piSGM		& $74.2$									& $74.1$									& $74.0$	 							  & $73.7$					& $71.5$									& $51.9$ 							\\	

\snSGM		& $74.1$									& $74.1$									& $74.0$							 	  & $73.6$					& $71.5$									& \underline{$52.3$}					\\

\pgSGM		& \underline{$75.6$}						& \underline{$75.5$}						& \underline{$75.4$}				 		& \underline{$75.1$}				& \underline{$72.9$}					& $51.4$ 							\\
 							
\bottomrule
\end{tabular}
\end{table} 

\subsection{Summarized Results and Comparison to Offline Multi-View Stereo Approaches}
\label{sec:experiments_comparisonToColmap}

Finally, before presenting and qualitatively assessing the results of the best configuration on real-world data, a short quantitative summary and comparison to the results produced by offline \gls*{MVS} is done in the following. %
As offline \gls*{MVS} approach, the widely used and open source \COLMAP toolbox \citep{Schoenberger2016mvs} is used. %
While \COLMAP provides the full reconstruction pipeline, \ie including a subsequent fusion of depth maps into a $3$D model, only the geometric depth maps were used in order to make a fair comparison, since the fusion into a $3$D model leads to a further filtering of outliers. %
The significance of a comparison between an online \gls*{MVS} approach, like the one presented in this work, with an offline approach can be questioned, nonetheless, since the two types of approaches make different assumptions and focus on different aspects within the processing, as further discussed in \Cref{sec:discussion_overallAccuracy}.

\begin{table}[t!] 
\caption{
Final quantitative results of the three SGM extensions on the DTU dataset, in combination with fronto-parallel plane-sweep sampling and post-filtering based on geometric consistency check. %
As reference, the quantitative results achieved by the geometric depth maps of the offline \gls*{MVS} toolbox \COLMAP\citep{Schoenberger2016mvs} are given. 
}
\label{tab:comparisonToColmap}
\footnotesize
\centering
\resizebox{\textwidth}{!}{
\begin{tabular}{ccccccccc}
\toprule
 \textbf{}	& \textbf{L1-abs} 	&  \textbf{L1-rel} 		& $\boldsymbol{\text{\textbf{F}}_{1.25}}$	& $\boldsymbol{\text{\textbf{F}}_{1.20}}$	& $\boldsymbol{\text{\textbf{F}}_{1.15}}$ & $\boldsymbol{\text{\textbf{F}}_{1.10}}$	& $\boldsymbol{\text{\textbf{F}}_{1.05}}$	& $\boldsymbol{\text{\textbf{F}}_{1.01}}$	\\
 			& \scriptsize{(in mm)} & \scriptsize{(in mm)} & \scriptsize{(in \%)} & \scriptsize{(in \%)} & \scriptsize{(in \%)} & \scriptsize{(in \%)} & \scriptsize{(in \%)} & \scriptsize{(in \%)} \\
\midrule	

\piSGM		& $8.549$\,\scriptsize{$\pm7.509$} & $0.012$\,\scriptsize{$\pm0.011$} & $74.2$								& $74.1$								& $74.0$							 & $73.7$							& $71.5$								& $51.9$ 							\\	

\snSGM		& $8.479$\,\scriptsize{$\pm7.559$} & $0.012$\,\scriptsize{$\pm0.011$} & $74.1$								& $74.1$								& $74.0$							 & $73.6$							& $71.5$								& $52.3$								\\

\pgSGM		& $8.722$\,\scriptsize{$\pm7.255$} & $0.013$\,\scriptsize{$\pm0.010$} & $75.6$								& $75.5$								& $75.4$				 			 & $75.1$							& $72.9$								& $51.4$ 							\\

\COLMAP 	& $3.745$\,\scriptsize{$\pm5.498$} & $0.006$\,\scriptsize{$\pm0.004$} & $80.2$								& $80.2$								& $80.1$				 			 & $80.0$							& $79.6$								& $74.4$ 							\\
 							
\bottomrule
\end{tabular}
}
\end{table}

\begin{table}[!b] 
\caption{
Run-time comparison between the three SGM extensions with $8$ aggregation paths and \gls*{MVS} approach of \COLMAP\citep{Schoenberger2016mvs}. %
Measurements were conducted on the dataset of the DTU benchmark and represent the mean run-time required by the different approaches to estimate a single depth map. %
While \COLMAP allows to only estimate photometric depth maps, the quantitative evaluation with respect to the accuracy of \COLMAP was done on the geometric depth maps (last column). %
}
\label{tab:comparisonToColmapRuntime}
\footnotesize
\centering
\begin{tabular}{cccccc}
\toprule
 \textbf{}	& \textbf{\piSGM} 	&  \textbf{\snSGM}		& \textbf{\pgSGM}	& \textbf{\COLMAP}	& \textbf{\COLMAP} 	\\
			& \scriptsize{$8$-Path}	& \scriptsize{$8$-Path}	& \scriptsize{$8$-Path}	& \scriptsize{Photometric} 				& \scriptsize{Photometric+Geometric}  \\
\midrule	
Run-time   	& \underline{640} & 895	 & 2079 	& 14687										& 34840								 	\\	
\scriptsize{(in \ms)} & \scriptsize{$\pm$3}		& \scriptsize{$\pm$2}	& \scriptsize{$\pm$3}	& \scriptsize{$\pm$1020}										& \scriptsize{$\pm$1243}									 	\\			
\bottomrule
\end{tabular}
\end{table}

Based on the previous experiments, the following setup is selected for the final experiments on the DTU dataset. %
The size of the input bundle is set to $|\Omega| = 5$ and the height of the image pyramids for the hierarchical processing is set to $n = 3$. %
As sweeping direction and plane normal for the multi-image plane-sweep sampling, a fronto-parallel plane orientation, \ie $\mathrm{n} = (0\ 0\ \text{-}1)^{\intercal}$, is used. %
For the image matching, the \acrlong*{NCC} with a support region of $5\times 5$ pixels is selected and the penalty $\varphi_1$ for the subsequent \gls*{SGM} optimization is set to $\varphi_1 = 100$, with $\varphi_2$ being adaptively adjusted according to the intensity difference between neighboring pixels as described in \Cref{sec:adaptive_p2}. %
For post-filtering of the resulting maps, the geometric consistency check based on the reprojection error achieves the best results and is thus selected. %
Since, the geometric consistency check is not applicable for the dataset of the 3DOMcity benchmark, as the individual image sequences of the dataset are too short, the final comparison is only performed on the dataset provided by the DTU benchmark. %

\begin{table}[b!] 
\caption{
Quantitative results of related work on the DTU benchmark, using the actual evaluation protocol of the benchmark. %
This protocol calculates the errors not on the individual depth maps, but on the actual point cloud and specifies the accuracy (Acc) and completeness (Cpl) in absolute differences between the estimated points and the reference (lower is better). %
It is thus somehow comparable to the \LOneAbs error.
}
\label{tab:comparisonToOther}
\footnotesize
\centering
\begin{tabular}{cccc}
\toprule
 \textbf{}							& \textbf{Acc} 	&  \textbf{Cpl} 		& \textbf{Overall}	\\
 									& \scriptsize{(in mm)} & \scriptsize{(in mm)} & \scriptsize{(in mm)} \\
\midrule	
\COLMAP \citep{Schoenberger2016mvs} 	& $0.400$ 									& $0.664$ 									& $0.532$ \\
Furu \citep{Furukawa2010pmvs} 	& $0.613$ 									& $0.941$ 									& $0.777$ \\
Gipuma \citep{Galliani2015}			& $0.283$									& $0.873$										& $0.578$ \\
MVSNet \citep{Yao2018mvsnet}			& $0.396$									& $0.527$										& $0.462$ \\
 							
\bottomrule
\end{tabular}
\end{table}

\Cref{tab:comparisonToColmap} lists the quantitative results of the summarized experiments, together with the results achieved by the geometric depth maps of \COLMAP. %
As to be expected, the results of the offline \gls*{MVS} toolbox \COLMAP have the lowest error with respect to all accuracy measures, since it uses a more global optimization scheme without any run-time constraints and, in turn, taking significantly longer to estimate a single depth map as the numbers in \Cref{tab:comparisonToColmapRuntime} show. %
Moreover, in contrast to an online approach, offline processing allows to take all input images of the sequence into account and to choose the most appropriate ones for the tasks of \gls*{DIM} and \gls*{MVS}. %
Nonetheless, the results of the presented approach for online \gls*{MVS} are not too far off, with the error not even being one magnitude higher than that of \COLMAP. %
Interestingly, while \pgSGM outperforms the other two \gls*{SGM} extensions with respect to the $\text{F}$-score, \snSGM has the lowest \LOne error. %
This can be explained by the density of the resulting depth maps. %
When using \pgSGM, more estimates pass the geometric consistency check, resulting in depth maps that are slightly more dense than those produced by \piSGM and \snSGM and increasing the $\text{F}$-score, while at the same time also increasing the $\text{L1}$ error. %
Quantitatively speaking, the difference, however, is only marginal and a conclusion whether one certain \gls*{SGM} extension is to be preferred over the others depends on the use-case and is to be drawn based on qualitative comparisons. % 

In a final comparison, \Cref{tab:comparisonToOther} lists the quantitative results of some representative approaches from literature, including \COLMAP, on the evaluation set of the DTU dataset, calculated with the actual evaluation protocol of the benchmark. %
Here, in contrast to the accuracy (Acc) and completeness (Cpl) used in this work, the Acc and Cpl are calculated based on the absolute difference between the estimates and reference. %
And instead of using the individual depth maps, the evaluation protocol calculates the error based on the fused point clouds. %
Thus, the values are not directly comparable to the ones in \Cref{tab:comparisonToColmap}. %
However, in both tables the results of \COLMAP are listed, allowing for tentative comparison.  %

\subsection{Use-Case-Specific Experiments Conducted on Real-World Datasets}
\label{sec:experiments_qualitative}

\begin{figure*}[!t]
     \centering
   	 \resizebox{\textwidth}{!}{\subimport{figures/}{use-case_qual_results-edited.pdf_tex}}
     \caption{
		Qualitative results of \pgSGM with $4$ aggregation paths achieved on the two real-world and use-case-specific datasets, namely the TMB dataset and the FB dataset. %
		As comparison, the corresponding depth maps estimated by \COLMAP are also visualized. %
		\textbf{Rows~1~\&~2}: TMB Building scene captured from an altitude of 15\m and 8\m, respectively. %
		\textbf{Rows~3~\&~4}: TMB Container scene. %
		\textbf{Rows~5~\&~6}: Two excerpts from the FB dataset. %
     }
     \label{fig:use-case_qual_results}
\end{figure*}
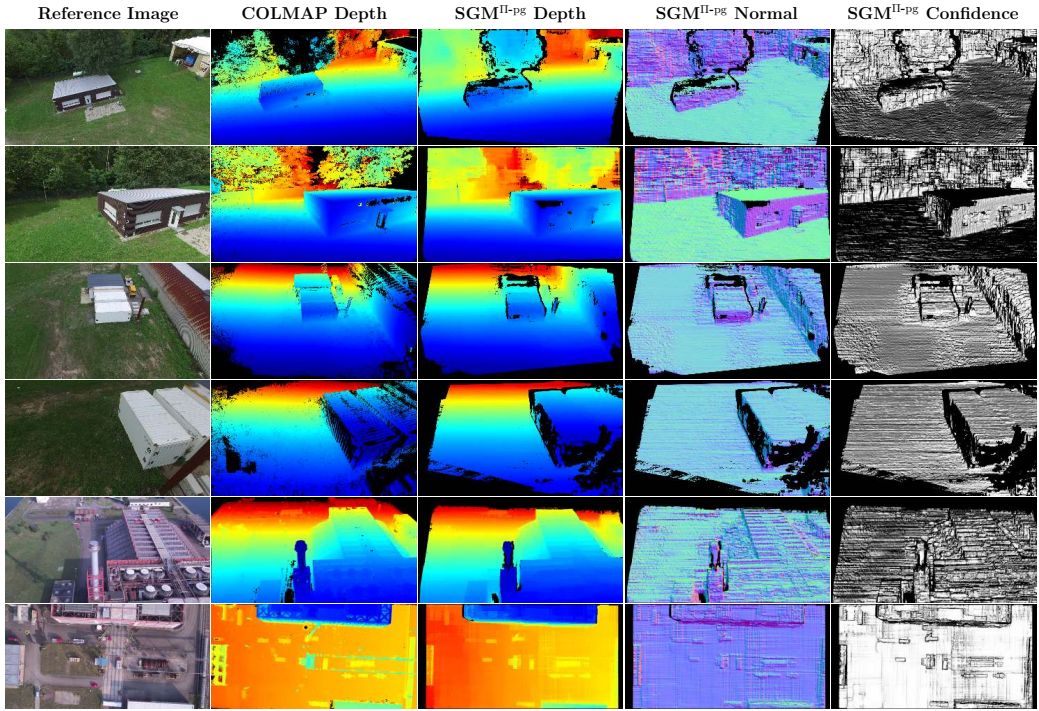

Finally, to demonstrate the performance of the presented approach on use-case-specific and real-world datasets, experiments using \pgSGM with $4$ aggregation paths and the configuration mentioned in \Cref{sec:experiments_comparisonToColmap} are conducted on the TMB and FB dataset (\cf \Cref{sec:experiments_datasets}). %
An excerpt of the computed depth, normal and confidence maps is depicted in \Cref{fig:use-case_qual_results}, together with corresponding depth maps estimated by \COLMAP as reference. %
Again, the experiments and timing measurements were conducted on a NVIDIA Titan X. %
The mean processing time in case of the TMB dataset is $690$\ms, partly varying from between $320$\ms and $1218$\ms depending on the arrangement of the input data and, in turn, the number of plane distances $\delta$ at which the scene is sampled. %
For the FB dataset, the mean processing time is $800$\ms, varying between $514$\ms and $1419$\ms again depending the arrangement of the input images. %

%% file: figures/sgmx_qual_comparison_DTU-edited.pdf_tex
%% Accompanies image file 'sgmx_qual_comparison_DTU.pdf' (pdf, eps, ps)
\begingroup%
  \makeatletter%
  \providecommand\color[2][]{%
    \errmessage{(Inkscape) Color is used for the text in Inkscape, but the package 'color.sty' is not loaded}%
    \renewcommand\color[2][]{}%
  }%
  \providecommand\transparent[1]{%
    \errmessage{(Inkscape) Transparency is used (non-zero) for the text in Inkscape, but the package 'transparent.sty' is not loaded}%
    \renewcommand\transparent[1]{}%
  }%
  \providecommand\rotatebox[2]{#2}%
  \newcommand*\fsize{\dimexpr\f@size pt\relax}%
  \newcommand*\lineheight[1]{\fontsize{\fsize}{#1\fsize}\selectfont}%
  \ifx\svgwidth\undefined%
    \setlength{\unitlength}{972.75829886bp}%
    \ifx\svgscale\undefined%
      \relax%
    \else%
      \setlength{\unitlength}{\unitlength * \real{\svgscale}}%
    \fi%
  \else%
    \setlength{\unitlength}{\svgwidth}%
  \fi%
  \global\let\svgwidth\undefined%
  \global\let\svgscale\undefined%
  \makeatother%
  \begin{picture}(1,0.7874984)%
  	\begin{scriptsize}
  	\lineheight{1}%
    \setlength\tabcolsep{0pt}%
    \put(0,0){\includegraphics[width=\unitlength,page=1]{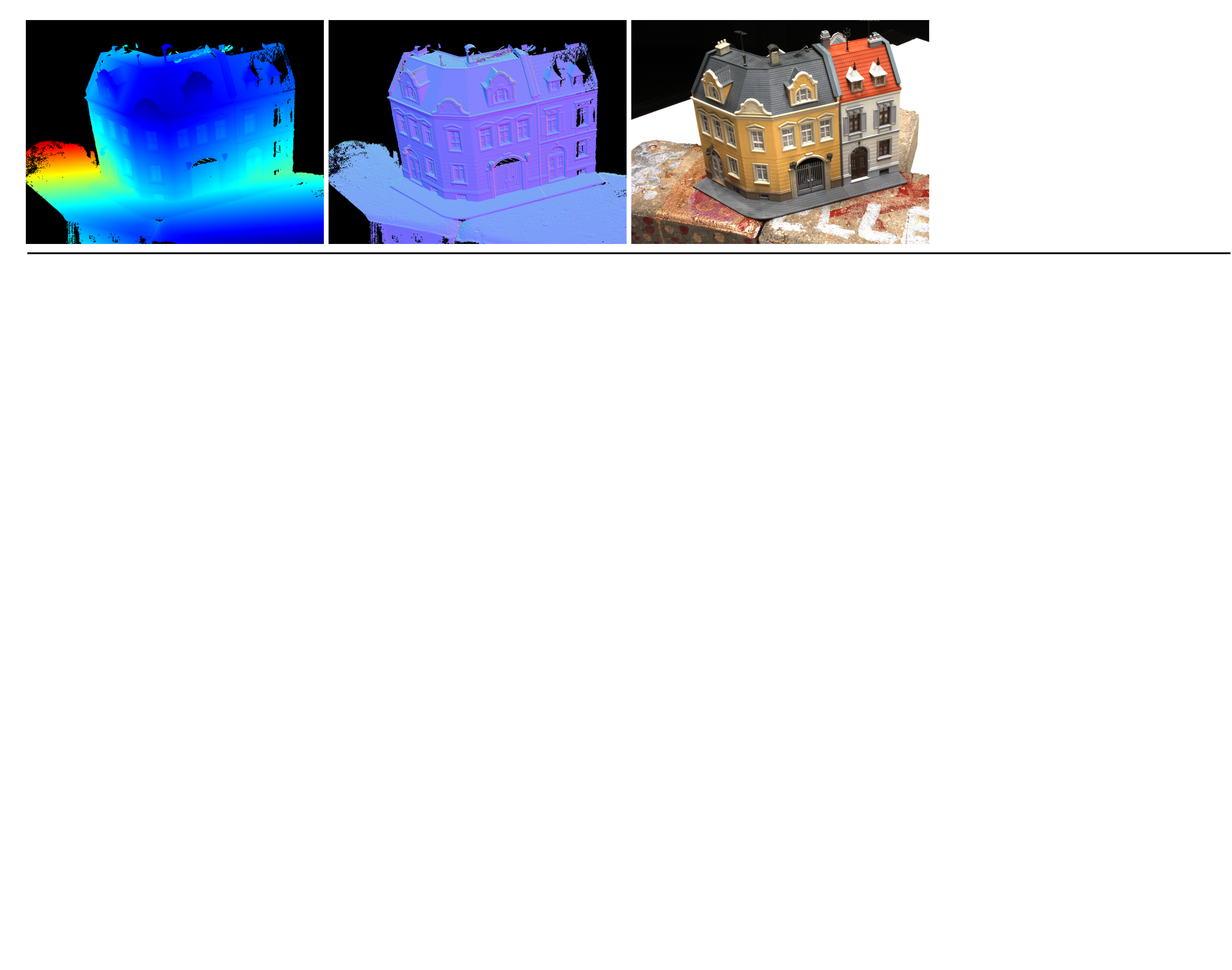}}%
    \put(0.14110714,0.7765638){\color[rgb]{0,0,0}\makebox(0,0)[t]{\lineheight{1.25}\smash{\begin{tabular}[t]{c}Ground Truth Depth\end{tabular}}}}%
    \put(0.38682653,0.77516887){\color[rgb]{0,0,0}\makebox(0,0)[t]{\lineheight{1.25}\smash{\begin{tabular}[t]{c}Ground Truth Normal\end{tabular}}}}%
    \put(0.63194161,0.7765638){\color[rgb]{0,0,0}\makebox(0,0)[t]{\lineheight{1.25}\smash{\begin{tabular}[t]{c}Reference Image\end{tabular}}}}%
    \put(0.13786752,0.55886815){\color[rgb]{0,0,0}\makebox(0,0)[t]{\lineheight{1.25}\smash{\begin{tabular}[t]{c}Depth\end{tabular}}}}%
    \put(0.38622219,0.55720969){\color[rgb]{0,0,0}\makebox(0,0)[t]{\lineheight{1.25}\smash{\begin{tabular}[t]{c}Normal\end{tabular}}}}%
    \put(0.63194161,0.55886815){\color[rgb]{0,0,0}\makebox(0,0)[t]{\lineheight{1.25}\smash{\begin{tabular}[t]{c}Confidence\end{tabular}}}}%
    \put(0,0){\includegraphics[width=\unitlength,page=2]{sgmx_qual_comparison_DTU.pdf}}%
    \put(0.0109346,0.45966367){\color[rgb]{0,0,0}\rotatebox{90}{\makebox(0,0)[t]{\lineheight{1.25}\smash{\begin{tabular}[t]{c}\piSGM\end{tabular}}}}}%
    \put(0.0109346,0.27693825){\color[rgb]{0,0,0}\rotatebox{90}{\makebox(0,0)[t]{\lineheight{1.25}\smash{\begin{tabular}[t]{c}\snSGM\end{tabular}}}}}%
    \put(0.0109346,0.09151432){\color[rgb]{0,0,0}\rotatebox{90}{\makebox(0,0)[t]{\lineheight{1.25}\smash{\begin{tabular}[t]{c}\pgSGM\end{tabular}}}}}%
    \put(0.87765534,0.55886815){\color[rgb]{0,0,0}\makebox(0,0)[t]{\lineheight{1.25}\smash{\begin{tabular}[t]{c}Depth Difference\end{tabular}}}}%
  	\end{scriptsize}
  \end{picture}%
\endgroup%

%% file: figures/sgmx_qual_comparison_3DOMcity-edited.pdf_tex
%% Accompanies image file 'sgmx_qual_comparison_3DOMcity.pdf' (pdf, eps, ps)
\begingroup%
  \makeatletter%
  \providecommand\color[2][]{%
    \errmessage{(Inkscape) Color is used for the text in Inkscape, but the package 'color.sty' is not loaded}%
    \renewcommand\color[2][]{}%
  }%
  \providecommand\transparent[1]{%
    \errmessage{(Inkscape) Transparency is used (non-zero) for the text in Inkscape, but the package 'transparent.sty' is not loaded}%
    \renewcommand\transparent[1]{}%
  }%
  \providecommand\rotatebox[2]{#2}%
  \newcommand*\fsize{\dimexpr\f@size pt\relax}%
  \newcommand*\lineheight[1]{\fontsize{\fsize}{#1\fsize}\selectfont}%
  \ifx\svgwidth\undefined%
    \setlength{\unitlength}{819.9312946bp}%
    \ifx\svgscale\undefined%
      \relax%
    \else%
      \setlength{\unitlength}{\unitlength * \real{\svgscale}}%
    \fi%
  \else%
    \setlength{\unitlength}{\svgwidth}%
  \fi%
  \global\let\svgwidth\undefined%
  \global\let\svgscale\undefined%
  \makeatother%
  \begin{picture}(1,1.11284001)%
  	\begin{scriptsize}
    \lineheight{1}%
    \setlength\tabcolsep{0pt}%
    \put(0,0){\includegraphics[width=\unitlength,page=1]{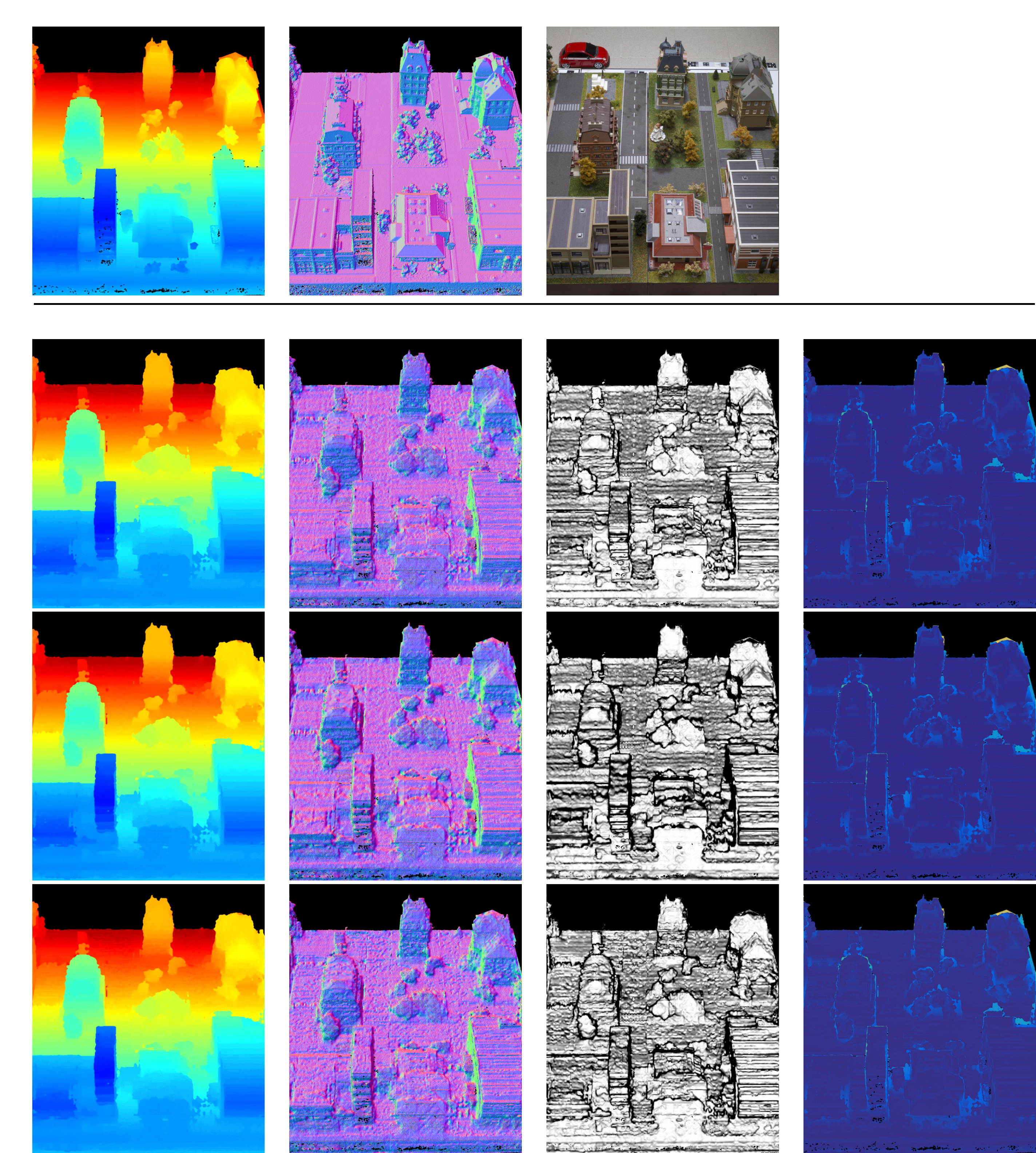}}%
    \put(0.14357774,1.09986732){\color[rgb]{0,0,0}\makebox(0,0)[t]{\lineheight{1.25}\smash{\begin{tabular}[t]{c}Ground Truth Depth\end{tabular}}}}%
    \put(0.39180645,1.09821238){\color[rgb]{0,0,0}\makebox(0,0)[t]{\lineheight{1.25}\smash{\begin{tabular}[t]{c}Ground Truth Normal\end{tabular}}}}%
    \put(0.63927236,1.09986732){\color[rgb]{0,0,0}\makebox(0,0)[t]{\lineheight{1.25}\smash{\begin{tabular}[t]{c}Reference Image\end{tabular}}}}%
    \put(0.14321924,0.79716658){\color[rgb]{0,0,0}\makebox(0,0)[t]{\lineheight{1.25}\smash{\begin{tabular}[t]{c}Depth\end{tabular}}}}%
    \put(0.391448,0.79551163){\color[rgb]{0,0,0}\makebox(0,0)[t]{\lineheight{1.25}\smash{\begin{tabular}[t]{c}Normal\end{tabular}}}}%
    \put(0.63963087,0.79551163){\color[rgb]{0,0,0}\makebox(0,0)[t]{\lineheight{1.25}\smash{\begin{tabular}[t]{c}Confidence\end{tabular}}}}%
    \put(0.0129727,0.65637571){\color[rgb]{0,0,0}\rotatebox{90}{\makebox(0,0)[t]{\lineheight{1.25}\smash{\begin{tabular}[t]{c}\piSGM\end{tabular}}}}}%
    \put(0.01380642,0.39342882){\color[rgb]{0,0,0}\rotatebox{90}{\makebox(0,0)[t]{\lineheight{1.25}\smash{\begin{tabular}[t]{c}\snSGM\end{tabular}}}}}%
    \put(0.01282263,0.13048198){\color[rgb]{0,0,0}\rotatebox{90}{\makebox(0,0)[t]{\lineheight{1.25}\smash{\begin{tabular}[t]{c}\pgSGM\end{tabular}}}}}%
    \put(0.8874245,0.79716658){\color[rgb]{0,0,0}\makebox(0,0)[t]{\lineheight{1.25}\smash{\begin{tabular}[t]{c}Depth Difference\end{tabular}}}}%
    \end{scriptsize}
  \end{picture}%
\endgroup%

%% file: figures/ROC_ACC_Density_combined-edited.pdf_tex
%% Accompanies image file 'ROC_ACC_Density_combined.pdf' (pdf, eps, ps)
\begingroup%
  \makeatletter%
  \providecommand\color[2][]{%
    \errmessage{(Inkscape) Color is used for the text in Inkscape, but the package 'color.sty' is not loaded}%
    \renewcommand\color[2][]{}%
  }%
  \providecommand\transparent[1]{%
    \errmessage{(Inkscape) Transparency is used (non-zero) for the text in Inkscape, but the package 'transparent.sty' is not loaded}%
    \renewcommand\transparent[1]{}%
  }%
  \providecommand\rotatebox[2]{#2}%
  \newcommand*\fsize{\dimexpr\f@size pt\relax}%
  \newcommand*\lineheight[1]{\fontsize{\fsize}{#1\fsize}\selectfont}%
  \ifx\svgwidth\undefined%
    \setlength{\unitlength}{936.75640869bp}%
    \ifx\svgscale\undefined%
      \relax%
    \else%
      \setlength{\unitlength}{\unitlength * \real{\svgscale}}%
    \fi%
  \else%
    \setlength{\unitlength}{\svgwidth}%
  \fi%
  \global\let\svgwidth\undefined%
  \global\let\svgscale\undefined%
  \makeatother%
  \begin{picture}(1,0.36973325)%
    \lineheight{1}%
    \setlength\tabcolsep{0pt}%
    \begin{Large}
    \put(0,0){\includegraphics[width=\unitlength,page=1]{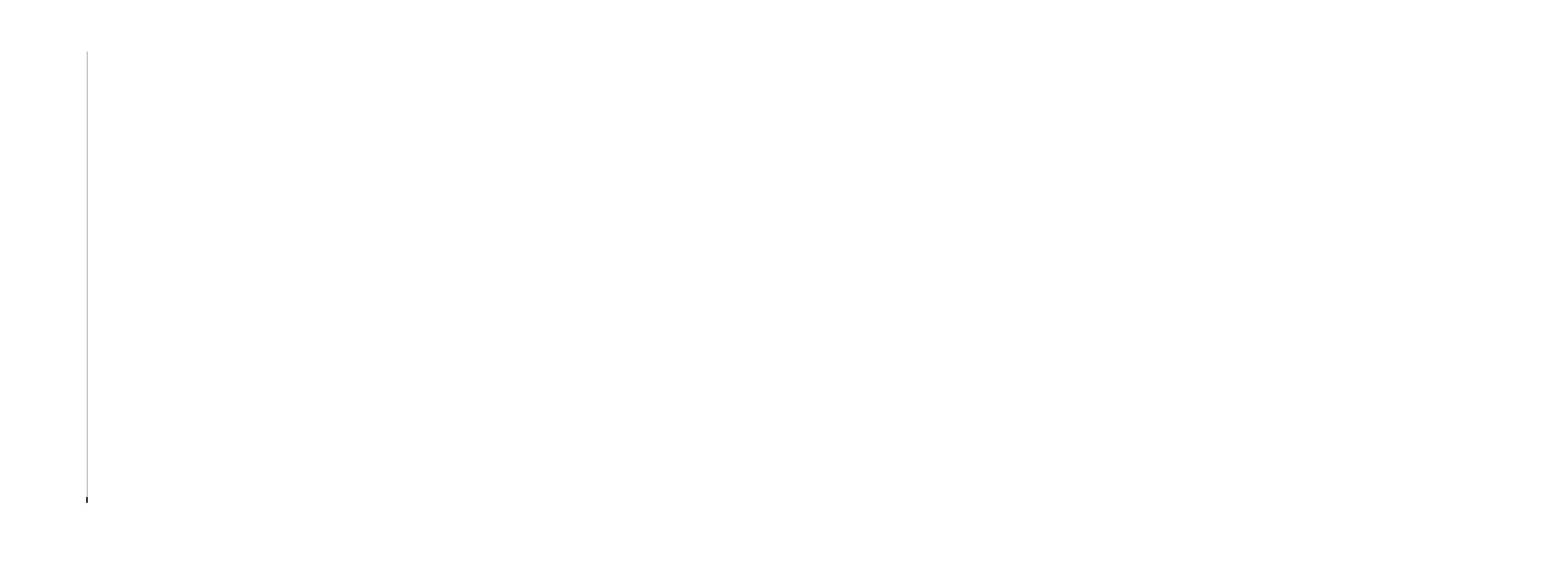}}%
    \put(0.05129902,0.03566562){\color[rgb]{0,0,0}\makebox(0,0)[lt]{\lineheight{1.25}\smash{\begin{tabular}[t]{l}$5$\end{tabular}}}}%
    \put(0,0){\includegraphics[width=\unitlength,page=2]{ROC_ACC_Density_combined.pdf}}%
    \put(0.11247086,0.03566562){\color[rgb]{0,0,0}\makebox(0,0)[lt]{\lineheight{1.25}\smash{\begin{tabular}[t]{l}$20$\end{tabular}}}}%
    \put(0,0){\includegraphics[width=\unitlength,page=3]{ROC_ACC_Density_combined.pdf}}%
    \put(0.19947094,0.03566562){\color[rgb]{0,0,0}\makebox(0,0)[lt]{\lineheight{1.25}\smash{\begin{tabular}[t]{l}$40$\end{tabular}}}}%
    \put(0,0){\includegraphics[width=\unitlength,page=4]{ROC_ACC_Density_combined.pdf}}%
    \put(0.28647104,0.03566562){\color[rgb]{0,0,0}\makebox(0,0)[lt]{\lineheight{1.25}\smash{\begin{tabular}[t]{l}$60$\end{tabular}}}}%
    \put(0,0){\includegraphics[width=\unitlength,page=5]{ROC_ACC_Density_combined.pdf}}%
    \put(0.37347113,0.03566562){\color[rgb]{0,0,0}\makebox(0,0)[lt]{\lineheight{1.25}\smash{\begin{tabular}[t]{l}$80$\end{tabular}}}}%
    \put(0,0){\includegraphics[width=\unitlength,page=6]{ROC_ACC_Density_combined.pdf}}%
    \put(0.45639298,0.03566562){\color[rgb]{0,0,0}\makebox(0,0)[lt]{\lineheight{1.25}\smash{\begin{tabular}[t]{l}$100$\end{tabular}}}}%
    \put(0.20150448,0.01574983){\color[rgb]{0,0,0}\makebox(0,0)[lt]{\lineheight{1.25}\smash{\begin{tabular}[t]{l}\huge{Density} \Large{(in \%)}\end{tabular}}}}%
    \put(0,0){\includegraphics[width=\unitlength,page=7]{ROC_ACC_Density_combined.pdf}}%
    \put(0.03159172,0.08394746){\color[rgb]{0,0,0}\makebox(0,0)[lt]{\lineheight{1.25}\smash{\begin{tabular}[t]{l}$10$\end{tabular}}}}%
    \put(0,0){\includegraphics[width=\unitlength,page=8]{ROC_ACC_Density_combined.pdf}}%
    \put(0.03159172,0.11988618){\color[rgb]{0,0,0}\makebox(0,0)[lt]{\lineheight{1.25}\smash{\begin{tabular}[t]{l}$11$\end{tabular}}}}%
    \put(0,0){\includegraphics[width=\unitlength,page=9]{ROC_ACC_Density_combined.pdf}}%
    \put(0.03159172,0.15582489){\color[rgb]{0,0,0}\makebox(0,0)[lt]{\lineheight{1.25}\smash{\begin{tabular}[t]{l}$12$\end{tabular}}}}%
    \put(0,0){\includegraphics[width=\unitlength,page=10]{ROC_ACC_Density_combined.pdf}}%
    \put(0.03159172,0.1917636){\color[rgb]{0,0,0}\makebox(0,0)[lt]{\lineheight{1.25}\smash{\begin{tabular}[t]{l}$13$\end{tabular}}}}%
    \put(0,0){\includegraphics[width=\unitlength,page=11]{ROC_ACC_Density_combined.pdf}}%
    \put(0.03159172,0.22770232){\color[rgb]{0,0,0}\makebox(0,0)[lt]{\lineheight{1.25}\smash{\begin{tabular}[t]{l}$14$\end{tabular}}}}%
    \put(0,0){\includegraphics[width=\unitlength,page=12]{ROC_ACC_Density_combined.pdf}}%
    \put(0.03159172,0.26364104){\color[rgb]{0,0,0}\makebox(0,0)[lt]{\lineheight{1.25}\smash{\begin{tabular}[t]{l}$15$\end{tabular}}}}%
    \put(0,0){\includegraphics[width=\unitlength,page=13]{ROC_ACC_Density_combined.pdf}}%
    \put(0.03159172,0.29957975){\color[rgb]{0,0,0}\makebox(0,0)[lt]{\lineheight{1.25}\smash{\begin{tabular}[t]{l}$16$\end{tabular}}}}%
    \put(0.02376885,0.12592421){\color[rgb]{0,0,0}\rotatebox{90}{\makebox(0,0)[lt]{\lineheight{1.25}\smash{\begin{tabular}[t]{l}\huge{Error rate} \Large{(in \%)}\end{tabular}}}}}%
    \put(0,0){\includegraphics[width=\unitlength,page=14]{ROC_ACC_Density_combined.pdf}}%
    \put(0.24170303,0.34320021){\color[rgb]{0,0,0}\makebox(0,0)[lt]{\lineheight{1.25}\smash{\begin{tabular}[t]{l}\huge{\textbf{DTU}}\end{tabular}}}}%
    \put(0,0){\includegraphics[width=\unitlength,page=15]{ROC_ACC_Density_combined.pdf}}%
    \put(0.39,0.31552492){\color[rgb]{0,0,0}\makebox(0,0)[lt]{\lineheight{1.25}\smash{\begin{tabular}[t]{l}\piSGM\end{tabular}}}}%
    \put(0,0){\includegraphics[width=\unitlength,page=16]{ROC_ACC_Density_combined.pdf}}%
    \put(0.39,0.29571){\color[rgb]{0,0,0}\makebox(0,0)[lt]{\lineheight{1.25}\smash{\begin{tabular}[t]{l}\snSGM\end{tabular}}}}%
    \put(0,0){\includegraphics[width=\unitlength,page=17]{ROC_ACC_Density_combined.pdf}}%
    \put(0.39,0.27689508){\color[rgb]{0,0,0}\makebox(0,0)[lt]{\lineheight{1.25}\smash{\begin{tabular}[t]{l}\pgSGM\end{tabular}}}}%
    \put(0,0){\includegraphics[width=\unitlength,page=18]{ROC_ACC_Density_combined.pdf}}%
    \put(0.5504221,0.03566562){\color[rgb]{0,0,0}\makebox(0,0)[lt]{\lineheight{1.25}\smash{\begin{tabular}[t]{l}$5$\end{tabular}}}}%
    \put(0,0){\includegraphics[width=\unitlength,page=19]{ROC_ACC_Density_combined.pdf}}%
    \put(0.61288338,0.03566562){\color[rgb]{0,0,0}\makebox(0,0)[lt]{\lineheight{1.25}\smash{\begin{tabular}[t]{l}$20$\end{tabular}}}}%
    \put(0,0){\includegraphics[width=\unitlength,page=20]{ROC_ACC_Density_combined.pdf}}%
    \put(0.70160273,0.03566562){\color[rgb]{0,0,0}\makebox(0,0)[lt]{\lineheight{1.25}\smash{\begin{tabular}[t]{l}$40$\end{tabular}}}}%
    \put(0,0){\includegraphics[width=\unitlength,page=21]{ROC_ACC_Density_combined.pdf}}%
    \put(0.79032208,0.03566562){\color[rgb]{0,0,0}\makebox(0,0)[lt]{\lineheight{1.25}\smash{\begin{tabular}[t]{l}$60$\end{tabular}}}}%
    \put(0,0){\includegraphics[width=\unitlength,page=22]{ROC_ACC_Density_combined.pdf}}%
    \put(0.87904143,0.03566562){\color[rgb]{0,0,0}\makebox(0,0)[lt]{\lineheight{1.25}\smash{\begin{tabular}[t]{l}$80$\end{tabular}}}}%
    \put(0,0){\includegraphics[width=\unitlength,page=23]{ROC_ACC_Density_combined.pdf}}%
    \put(0.96368254,0.03566562){\color[rgb]{0,0,0}\makebox(0,0)[lt]{\lineheight{1.25}\smash{\begin{tabular}[t]{l}$100$\end{tabular}}}}%
    \put(0.70471081,0.01574983){\color[rgb]{0,0,0}\makebox(0,0)[lt]{\lineheight{1.25}\smash{\begin{tabular}[t]{l}\huge{Density} \Large{(in \%)}\end{tabular}}}}%
    \put(0,0){\includegraphics[width=\unitlength,page=24]{ROC_ACC_Density_combined.pdf}}%
    \put(0.53887128,0.05418083){\color[rgb]{0,0,0}\makebox(0,0)[lt]{\lineheight{1.25}\smash{\begin{tabular}[t]{l}$2$\end{tabular}}}}%
    \put(0,0){\includegraphics[width=\unitlength,page=25]{ROC_ACC_Density_combined.pdf}}%
    \put(0.53887128,0.11590166){\color[rgb]{0,0,0}\makebox(0,0)[lt]{\lineheight{1.25}\smash{\begin{tabular}[t]{l}$3$\end{tabular}}}}%
    \put(0,0){\includegraphics[width=\unitlength,page=26]{ROC_ACC_Density_combined.pdf}}%
    \put(0.53887128,0.1776225){\color[rgb]{0,0,0}\makebox(0,0)[lt]{\lineheight{1.25}\smash{\begin{tabular}[t]{l}$4$\end{tabular}}}}%
    \put(0,0){\includegraphics[width=\unitlength,page=27]{ROC_ACC_Density_combined.pdf}}%
    \put(0.53887128,0.23934334){\color[rgb]{0,0,0}\makebox(0,0)[lt]{\lineheight{1.25}\smash{\begin{tabular}[t]{l}$5$\end{tabular}}}}%
    \put(0,0){\includegraphics[width=\unitlength,page=28]{ROC_ACC_Density_combined.pdf}}%
    \put(0.53887128,0.30106417){\color[rgb]{0,0,0}\makebox(0,0)[lt]{\lineheight{1.25}\smash{\begin{tabular}[t]{l}$6$\end{tabular}}}}%
    \put(0.53104841,0.12592421){\color[rgb]{0,0,0}\rotatebox{90}{\makebox(0,0)[lt]{\lineheight{1.25}\smash{\begin{tabular}[t]{l}\huge{Error rate} \Large{(in \%)}\end{tabular}}}}}%
    \put(0,0){\includegraphics[width=\unitlength,page=29]{ROC_ACC_Density_combined.pdf}}%
    \put(0.71843837,0.34320021){\color[rgb]{0,0,0}\makebox(0,0)[lt]{\lineheight{1.25}\smash{\begin{tabular}[t]{l}\huge{\textbf{3DOMcity}}\end{tabular}}}}%
    \put(0,0){\includegraphics[width=\unitlength,page=30]{ROC_ACC_Density_combined.pdf}}%
    \put(0.896,0.31452492){\color[rgb]{0,0,0}\makebox(0,0)[lt]{\lineheight{1.25}\smash{\begin{tabular}[t]{l}\piSGM\end{tabular}}}}%
    \put(0,0){\includegraphics[width=\unitlength,page=31]{ROC_ACC_Density_combined.pdf}}%
    \put(0.896,0.29571){\color[rgb]{0,0,0}\makebox(0,0)[lt]{\lineheight{1.25}\smash{\begin{tabular}[t]{l}\snSGM\end{tabular}}}}%
    \put(0,0){\includegraphics[width=\unitlength,page=32]{ROC_ACC_Density_combined.pdf}}%
    \put(0.896,0.27689508){\color[rgb]{0,0,0}\makebox(0,0)[lt]{\lineheight{1.25}\smash{\begin{tabular}[t]{l}\pgSGM\end{tabular}}}}%  
    \end{Large}
  \end{picture}%
\endgroup%

%% file: figures/diffPlane_qual_comparison-edited.pdf_tex
%% Accompanies image file 'diffPlane_qual_comparison.pdf' (pdf, eps, ps)
\begingroup%
  \makeatletter%
  \providecommand\color[2][]{%
    \errmessage{(Inkscape) Color is used for the text in Inkscape, but the package 'color.sty' is not loaded}%
    \renewcommand\color[2][]{}%
  }%
  \providecommand\transparent[1]{%
    \errmessage{(Inkscape) Transparency is used (non-zero) for the text in Inkscape, but the package 'transparent.sty' is not loaded}%
    \renewcommand\transparent[1]{}%
  }%
  \providecommand\rotatebox[2]{#2}%
  \newcommand*\fsize{\dimexpr\f@size pt\relax}%
  \newcommand*\lineheight[1]{\fontsize{\fsize}{#1\fsize}\selectfont}%
  \ifx\svgwidth\undefined%
    \setlength{\unitlength}{723.66138849bp}%
    \ifx\svgscale\undefined%
      \relax%
    \else%
      \setlength{\unitlength}{\unitlength * \real{\svgscale}}%
    \fi%
  \else%
    \setlength{\unitlength}{\svgwidth}%
  \fi%
  \global\let\svgwidth\undefined%
  \global\let\svgscale\undefined%
  \makeatother%
  \begin{picture}(1,0.3991818)%
  \begin{scriptsize}
    \lineheight{1}%
    \setlength\tabcolsep{0pt}%
    \put(0,0){\includegraphics[width=\unitlength,page=1]{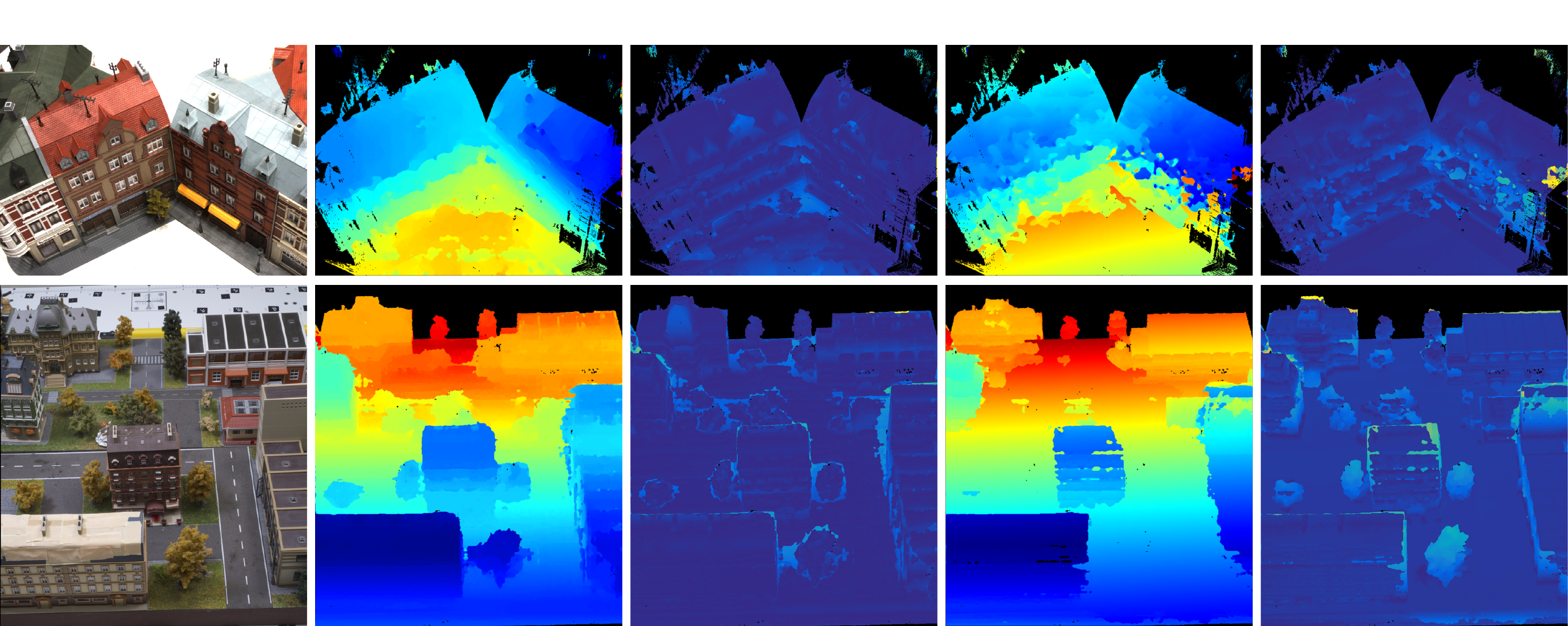}}%
    \put(0.1,0.38448333){\color[rgb]{0,0,0}\makebox(0,0)[ct]{\lineheight{1.25}\smash{\begin{tabular}[t]{c}Reference Image\end{tabular}}}}%
    \put(0.4,0.38448333){\color[rgb]{0,0,0}\makebox(0,0)[ct]{\lineheight{1.25}\smash{\begin{tabular}[t]{c}Fronto-Parallel Sampling\end{tabular}}}}%
    \put(0.8,0.38448333){\color[rgb]{0,0,0}\makebox(0,0)[ct]{\lineheight{1.25}\smash{\begin{tabular}[t]{c}Horizontal Sampling\end{tabular}}}}%
    \end{scriptsize}
  \end{picture}%
\endgroup%

%% file: figures/postfilteringAccCpl_combined-edited.pdf_tex
%% Accompanies image file 'postfilteringAccCpl_combined.pdf' (pdf, eps, ps)
\begingroup%
  \makeatletter%
  \providecommand\color[2][]{%
    \errmessage{(Inkscape) Color is used for the text in Inkscape, but the package 'color.sty' is not loaded}%
    \renewcommand\color[2][]{}%
  }%
  \providecommand\transparent[1]{%
    \errmessage{(Inkscape) Transparency is used (non-zero) for the text in Inkscape, but the package 'transparent.sty' is not loaded}%
    \renewcommand\transparent[1]{}%
  }%
  \providecommand\rotatebox[2]{#2}%
  \newcommand*\fsize{\dimexpr\f@size pt\relax}%
  \newcommand*\lineheight[1]{\fontsize{\fsize}{#1\fsize}\selectfont}%
  \ifx\svgwidth\undefined%
    \setlength{\unitlength}{1430.25576782bp}%
    \ifx\svgscale\undefined%
      \relax%
    \else%
      \setlength{\unitlength}{\unitlength * \real{\svgscale}}%
    \fi%
  \else%
    \setlength{\unitlength}{\svgwidth}%
  \fi%
  \global\let\svgwidth\undefined%
  \global\let\svgscale\undefined%
  \makeatother%
  \begin{picture}(1,0.24215948)%
    \lineheight{1}%
    \setlength\tabcolsep{0pt}%
    \begin{Large}
    \put(0,0){\includegraphics[width=\unitlength,page=1]{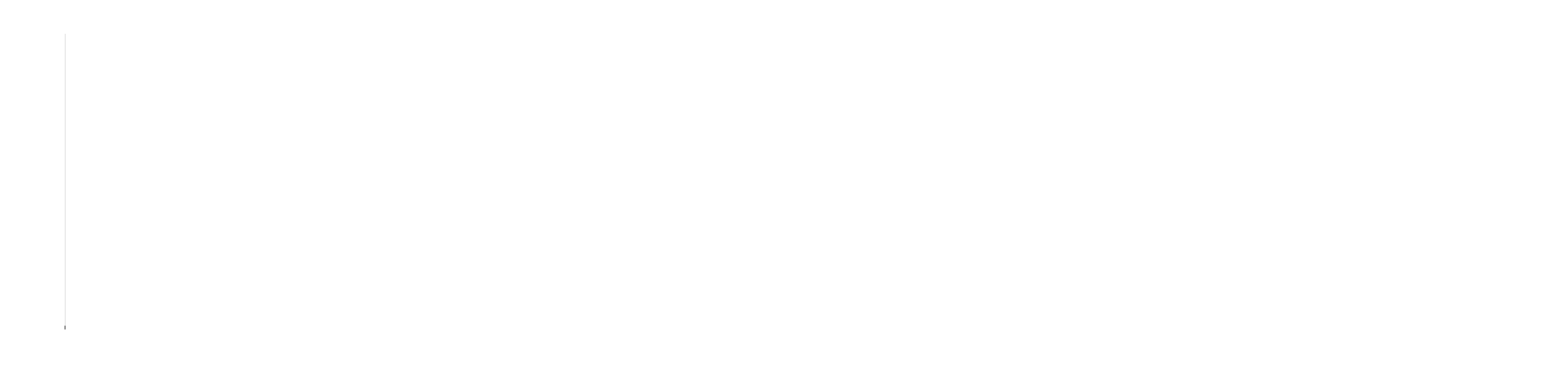}}%
    \put(0,0){\includegraphics[width=\unitlength,page=2]{postfilteringAccCpl_combined.pdf}}%
    \put(0,0){\includegraphics[width=\unitlength,page=3]{postfilteringAccCpl_combined.pdf}}%
    \put(0,0){\includegraphics[width=\unitlength,page=4]{postfilteringAccCpl_combined.pdf}}%
    \put(0,0){\includegraphics[width=\unitlength,page=5]{postfilteringAccCpl_combined.pdf}}%
    \put(0,0){\includegraphics[width=\unitlength,page=6]{postfilteringAccCpl_combined.pdf}}%
    \put(0,0){\includegraphics[width=\unitlength,page=7]{postfilteringAccCpl_combined.pdf}}%
    \put(0,0){\includegraphics[width=\unitlength,page=8]{postfilteringAccCpl_combined.pdf}}%
    \put(0,0){\includegraphics[width=\unitlength,page=9]{postfilteringAccCpl_combined.pdf}}%
    \put(0,0){\includegraphics[width=\unitlength,page=10]{postfilteringAccCpl_combined.pdf}}%
    \put(0,0){\includegraphics[width=\unitlength,page=11]{postfilteringAccCpl_combined.pdf}}%
    \put(0,0){\includegraphics[width=\unitlength,page=12]{postfilteringAccCpl_combined.pdf}}%
    \put(0,0){\includegraphics[width=\unitlength,page=13]{postfilteringAccCpl_combined.pdf}}%
    \put(0,0){\includegraphics[width=\unitlength,page=14]{postfilteringAccCpl_combined.pdf}}%
    \put(0,0){\includegraphics[width=\unitlength,page=15]{postfilteringAccCpl_combined.pdf}}%
    \put(0,0){\includegraphics[width=\unitlength,page=16]{postfilteringAccCpl_combined.pdf}}%
    \put(0,0){\includegraphics[width=\unitlength,page=17]{postfilteringAccCpl_combined.pdf}}%
    \put(0,0){\includegraphics[width=\unitlength,page=18]{postfilteringAccCpl_combined.pdf}}%
    \put(0,0){\includegraphics[width=\unitlength,page=19]{postfilteringAccCpl_combined.pdf}}%
    \put(0,0){\includegraphics[width=\unitlength,page=20]{postfilteringAccCpl_combined.pdf}}%
    \put(0,0){\includegraphics[width=\unitlength,page=21]{postfilteringAccCpl_combined.pdf}}%
    \put(0,0){\includegraphics[width=\unitlength,page=22]{postfilteringAccCpl_combined.pdf}}%
    \put(0.03615046,0.02335946){\color[rgb]{0,0,0}\makebox(0,0)[lt]{\lineheight{1.25}\smash{\begin{tabular}[t]{l}$30$\end{tabular}}}}%
    \put(0,0){\includegraphics[width=\unitlength,page=23]{postfilteringAccCpl_combined.pdf}}%
    \put(0.07407025,0.02335946){\color[rgb]{0,0,0}\makebox(0,0)[lt]{\lineheight{1.25}\smash{\begin{tabular}[t]{l}$40$\end{tabular}}}}%
    \put(0,0){\includegraphics[width=\unitlength,page=24]{postfilteringAccCpl_combined.pdf}}%
    \put(0.11199004,0.02335946){\color[rgb]{0,0,0}\makebox(0,0)[lt]{\lineheight{1.25}\smash{\begin{tabular}[t]{l}$50$\end{tabular}}}}%
    \put(0,0){\includegraphics[width=\unitlength,page=25]{postfilteringAccCpl_combined.pdf}}%
    \put(0.14990983,0.02335946){\color[rgb]{0,0,0}\makebox(0,0)[lt]{\lineheight{1.25}\smash{\begin{tabular}[t]{l}$60$\end{tabular}}}}%
    \put(0,0){\includegraphics[width=\unitlength,page=26]{postfilteringAccCpl_combined.pdf}}%
    \put(0.18782962,0.02335946){\color[rgb]{0,0,0}\makebox(0,0)[lt]{\lineheight{1.25}\smash{\begin{tabular}[t]{l}$70$\end{tabular}}}}%
    \put(0,0){\includegraphics[width=\unitlength,page=27]{postfilteringAccCpl_combined.pdf}}%
    \put(0.22574941,0.02335946){\color[rgb]{0,0,0}\makebox(0,0)[lt]{\lineheight{1.25}\smash{\begin{tabular}[t]{l}$80$\end{tabular}}}}%
    \put(0,0){\includegraphics[width=\unitlength,page=28]{postfilteringAccCpl_combined.pdf}}%
    \put(0.2636692,0.02335946){\color[rgb]{0,0,0}\makebox(0,0)[lt]{\lineheight{1.25}\smash{\begin{tabular}[t]{l}$90$\end{tabular}}}}%
    \put(0,0){\includegraphics[width=\unitlength,page=29]{postfilteringAccCpl_combined.pdf}}%
    \put(0.29891792,0.02335946){\color[rgb]{0,0,0}\makebox(0,0)[lt]{\lineheight{1.25}\smash{\begin{tabular}[t]{l}$100$\end{tabular}}}}%
    \put(0.1160765,0.00831546){\color[rgb]{0,0,0}\makebox(0,0)[lt]{\lineheight{1.25}\smash{\begin{tabular}[t]{l}\huge{Completeness} \Large{(in \%)}\end{tabular}}}}%
    \put(0,0){\includegraphics[width=\unitlength,page=30]{postfilteringAccCpl_combined.pdf}}%
    \put(0.02591409,0.03144368){\color[rgb]{0,0,0}\makebox(0,0)[lt]{\lineheight{1.25}\smash{\begin{tabular}[t]{l}$30$\end{tabular}}}}%
    \put(0,0){\includegraphics[width=\unitlength,page=31]{postfilteringAccCpl_combined.pdf}}%
    \put(0.02591409,0.05800835){\color[rgb]{0,0,0}\makebox(0,0)[lt]{\lineheight{1.25}\smash{\begin{tabular}[t]{l}$40$\end{tabular}}}}%
    \put(0,0){\includegraphics[width=\unitlength,page=32]{postfilteringAccCpl_combined.pdf}}%
    \put(0.02591409,0.08457303){\color[rgb]{0,0,0}\makebox(0,0)[lt]{\lineheight{1.25}\smash{\begin{tabular}[t]{l}$50$\end{tabular}}}}%
    \put(0,0){\includegraphics[width=\unitlength,page=33]{postfilteringAccCpl_combined.pdf}}%
    \put(0.02591409,0.11113771){\color[rgb]{0,0,0}\makebox(0,0)[lt]{\lineheight{1.25}\smash{\begin{tabular}[t]{l}$60$\end{tabular}}}}%
    \put(0,0){\includegraphics[width=\unitlength,page=34]{postfilteringAccCpl_combined.pdf}}%
    \put(0.02591409,0.13770239){\color[rgb]{0,0,0}\makebox(0,0)[lt]{\lineheight{1.25}\smash{\begin{tabular}[t]{l}$70$\end{tabular}}}}%
    \put(0,0){\includegraphics[width=\unitlength,page=35]{postfilteringAccCpl_combined.pdf}}%
    \put(0.02591409,0.16426707){\color[rgb]{0,0,0}\makebox(0,0)[lt]{\lineheight{1.25}\smash{\begin{tabular}[t]{l}$80$\end{tabular}}}}%
    \put(0,0){\includegraphics[width=\unitlength,page=36]{postfilteringAccCpl_combined.pdf}}%
    \put(0.02591409,0.19083174){\color[rgb]{0,0,0}\makebox(0,0)[lt]{\lineheight{1.25}\smash{\begin{tabular}[t]{l}$90$\end{tabular}}}}%
    \put(0,0){\includegraphics[width=\unitlength,page=37]{postfilteringAccCpl_combined.pdf}}%
    \put(0.02057195,0.21739643){\color[rgb]{0,0,0}\makebox(0,0)[lt]{\lineheight{1.25}\smash{\begin{tabular}[t]{l}$100$\end{tabular}}}}%
    \put(0.01544831,0.08370946){\color[rgb]{0,0,0}\rotatebox{90}{\makebox(0,0)[lt]{\lineheight{1.25}\smash{\begin{tabular}[t]{l}\huge{Accuracy} \Large{(in \%)}\end{tabular}}}}}%
    \put(0,0){\includegraphics[width=\unitlength,page=38]{postfilteringAccCpl_combined.pdf}}%
    \put(0.15417065,0.22478147){\color[rgb]{0,0,0}\makebox(0,0)[lt]{\lineheight{1.25}\smash{\begin{tabular}[t]{l}\huge{\textbf{\piSGM}}\end{tabular}}}}%
    \put(0,0){\includegraphics[width=\unitlength,page=39]{postfilteringAccCpl_combined.pdf}}%
    \put(0.23888962,0.08113103){\color[rgb]{0,0,0}\makebox(0,0)[lt]{\lineheight{1.25}\smash{\begin{tabular}[t]{l}DoG\end{tabular}}}}%
    \put(0,0){\includegraphics[width=\unitlength,page=40]{postfilteringAccCpl_combined.pdf}}%
    \put(0.23888962,0.06880806){\color[rgb]{0,0,0}\makebox(0,0)[lt]{\lineheight{1.25}\smash{\begin{tabular}[t]{l}Geom.\end{tabular}}}}%
    \put(0,0){\includegraphics[width=\unitlength,page=41]{postfilteringAccCpl_combined.pdf}}%
    \put(0.23888962,0.05648509){\color[rgb]{0,0,0}\makebox(0,0)[lt]{\lineheight{1.25}\smash{\begin{tabular}[t]{l}DoG\&Geom.\end{tabular}}}}%
    \put(0,0){\includegraphics[width=\unitlength,page=42]{postfilteringAccCpl_combined.pdf}}%
    \put(0.23888962,0.0439327){\color[rgb]{0,0,0}\makebox(0,0)[lt]{\lineheight{1.25}\smash{\begin{tabular}[t]{l}No filter\end{tabular}}}}%
    \put(0,0){\includegraphics[width=\unitlength,page=43]{postfilteringAccCpl_combined.pdf}}%
    \put(0.3747982,0.02335946){\color[rgb]{0,0,0}\makebox(0,0)[lt]{\lineheight{1.25}\smash{\begin{tabular}[t]{l}$30$\end{tabular}}}}%
    \put(0,0){\includegraphics[width=\unitlength,page=44]{postfilteringAccCpl_combined.pdf}}%
    \put(0.41271798,0.02335946){\color[rgb]{0,0,0}\makebox(0,0)[lt]{\lineheight{1.25}\smash{\begin{tabular}[t]{l}$40$\end{tabular}}}}%
    \put(0,0){\includegraphics[width=\unitlength,page=45]{postfilteringAccCpl_combined.pdf}}%
    \put(0.45063777,0.02335946){\color[rgb]{0,0,0}\makebox(0,0)[lt]{\lineheight{1.25}\smash{\begin{tabular}[t]{l}$50$\end{tabular}}}}%
    \put(0,0){\includegraphics[width=\unitlength,page=46]{postfilteringAccCpl_combined.pdf}}%
    \put(0.48855756,0.02335946){\color[rgb]{0,0,0}\makebox(0,0)[lt]{\lineheight{1.25}\smash{\begin{tabular}[t]{l}$60$\end{tabular}}}}%
    \put(0,0){\includegraphics[width=\unitlength,page=47]{postfilteringAccCpl_combined.pdf}}%
    \put(0.52647735,0.02335946){\color[rgb]{0,0,0}\makebox(0,0)[lt]{\lineheight{1.25}\smash{\begin{tabular}[t]{l}$70$\end{tabular}}}}%
    \put(0,0){\includegraphics[width=\unitlength,page=48]{postfilteringAccCpl_combined.pdf}}%
    \put(0.56439714,0.02335946){\color[rgb]{0,0,0}\makebox(0,0)[lt]{\lineheight{1.25}\smash{\begin{tabular}[t]{l}$80$\end{tabular}}}}%
    \put(0,0){\includegraphics[width=\unitlength,page=49]{postfilteringAccCpl_combined.pdf}}%
    \put(0.60231693,0.02335946){\color[rgb]{0,0,0}\makebox(0,0)[lt]{\lineheight{1.25}\smash{\begin{tabular}[t]{l}$90$\end{tabular}}}}%
    \put(0,0){\includegraphics[width=\unitlength,page=50]{postfilteringAccCpl_combined.pdf}}%
    \put(0.63756566,0.02335946){\color[rgb]{0,0,0}\makebox(0,0)[lt]{\lineheight{1.25}\smash{\begin{tabular}[t]{l}$100$\end{tabular}}}}%
    \put(0.45472424,0.00831546){\color[rgb]{0,0,0}\makebox(0,0)[lt]{\lineheight{1.25}\smash{\begin{tabular}[t]{l}\huge{Completeness} \Large{(in \%)}\end{tabular}}}}%
    \put(0,0){\includegraphics[width=\unitlength,page=51]{postfilteringAccCpl_combined.pdf}}%
    \put(0.36456183,0.03144368){\color[rgb]{0,0,0}\makebox(0,0)[lt]{\lineheight{1.25}\smash{\begin{tabular}[t]{l}$30$\end{tabular}}}}%
    \put(0,0){\includegraphics[width=\unitlength,page=52]{postfilteringAccCpl_combined.pdf}}%
    \put(0.36456183,0.05800835){\color[rgb]{0,0,0}\makebox(0,0)[lt]{\lineheight{1.25}\smash{\begin{tabular}[t]{l}$40$\end{tabular}}}}%
    \put(0,0){\includegraphics[width=\unitlength,page=53]{postfilteringAccCpl_combined.pdf}}%
    \put(0.36456183,0.08457303){\color[rgb]{0,0,0}\makebox(0,0)[lt]{\lineheight{1.25}\smash{\begin{tabular}[t]{l}$50$\end{tabular}}}}%
    \put(0,0){\includegraphics[width=\unitlength,page=54]{postfilteringAccCpl_combined.pdf}}%
    \put(0.36456183,0.11113771){\color[rgb]{0,0,0}\makebox(0,0)[lt]{\lineheight{1.25}\smash{\begin{tabular}[t]{l}$60$\end{tabular}}}}%
    \put(0,0){\includegraphics[width=\unitlength,page=55]{postfilteringAccCpl_combined.pdf}}%
    \put(0.36456183,0.13770239){\color[rgb]{0,0,0}\makebox(0,0)[lt]{\lineheight{1.25}\smash{\begin{tabular}[t]{l}$70$\end{tabular}}}}%
    \put(0,0){\includegraphics[width=\unitlength,page=56]{postfilteringAccCpl_combined.pdf}}%
    \put(0.36456183,0.16426707){\color[rgb]{0,0,0}\makebox(0,0)[lt]{\lineheight{1.25}\smash{\begin{tabular}[t]{l}$80$\end{tabular}}}}%
    \put(0,0){\includegraphics[width=\unitlength,page=57]{postfilteringAccCpl_combined.pdf}}%
    \put(0.36456183,0.19083174){\color[rgb]{0,0,0}\makebox(0,0)[lt]{\lineheight{1.25}\smash{\begin{tabular}[t]{l}$90$\end{tabular}}}}%
    \put(0,0){\includegraphics[width=\unitlength,page=58]{postfilteringAccCpl_combined.pdf}}%
    \put(0.35921969,0.21739643){\color[rgb]{0,0,0}\makebox(0,0)[lt]{\lineheight{1.25}\smash{\begin{tabular}[t]{l}$100$\end{tabular}}}}%
    \put(0.35409604,0.08370946){\color[rgb]{0,0,0}\rotatebox{90}{\makebox(0,0)[lt]{\lineheight{1.25}\smash{\begin{tabular}[t]{l}\huge{Accuracy} \Large{(in \%)}\end{tabular}}}}}%
    \put(0,0){\includegraphics[width=\unitlength,page=59]{postfilteringAccCpl_combined.pdf}}%
    \put(0.4912944,0.22478147){\color[rgb]{0,0,0}\makebox(0,0)[lt]{\lineheight{1.25}\smash{\begin{tabular}[t]{l}\huge{\textbf{\snSGM}}\end{tabular}}}}%
    \put(0,0){\includegraphics[width=\unitlength,page=60]{postfilteringAccCpl_combined.pdf}}%
    \put(0.57753736,0.08113103){\color[rgb]{0,0,0}\makebox(0,0)[lt]{\lineheight{1.25}\smash{\begin{tabular}[t]{l}DoG\end{tabular}}}}%
    \put(0,0){\includegraphics[width=\unitlength,page=61]{postfilteringAccCpl_combined.pdf}}%
    \put(0.57753736,0.06880806){\color[rgb]{0,0,0}\makebox(0,0)[lt]{\lineheight{1.25}\smash{\begin{tabular}[t]{l}Geom.\end{tabular}}}}%
    \put(0,0){\includegraphics[width=\unitlength,page=62]{postfilteringAccCpl_combined.pdf}}%
    \put(0.57753736,0.05648509){\color[rgb]{0,0,0}\makebox(0,0)[lt]{\lineheight{1.25}\smash{\begin{tabular}[t]{l}DoG\&Geom.\end{tabular}}}}%
    \put(0,0){\includegraphics[width=\unitlength,page=63]{postfilteringAccCpl_combined.pdf}}%
    \put(0.57753736,0.0439327){\color[rgb]{0,0,0}\makebox(0,0)[lt]{\lineheight{1.25}\smash{\begin{tabular}[t]{l}No filter\end{tabular}}}}%
    \put(0.71344594,0.02335946){\color[rgb]{0,0,0}\makebox(0,0)[lt]{\lineheight{1.25}\smash{\begin{tabular}[t]{l}$30$\end{tabular}}}}%
    \put(0.75136573,0.02335946){\color[rgb]{0,0,0}\makebox(0,0)[lt]{\lineheight{1.25}\smash{\begin{tabular}[t]{l}$40$\end{tabular}}}}%
    \put(0.78928552,0.02335946){\color[rgb]{0,0,0}\makebox(0,0)[lt]{\lineheight{1.25}\smash{\begin{tabular}[t]{l}$50$\end{tabular}}}}%
    \put(0.82720531,0.02335946){\color[rgb]{0,0,0}\makebox(0,0)[lt]{\lineheight{1.25}\smash{\begin{tabular}[t]{l}$60$\end{tabular}}}}%
    \put(0.8651251,0.02335946){\color[rgb]{0,0,0}\makebox(0,0)[lt]{\lineheight{1.25}\smash{\begin{tabular}[t]{l}$70$\end{tabular}}}}%
    \put(0.90304489,0.02335946){\color[rgb]{0,0,0}\makebox(0,0)[lt]{\lineheight{1.25}\smash{\begin{tabular}[t]{l}$80$\end{tabular}}}}%
    \put(0.94096468,0.02335946){\color[rgb]{0,0,0}\makebox(0,0)[lt]{\lineheight{1.25}\smash{\begin{tabular}[t]{l}$90$\end{tabular}}}}%
    \put(0.9762134,0.02335946){\color[rgb]{0,0,0}\makebox(0,0)[lt]{\lineheight{1.25}\smash{\begin{tabular}[t]{l}$100$\end{tabular}}}}%
    \put(0.79337198,0.00831546){\color[rgb]{0,0,0}\makebox(0,0)[lt]{\lineheight{1.25}\smash{\begin{tabular}[t]{l}\huge{Completeness} \Large{(in \%)}\end{tabular}}}}%
    \put(0.70320957,0.03144368){\color[rgb]{0,0,0}\makebox(0,0)[lt]{\lineheight{1.25}\smash{\begin{tabular}[t]{l}$30$\end{tabular}}}}%
    \put(0.70320957,0.05800835){\color[rgb]{0,0,0}\makebox(0,0)[lt]{\lineheight{1.25}\smash{\begin{tabular}[t]{l}$40$\end{tabular}}}}%
    \put(0.70320957,0.08457303){\color[rgb]{0,0,0}\makebox(0,0)[lt]{\lineheight{1.25}\smash{\begin{tabular}[t]{l}$50$\end{tabular}}}}%
    \put(0.70320957,0.11113771){\color[rgb]{0,0,0}\makebox(0,0)[lt]{\lineheight{1.25}\smash{\begin{tabular}[t]{l}$60$\end{tabular}}}}%
    \put(0.70320957,0.13770239){\color[rgb]{0,0,0}\makebox(0,0)[lt]{\lineheight{1.25}\smash{\begin{tabular}[t]{l}$70$\end{tabular}}}}%
    \put(0.70320957,0.16426707){\color[rgb]{0,0,0}\makebox(0,0)[lt]{\lineheight{1.25}\smash{\begin{tabular}[t]{l}$80$\end{tabular}}}}%
    \put(0.70320957,0.19083174){\color[rgb]{0,0,0}\makebox(0,0)[lt]{\lineheight{1.25}\smash{\begin{tabular}[t]{l}$90$\end{tabular}}}}%
    \put(0.69786743,0.21739643){\color[rgb]{0,0,0}\makebox(0,0)[lt]{\lineheight{1.25}\smash{\begin{tabular}[t]{l}$100$\end{tabular}}}}%
    \put(0.69274379,0.08370946){\color[rgb]{0,0,0}\rotatebox{90}{\makebox(0,0)[lt]{\lineheight{1.25}\smash{\begin{tabular}[t]{l}\huge{Accuracy} \Large{(in \%)}\end{tabular}}}}}%
    \put(0.82922112,0.22478147){\color[rgb]{0,0,0}\makebox(0,0)[lt]{\lineheight{1.25}\smash{\begin{tabular}[t]{l}\huge{\textbf{\pgSGM}}\end{tabular}}}}%
    \put(0.9161851,0.08113103){\color[rgb]{0,0,0}\makebox(0,0)[lt]{\lineheight{1.25}\smash{\begin{tabular}[t]{l}DoG\end{tabular}}}}%
    \put(0.9161851,0.06880806){\color[rgb]{0,0,0}\makebox(0,0)[lt]{\lineheight{1.25}\smash{\begin{tabular}[t]{l}Geom.\end{tabular}}}}%
    \put(0.9161851,0.05648509){\color[rgb]{0,0,0}\makebox(0,0)[lt]{\lineheight{1.25}\smash{\begin{tabular}[t]{l}DoG\&Geom.\end{tabular}}}}%
    \put(0.9161851,0.0439327){\color[rgb]{0,0,0}\makebox(0,0)[lt]{\lineheight{1.25}\smash{\begin{tabular}[t]{l}No filter\end{tabular}}}}%
    \end{Large}
  \end{picture}%
\endgroup%

%% file: figures/use-case_qual_results-edited.pdf_tex
%% Accompanies image file 'use-case_qual_results.pdf' (pdf, eps, ps)
\begingroup%
  \makeatletter%
  \providecommand\color[2][]{%
    \errmessage{(Inkscape) Color is used for the text in Inkscape, but the package 'color.sty' is not loaded}%
    \renewcommand\color[2][]{}%
  }%
  \providecommand\transparent[1]{%
    \errmessage{(Inkscape) Transparency is used (non-zero) for the text in Inkscape, but the package 'transparent.sty' is not loaded}%
    \renewcommand\transparent[1]{}%
  }%
  \providecommand\rotatebox[2]{#2}%
  \newcommand*\fsize{\dimexpr\f@size pt\relax}%
  \newcommand*\lineheight[1]{\fontsize{\fsize}{#1\fsize}\selectfont}%
  \ifx\svgwidth\undefined%
    \setlength{\unitlength}{715.51801066bp}%
    \ifx\svgscale\undefined%
      \relax%
    \else%
      \setlength{\unitlength}{\unitlength * \real{\svgscale}}%
    \fi%
  \else%
    \setlength{\unitlength}{\svgwidth}%
  \fi%
  \global\let\svgwidth\undefined%
  \global\let\svgscale\undefined%
  \makeatother%
  \begin{picture}(1,0.65886767)%
    \lineheight{1}%
    \setlength\tabcolsep{0pt}%
    \put(0,0){\includegraphics[width=\unitlength,page=1]{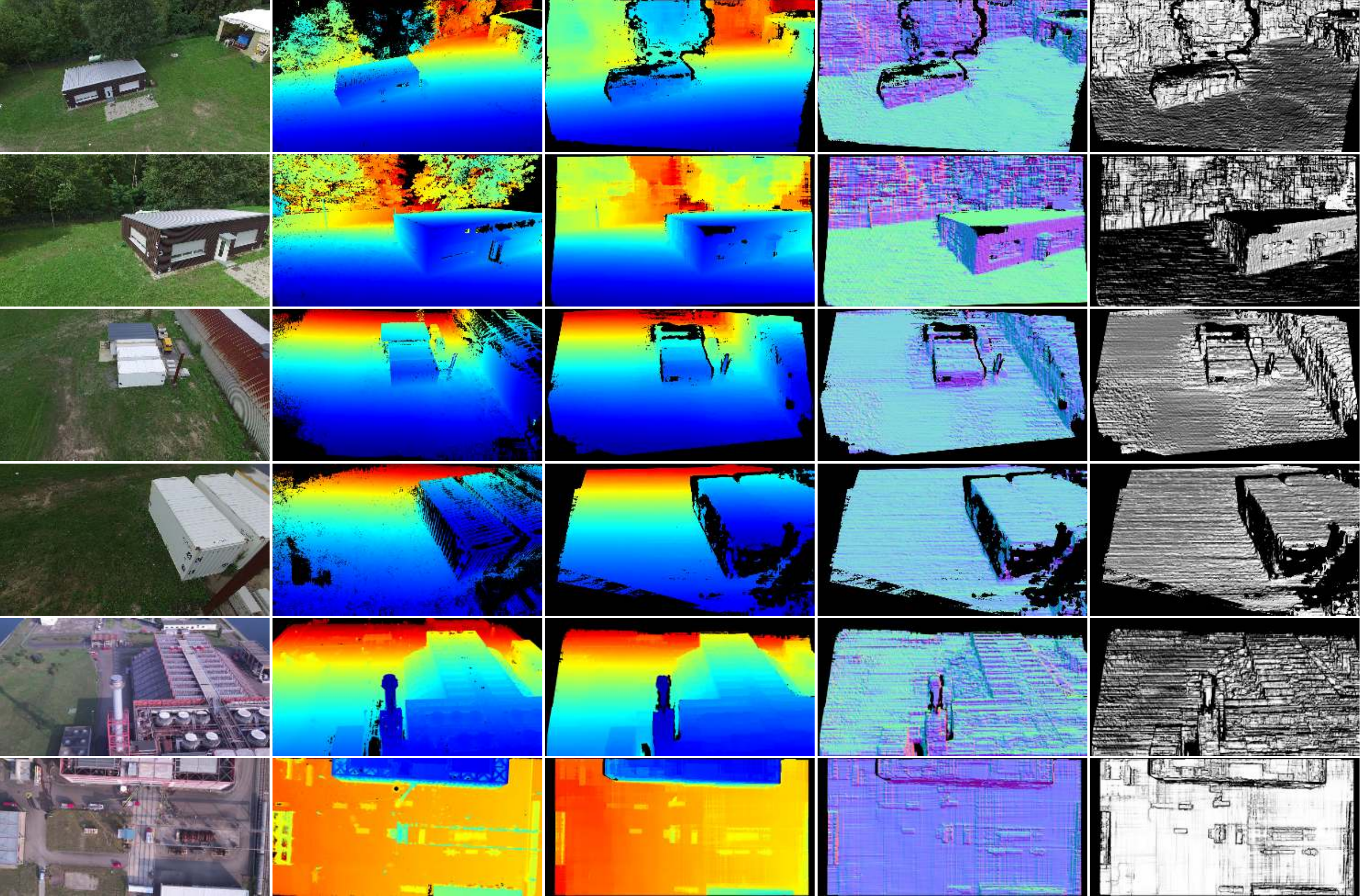}}%
    \put(0.1,0.66881392){\color[rgb]{0,0,0}\makebox(0,0)[ct]{\lineheight{1.25}\smash{\begin{tabular}[t]{c}\textbf{Reference Image}\end{tabular}}}}%
    \put(0.3,0.66881392){\color[rgb]{0,0,0}\makebox(0,0)[ct]{\lineheight{1.25}\smash{\begin{tabular}[t]{c}\textbf{\COLMAP Depth}\end{tabular}}}}%
    \put(0.5,0.66881392){\color[rgb]{0,0,0}\makebox(0,0)[ct]{\lineheight{1.25}\smash{\begin{tabular}[t]{c}\textbf{\pgSGM Depth}\end{tabular}}}}%
    \put(0.7,0.66881392){\color[rgb]{0,0,0}\makebox(0,0)[ct]{\lineheight{1.25}\smash{\begin{tabular}[t]{c}\textbf{\pgSGM Normal}\end{tabular}}}}%
    \put(0.9,0.66956263){\color[rgb]{0,0,0}\makebox(0,0)[ct]{\lineheight{1.25}\smash{\begin{tabular}[t]{c}\textbf{\pgSGM Confidence}\end{tabular}}}}%
  \end{picture}%
\endgroup%

%% file: contents/05_discussion.tex
\section{Discussion}
\label{sec:discussion}

In the following, the findings of the conducted experiments are discussed with respect to different aspects, namely the overall accuracy and the comparison to offline \gls*{MVS} approaches (\Cref{sec:discussion_overallAccuracy}), the ability of the presented approach to reconstruct slanted surface structures (\Cref{sec:discussion_slantedSurfaces}), the run-time and the support for online processing (\Cref{sec:discussion_runtime}), as well as effects of the employed post-filtering algorithms and the relevance of the confidence estimates (\Cref{sec:discussion_postFilteringConfidence}). %

\subsection{Overall Accuracy}
\label{sec:discussion_overallAccuracy}

As the results in \Cref{tab:comparisonToColmap} show, the overall accuracies of the depth maps estimated by the presented approach are lower than those of the geometric depth maps from \COLMAP, and presumably also lower than the results of other offline \gls*{MVS} approaches as the numbers in \Cref{tab:comparisonToOther} suggest. %
This is unsurprising, since the procedure and the assumptions involved greatly differ between online and incremental approaches, such as the one presented in this work, and offline approaches. %
Offline approaches, such as \COLMAP, assume that all input images are available at the time of reconstruction, allowing to optimize the set of input images that are considered for the reconstruction of a certain viewpoint. %
In contrast, online approaches that incrementally perform \gls*{MVS} only consider input images within a temporally confined window, at most all images that were captured up to a certain point in time.  
Furthermore, offline approaches typically also do not have any time constraints. %
Nonetheless, the quantitative differences between the results achieved with the presented approach and \COLMAP are not that big, less than an order of magnitude, especially when using a geometric consistency filtering in post-processing. %
Also, a qualitative comparison on use-case-specific input data renders the results of the presented approach very satisfactory. %
Compared to the geometric depth maps of \COLMAP, the depth maps of \pgSGM lack the fine-grained details, such as the roof structures in Rows $5$~\&~$6$ of \Cref{fig:use-case_qual_results}, which are caused by the coarse-to-fine processing. %
Bigger structures, however, are well represented and the quality of their reconstruction is comparable, as is the overall density. %
Yet, even though the fronto-parallel bias of \gls*{SGM} is reduced, some artifacts of the fronto-parallel sampling are still visible, especially in the normal maps of \Cref{fig:use-case_qual_results}. %

Unfortunately, it is hard to make a quantitative comparison against other online approaches, such as the ones presented by \citet{Gallup2007} or \citet{Pollefeys2008}, with similar assumption, aims and use-cases. %
This is mainly due to the lack of appropriate datasets that are publicly available and provide accurate ground truth that in turn allows for a thorough investigation and benchmarking. %
This holds especially with respect to the task of online and incremental \gls*{MVS}. %
To the best of our knowledge, current publicly available datasets and benchmarks, used to evaluate the performance of algorithms for the task of \gls*{DIM} and \gls*{MVS}, aim at the benchmarking of offline approaches, providing a fixed set of input images acquired from predefined viewpoints. %
However, by our experience, in case of online approaches, the selection and sampling of images from an input sequence, that are used as input to the approach, have a great effect on the final result. %
Depending on the configuration of the input images, the range of the scene depth to be sampled can be very large, requiring higher Gaussian pyramids in order to keep the required resources on a reasonable level and not exceed the run-time constraint. %
However, a higher number of pyramid levels also reduces the initial image size, which in turn will reduce the level of detail and the accuracy of the resulting depth map. %
Thus, a clever assembling of the input bundle is just as important as the correct height of the Gaussian pyramids. %

\subsection{Ability to Account for Non-Fronto-Parallel Surfaces}
\label{sec:discussion_slantedSurfaces}

To further increase the accuracy in the reconstruction of slanted, non-fronto-parallel surface structures, this work proposes, apart from \piSGM, two extensions to the \gls*{SGM} algorithm that should reduce the fronto-parallel bias. %
Namely the incorporation of surface-normals to adjust the zero-cost transition in the \gls*{SGM} path aggregation (\snSGM) and the penalization of deviations from the gradient of the minimum cost path (\pgSGM). %
The conducted experiments reveal that these extensions only provide a slight quantitative improvement over the standard \gls*{SGM} adaptation (\piSGM) to the plane-sweep sampling. %
This insight, however, stands in contrast to the experiments conducted by \citet{Scharstein2018surface}. %
There are at least two reasons that could explain this discrepancy. %
First, \citet{Scharstein2018surface} demonstrate their implementation on a two-view stereo dataset, in which the input images are captured by two cameras mounted on a fixed rig and orientated in the same direction. %
Before being processed, the images are also rectified, \ie transformed so that both lie on the same image plane and that the epipolar lines coincide with the image rows. %
Thus, in the process of \acrlong*{DIM}, the images are equidistantly sampled with a step-size of $1$\,pixel. %
In case of the presented approach however, the distances of the sampling planes and, in turn, the sampling points are chosen in such a way that the disparity shift along the epipolar line between two consecutive planes is less than or equal to $1$. %
This leads to a sampling with a much higher density, already reducing the stair-casing effect in case of \piSGM. %
And secondly, \citet{Scharstein2018surface} propose to use a ground truth normal map for the adjustment of the zero-cost transition, whereas in the presented approach the upscaled normal map of the previous iteration of the hierarchical processing is used. %
This is bootstrapped with \piSGM on the highest pyramid level, introducing inaccuracies, which probably cannot be fully compensated. %
The qualitative analysis, however, reveals that \snSGM and \pgSGM clearly lead to smoother normal maps and that the stair-casing artifacts  in the depth maps are reduced, which is also why in case of the use-case-specific experiments only \pgSGM is considered. %

Apart from reducing the fronto-parallel bias in the \gls*{SGM} path aggregation, the plane-sweep algorithm within our presented approach allows to adjust the image matching to the surface structures in the scene by selecting appropriate normal vectors and sweeping directions. %
In a short qualitative experiment (\cf \Cref{fig:diffPlane_result}), the effects of a horizontal plane sampling, compared to a fronto-parallel sampling, both in combination with \piSGM, are studied. %
The results reveal, that the horizontal sampling leads to more consistent depth estimates with little or no stair-casing artifacts in areas where the surface structure coincides with the plane orientation, \eg the ground plane. %
In areas, where the surface structures are not horizontal, however, the non-fronto-parallel sampling leads to considerable errors. %
To overcome this effect, a splitting of the scene into local regions can be considered, which are sampled individually with different plane orientations, similar to the local-plane-sweep approach presented by \citet{Sinha2014}. %
This, however, again comes at the cost of a higher computational complexity. %
Another remedy is to repeat the plane-sweep image matching multiple times on the whole image domain prior to the \gls*{SGM} optimization, with different sweeping directions and perform a pixel-wise pre-selection of the best plane orientation based on the matching costs, similar to the approach of \citet{Pollefeys2008}. %
This leads to a smaller increase in computational complexity compared to the first option. %

\subsection{Run-Time and Online Processing}
\label{sec:discussion_runtime}

Given the run-time measurements in \Cref{tab:surface-aware_sgmx_runtime} and \Cref{tab:comparisonToColmapRuntime}, the presented approach is obviously not capable of real-time and low-latency processing, in the sense that for each input frame a depth map is computed at similar frame-rates as given for the input stream. 
Considering the nature of the approach and the expected input data, however, the run-time is generally sufficient for online processing, which will be explained in the following section. %
The presented approach takes a bundle of three or more input images, with a bundle size of $5$ images actually yielding better results, and performs \gls*{MVS} for one reference image of the input bundle, which is typically the middle one. %
While these input images could be provided by individual cameras, it is assumed that the images are extracted from an input sequence, which is captured by only one camera that is moving around a static scene. %
In addition, not every image of the input sequence can be used, since an appropriate baseline needs to be in between each input image in order to allow for the estimation of scene depth. %
This is obviously dependent on the depth range that is to be sampled and the scene structure. %
In case of the TMB dataset, the mean distance between the individual input images is $1.8$\m and $1.03$\m for a flight altitude of $15$\m and $8$\m, respectively. %
This increases with higher flight altitudes, due to a larger scene depth. %
Modern \gls*{COTS} rotor-based \glspl*{UAV} can fly up to a speed of above $10$\m/s. %
The typical flight speed when capturing image data, however, is rather $1\text{-}3$\m/s \citep{DJI2020matrice, DJI2020mavic}. %
Thus, if the sets of input images are disjoint, then an estimation needs to be performed at least only every $3$\s, considering a low flight altitude, together with a high flight speed of approximately $3$\m/s and an input bundle size of $3$ images. %
If a maximum overlap between the input bundles is desired, meaning that the estimation of a new depth map is triggered with every new input frame that is appropriate and that it reuses $4$ images from the previous bundle, the required run-time is significantly less. %
As the use-case-specific experiments for the TMB and FB dataset, however, show, the average processing rate of \pgSGM, which is the computationally most complex variant, is between $1$-$2$\Hz, depending on the arrangement of the input images. %
Another possibility to reduce the run-time is the use of higher Gaussian pyramids, which again comes at the cost of a reduced level of detail as already pointed out in the discussion on the overall accuracy (\cf \Cref{sec:discussion_overallAccuracy}). %
In short, there are a number of possible settings in both the acquisition of the input data, \eg regarding the flight speed or size and overlap between the input bundles, and the configuration of the presented approach, \eg regarding the Gaussian pyramid height, depth range or optimization strategy, that allow to tweak the run-time to fit the rate of the input images and, in turn, allow for online processing. 
All in all, the numbers in \Cref{tab:comparisonToColmapRuntime} reveal the superiority of the presented approach in terms of run-time compared to state-of-the-art approaches for offline \acrlong*{MVS}. %

The emergence of high-performance \glspl*{SoC} with embedded \acrshortpl*{GPU}, like the NVIDIA Jetson series, allow to bring approaches like \FASSMVS directly onto the sensor carrier, \eg the \gls*{UAV}, for on-board processing. %
To evaluate the feasibility of running \FASSMVS on-board an embedded device, additionally a few run-time measurements were conducted on the NVIDIA Jetson AGX, equipped with an $8$-core $64$-bit ARMv$8.2$ CPU and a $512$-core Volta GPU. %
On an excerpt of the TMB dataset with in image size of $1920\times1080$\px, \FASSMVS with \piSGM, parameterized by the final configuration as presented in \Cref{sec:experiments_comparisonToColmap}, achieves an average run-time of $727$\ms on the Jetson AGX, compared to an average run-time of approximately $403$\ms achieved on the NVIDIA Titan X. %
As already discussed, the run-time can further be reduced by increasing the pyramid height to $n=4$ and $n=5$, for example, while at the same time accepting a decrease in the quality of the results.
This results in an average run-time of $444$\ms and $385$\ms, respectively. %
These experiments show, that \FASSMVS is capable of on-board processing by utilizing a high-performance embedded \gls*{SoC} like the NVIDIA Jetson AGX. %
This can be of particular interest when considering a deployment on a sensor-carrier that does not suffer from such strong energy constraints as \gls*{COTS} \glspl*{UAV}. %

\subsection{Post-Filtering and the Relevance of the Estimated Confidence Values}
\label{sec:discussion_postFilteringConfidence}

In the following section, the improvements gained by the post-filtering based on the \gls*{DoG} filter and geometric consistency and the effects the filtering has on the online processing, as well as the relevance and the expressiveness of the confidence estimates are discussed. %
A comparison of the \LOne errors in \Cref{tab:comparisonToColmap} to those listed in \Cref{tab:surface-aware_sgmx}, it reveals that with the use of post-filtering based on geometric consistency, the mean errors can drastically be reduced by approximately $40$\percent. 
The prize for this improvement, however, is a loss in density of the depth map and increase in the latency between the input and the results. %
The latter one is due to the additional sliding window that is introduced by the geometric-consistency-based filtering. %
In addition to the bundle of input images, to which only one set of estimates is produced, the geometric filter further requires $2$ or more depth maps for processing. %
Furthermore, the geometric filter is computationally more complex than the \gls*{DoG} filter. %
Again, whether to use the \gls*{DoG} or geometric filter depends on the application. %
If the presented approach is for example used for the task of online $3$D reconstruction, meaning that a subsequent depth map fusion step is employed \citep{Hermann2021realtime}, the geometric-consistency-based filtering is typically done in the fusion of depth maps and can thus be omitted. %
The \gls*{DoG} filter on the other hand is very efficient and does not introduce additional latency. %
However, as already mentioned in the experiments, the \gls*{DoG} filter might also remove potentially good estimates, as it is only executed based on the data provided by the input image. %
Nonetheless, especially when working with input data that contains a lot of homogeneous areas with little to no textures, \eg a clear or cloudy sky in case of extreme oblique viewpoints, the \gls*{DoG} filter is of great benefit. %

Lastly, as a third output, the presented approach computes a confidence map containing pixel-wise confidence scores corresponding to the depth estimates. %
In the scope of this work, these confidence measures are used to perform a comparison between the different \gls*{SGM} extensions based on a \gls*{ROC} analysis (\cf \Cref{fig:sgmx_roc}). %
As already pointed out, the fact that some of the curves are not monotonically increasing suggests that the confidence values do not represent the certainty of the estimates appropriately. %
For one, the reason for this might lie in the normal map. %
Since confidence values are calculated based on the surface orientation stored in the normal map, errors in the normal map inevitably lead to unreliable confidence estimates. %
However, the qualitative excerpts in \Cref{fig:sgmx_DTU_result}, \Cref{fig:sgmx_3DOMcity_result} and \Cref{fig:use-case_qual_results} object this explanation. %
For example, solely the fact that the scene in Row $6$ of \Cref{fig:use-case_qual_results} mostly consists of fronto-parallel structures leads to a confidence map with high certainty values, while the confidence map in Row $2$ in \Cref{fig:use-case_qual_results} renders the estimation of the building roof, which qualitatively appears very accurate, as fully uncertain. %
Similar observation can be seen in the confidence maps depicted in \Cref{fig:sgmx_DTU_result}. %
Only because the ground plane is greatly slanted with respect to the image plane, the confidence of the corresponding estimates is rendered very low even though qualitatively they do not appear more accurate than the estimates on the building facade. %
The most likely reason is that the modeling of a confidence score based on the surface orientation alone is not very expressive. %
An incorporation of additional heuristics, that are based on internal characteristics of the algorithm, as done in previous work \citep{Ruf2019efficient}, might improve the certainty estimation, but this still requires a cumbersome empirical study of the hyper-parameters. %
In recent years, however, the performance of learning-based approaches for the task of confidence estimation \citep{Poggi2020learning, Heinrich2021learning} has greatly increased. %
They are often agnostic to the internals of the algorithm and can be trained on any data for which both estimated and reference depth or disparity maps are available. %

%% file: contents/06_conclusion.tex
\section{Conclusions}
\label{sec:conclusion}

% KAO: Sloppy spacing ensures non-overfull lines. Can be removed if this is not an issue.
\sloppy

\glsreset{MVS}
\glsreset{SGM}

In conclusion, we present an approach for \gls*{MVS} from \gls*{UAV}-borne imagery that allows to facilitate fast, dense and incremental $3$D mapping. %
This approach consists of a hierarchical processing scheme, in which dense depth maps as well as corresponding normal and confidence maps are estimated. %
For the depth map computation, dense multi-image matching, by utilizing the plane-sweep algorithm, is used to produce pixel-wise depth hypotheses.  %
From these hypotheses, a dense depth map is extracted by adopting the optimization scheme of the widely used \gls*{SGM} algorithm. %
Here, the \gls*{SGM} algorithm is not only adapted to work with the multi-image matching of the plane-sweep algorithm, but also extended in order to reduce the fronto-parallel bias and, in turn, also account for slanted surface structure, by introducing two additional regularization schemes. %
The successive normal and confidence map estimation is done separately on the results of the depth estimation. %
In a final filtering step, geometric consistency over multiple depth maps is enforced, which greatly increases the overall accuracy of the resulting depth maps. %

The performance of our approach is quantitatively evaluated on two public datasets containing image data of model-scaled scenes, captured from an aerial perspective and providing an accurate ground truth. %
The experiments show that, for the best configuration, the estimated depth maps have a mean absolute \LOne error of only $8.5$\mm, that is $1$\percent with respect to the maximum depth of the reconstructed scene. %
In comparison, the geometric depth maps from COLMAP, a widely used open-source toolbox for offline \gls*{MVS}, achieve a mean absolute error of $3.8$\mm. %
Thus, even though the presented approach does not have all image data of the input sequence available during the time of reconstruction and is subjected to run-time constraints in order to ensure fast and online processing, its quantitative results are not too far off from state-of-the-art offline approaches. %
While the quantitative results do not show a significant improvement by the presented \gls*{SGM} extensions to account for slanted surface structures, a qualitative comparison reveals their ability to account for non-fronto-parallel surfaces. %
Thus, in case of oblique aerial imagery, which contains a lot of slanted surfaces, the presented \gls*{SGM} extension that penalizes deviations from the gradient of the minimum cost path, \ie \pgSGM, is the best choice, despite its computationally higher complexity. %
Concluding experiments on real-world and use-case-specific datasets have shown, that in terms of run-time the presented approach is well-suited for online processing by achieving a processing rate of $1$-$2$\Hz, meaning that it keeps up with the monocular input stream and allows for incremental $3$D mapping, while the input data is being received. %
A fast $3$D mapping, in turn, can facilitate other important applications or tasks, such as the fast assessment of inaccessible areas by emergency forces, \eg after a flood or earthquake, in order to accomplish disaster relief or search and rescue missions. 

Finally, there are also some aspects to be considered in future work. %
Even though the approach supports different plane orientations in the plane-sweep multi-image matching, to this end, each estimation is done with only one orientation. %
In future, the approach should be extended to use multiple plane orientations within the computation of a single depth map. %
This allows to reconstruct large planar surfaces, such as the ground plane more smoothly, but also allows to maintain higher accuracy in other regions by using fronto-parallel sampling. %
Furthermore, although we state that the processing rate is sufficient, a further decrease in run-time and a more efficient use of GPU resources would make up more opportunities for other concurrent tasks, such as the fusion of depth maps or the generation of orthographic photos. %
Thus, further optimization in terms of run-time and utilization of processing resources is an ongoing task.
Moreover, due to the ongoing development and fast advancements of deep-learning-based approaches for the task of \gls*{MVS}, we want to investigate, whether individual steps or even the whole approach can be substituted by an appropriate learning-based approach, while maintaining the reliability for the use in the scope of critical applications. %
Lastly, in their work, \citet{Nex2011lidar} have shown that the complementary use of \gls*{LIDAR} and image-based techniques for photogrammetric tasks has great potential. %
In addition, with the improvements of \gls*{LIDAR} sensors and the possibility of equipping more and more commodity \glspl*{UAV} with such sensors, like the Zenmuse L$1$\footnote{\url{https://www.dji.com/de/zenmuse-l1}}, their use in order to facilitate fast and incremental $3$D mapping is thus inevitably to be considered in future work. 

%% file: contents/08_appendix.tex
\appendix

\section{The Cross-Ratio as Invariant under Perspective Projection}
\label{sec:crossratio}

\begin{figure}[ht!]%
	\centering
	\def\svgwidth{\columnwidth}
	\resizebox{0.5\columnwidth}{!}{\subimport{figures/}{doppelverhaeltnis-edited.pdf_tex}}%
	\caption{The cross-ratio between four collinear points is the same on all three lines and thus invariant under the perspective projection.}%
	\label{fig:doppelverhaeltnis}%
\end{figure}
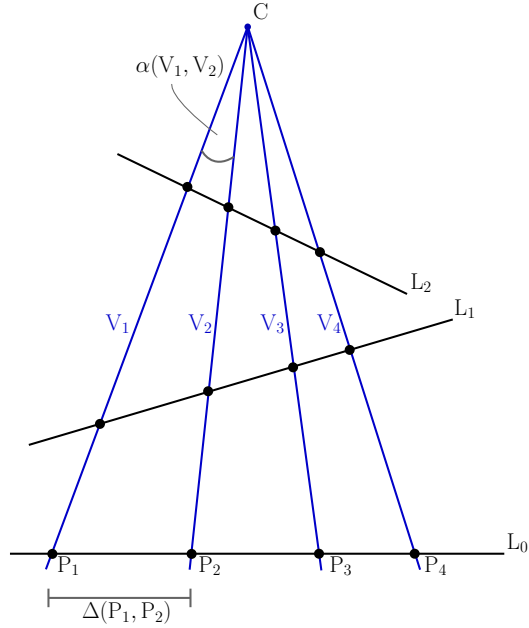

Given four collinear points, the cross-ratio describes the relative distance between them. %
While under perspective projection the relative distances between the points change, their cross-ratio does not and is thus invariant. %
As illustrated by \Cref{fig:doppelverhaeltnis}, four points $\mathrm{P}_1$, $\mathrm{P}_2$, $\mathrm{P}_3$ and $\mathrm{P}_4$ that lie on a straight line $\mathrm{L}_0$ are perspectively projected onto further non-parallel lines, \eg $\mathrm{L}_1$ and $\mathrm{L}_2$. %
If the distance $\Delta(\mathrm{P}_i,\mathrm{P}_j)$ between two points $\mathrm{P}_i$ and $\mathrm{P}_j$ is known, the cross-ratio $Q(\cdot)$ is calculated according to: %
\begin{equation}
\label{eq:doppelverhaeltnisA}
Q(\mathrm{P}_1, \mathrm{P}_2, \mathrm{P}_3, \mathrm{P}_4) = \frac{\Delta(\mathrm{P}_1,\mathrm{P}_3) \cdot \Delta(\mathrm{P}_2,\mathrm{P}_4)}{\Delta(\mathrm{P}_1,\mathrm{P}_4) \cdot \Delta(\mathrm{P}_2,\mathrm{P}_3)} .
\end{equation}
It is the same on all three lines. %
Thus, if the positions of the points on one line are known, their positions on the other lines can be deduced. %
Moreover, given the four rays $\mathrm{V}_1$, $\mathrm{V}_2$, $\mathrm{V}_3$ and $\mathrm{V}_4$, which are going through the center of projection $\mathrm{C}$ and each of the four points, as well as their pairwise enclosed angles $\alpha(\mathrm{V}_i,\mathrm{V}_j)$, the cross-ratio can be extended to:
\begin{equation}
\begin{aligned}
Q(\mathrm{P}_1, \mathrm{P}_2, \mathrm{P}_3, \mathrm{P}_4) &= Q(\mathrm{V}_1, \mathrm{V}_2, \mathrm{V}_3, \mathrm{V}_4) \\ &= \frac{\sin\alpha(\mathrm{V}_1,\mathrm{V}_3) \cdot \sin\alpha(\mathrm{V}_2,\mathrm{V}_4)}{\sin\alpha(\mathrm{V}_1,\mathrm{V}_4) \cdot \sin\alpha(\mathrm{V}_2,\mathrm{V}_3)} .
\end{aligned}
\label{eq:doppelverhaeltnisB}
\end{equation}

%% file: figures/doppelverhaeltnis-edited.pdf_tex
%% Accompanies image file 'doppelverhaeltnis.pdf' (pdf, eps, ps)
\begingroup%
  \makeatletter%
  \providecommand\color[2][]{%
    \errmessage{(Inkscape) Color is used for the text in Inkscape, but the package 'color.sty' is not loaded}%
    \renewcommand\color[2][]{}%
  }%
  \providecommand\transparent[1]{%
    \errmessage{(Inkscape) Transparency is used (non-zero) for the text in Inkscape, but the package 'transparent.sty' is not loaded}%
    \renewcommand\transparent[1]{}%
  }%
  \providecommand\rotatebox[2]{#2}%
  \newcommand*\fsize{\dimexpr\f@size pt\relax}%
  \newcommand*\lineheight[1]{\fontsize{\fsize}{#1\fsize}\selectfont}%
  \ifx\svgwidth\undefined%
    \setlength{\unitlength}{200.89450836bp}%
    \ifx\svgscale\undefined%
      \relax%
    \else%
      \setlength{\unitlength}{\unitlength * \real{\svgscale}}%
    \fi%
  \else%
    \setlength{\unitlength}{\svgwidth}%
  \fi%
  \global\let\svgwidth\undefined%
  \global\let\svgscale\undefined%
  \makeatother%
  \begin{picture}(1,1.21244757)%
    \lineheight{1}%
    \setlength\tabcolsep{0pt}%    
    \begin{Large}
    \put(0,0){\includegraphics[width=\unitlength,page=1]{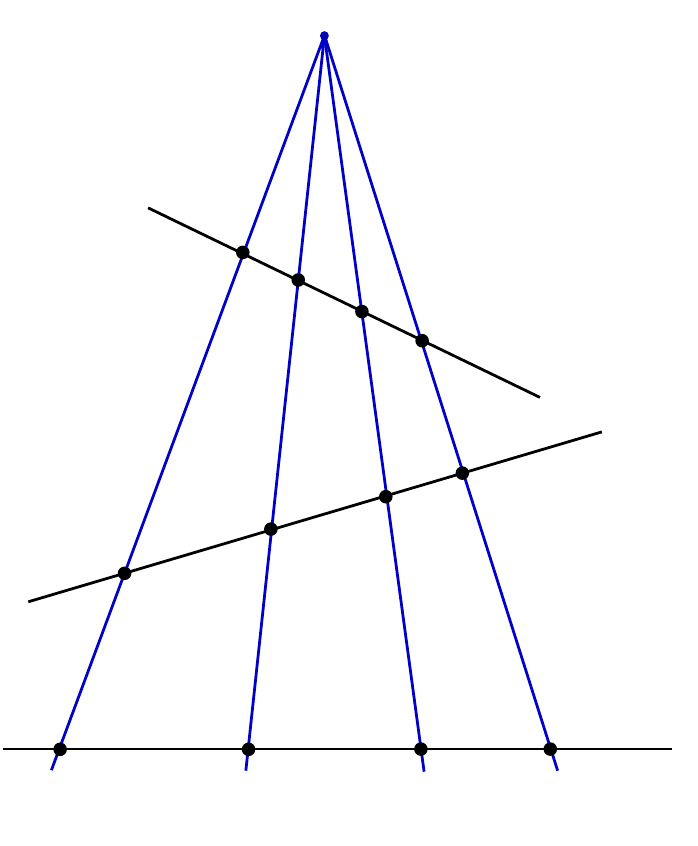}}%
    \put(0.09544961,0.103){\color[rgb]{0,0,0}\makebox(0,0)[lt]{\lineheight{1.25}\smash{\begin{tabular}[t]{l}$\mathrm{P}_1$\end{tabular}}}}%
    \put(0.36970176,0.103){\color[rgb]{0,0,0}\makebox(0,0)[lt]{\lineheight{1.25}\smash{\begin{tabular}[t]{l}$\mathrm{P}_2$\end{tabular}}}}%
    \put(0.62356636,0.103){\color[rgb]{0,0,0}\makebox(0,0)[lt]{\lineheight{1.25}\smash{\begin{tabular}[t]{l}$\mathrm{P}_3$\end{tabular}}}}%
    \put(0.80774257,0.103){\color[rgb]{0,0,0}\makebox(0,0)[lt]{\lineheight{1.25}\smash{\begin{tabular}[t]{l}$\mathrm{P}_4$\end{tabular}}}}%
    \put(0.96783808,0.14771485){\color[rgb]{0,0,0}\makebox(0,0)[lt]{\lineheight{1.25}\smash{\begin{tabular}[t]{l}$\mathrm{L}_0$\end{tabular}}}}%
    \put(0.86590727,0.60204929){\color[rgb]{0,0,0}\makebox(0,0)[lt]{\lineheight{1.25}\smash{\begin{tabular}[t]{l}$\mathrm{L}_1$\end{tabular}}}}%
    \put(0.78070241,0.65240109){\color[rgb]{0,0,0}\makebox(0,0)[lt]{\lineheight{1.25}\smash{\begin{tabular}[t]{l}$\mathrm{L}_2$\end{tabular}}}}%
    \put(0,0){\includegraphics[width=\unitlength,page=2]{doppelverhaeltnis.pdf}}%
    \put(0.14415754,0.01356336){\color[rgb]{0,0,0}\makebox(0,0)[lt]{\lineheight{1.25}\smash{\begin{tabular}[t]{l}$\Delta(\mathrm{P}_1,\mathrm{P}_2)$\end{tabular}}}}%
    \put(0,0){\includegraphics[width=\unitlength,page=3]{doppelverhaeltnis.pdf}}%
    \put(0.24839621,1.07337958){\color[rgb]{0,0,0}\makebox(0,0)[lt]{\lineheight{1.25}\smash{\begin{tabular}[t]{l}$\alpha(\mathrm{V}_1,\mathrm{V}_2)$\end{tabular}}}}%
    \put(0,0){\includegraphics[width=\unitlength,page=4]{doppelverhaeltnis.pdf}}%
    \put(0.47547368,1.17656284){\color[rgb]{0,0,0}\makebox(0,0)[lt]{\lineheight{1.25}\smash{\begin{tabular}[t]{l}$\mathrm{C}$\end{tabular}}}}%
    \put(0.19,0.57364528){\color[rgb]{0,0,0.78431373}\makebox(0,0)[lt]{\lineheight{1.25}\smash{\begin{tabular}[t]{l}$\mathrm{V}_1$\end{tabular}}}}%
    \put(0.35,0.57364528){\color[rgb]{0,0,0.78431373}\makebox(0,0)[lt]{\lineheight{1.25}\smash{\begin{tabular}[t]{l}$\mathrm{V}_2$\end{tabular}}}}% 
    \put(0.49,0.57364528){\color[rgb]{0,0,0.78431373}\makebox(0,0)[lt]{\lineheight{1.25}\smash{\begin{tabular}[t]{l}$\mathrm{V}_3$\end{tabular}}}}% 
    \put(0.60,0.57364528){\color[rgb]{0,0,0.78431373}\makebox(0,0)[lt]{\lineheight{1.25}\smash{\begin{tabular}[t]{l}$\mathrm{V}_4$\end{tabular}}}}%     
    \end{Large}
  \end{picture}%
\endgroup%

%% file: multiviewsgm.bbl
\begin{thebibliography}{73}
\expandafter\ifx\csname natexlab\endcsname\relax\def\natexlab#1{#1}\fi
\providecommand{\url}[1]{\texttt{#1}}
\providecommand{\href}[2]{#2}
\providecommand{\path}[1]{#1}
\providecommand{\DOIprefix}{doi:}
\providecommand{\ArXivprefix}{arXiv:}
\providecommand{\URLprefix}{URL: }
\providecommand{\Pubmedprefix}{pmid:}
\providecommand{\doi}[1]{\href{http://dx.doi.org/#1}{\path{#1}}}
\providecommand{\Pubmed}[1]{\href{pmid:#1}{\path{#1}}}
\providecommand{\bibinfo}[2]{#2}
\ifx\xfnm\relax \def\xfnm[#1]{\unskip,\space#1}\fi
%Type = Article
\bibitem[{Aan{\ae}s et~al.(2016)Aan{\ae}s, Jensen, Vogiatzis, Tola and
  Dahl}]{Aanaes2016Large}
\bibinfo{author}{Aan{\ae}s, H.}, \bibinfo{author}{Jensen, R.R.},
  \bibinfo{author}{Vogiatzis, G.}, \bibinfo{author}{Tola, E.},
  \bibinfo{author}{Dahl, A.B.}, \bibinfo{year}{2016}.
\newblock \bibinfo{title}{Large-scale data for multiple-view stereopsis}.
\newblock \bibinfo{journal}{International Journal of Computer Vision}
  \bibinfo{volume}{120}, \bibinfo{pages}{153--168}.
%Type = Inproceedings
\bibitem[{Banz et~al.(2011)Banz, Blume and Pirsch}]{Banz2011real}
\bibinfo{author}{Banz, C.}, \bibinfo{author}{Blume, H.},
  \bibinfo{author}{Pirsch, P.}, \bibinfo{year}{2011}.
\newblock \bibinfo{title}{Real-time semi-global matching disparity estimation
  on the {GPU}}, in: \bibinfo{booktitle}{Proceedings of the IEEE International
  Conference on Computer Vision Workshops}, pp. \bibinfo{pages}{514--521}.
%Type = Inproceedings
\bibitem[{Banz et~al.(2010)Banz, Hesselbarth, Flatt, Blume and
  Pirsch}]{Banz2010realtime}
\bibinfo{author}{Banz, C.}, \bibinfo{author}{Hesselbarth, S.},
  \bibinfo{author}{Flatt, H.}, \bibinfo{author}{Blume, H.},
  \bibinfo{author}{Pirsch, P.}, \bibinfo{year}{2010}.
\newblock \bibinfo{title}{Real-time stereo vision system using semi-global
  matching disparity estimation: Architecture and {FPGA}-implementation}, in:
  \bibinfo{booktitle}{Proceedings of the International Conference on Embedded
  Computer Systems: Architectures, Modeling and Simulation}, pp.
  \bibinfo{pages}{93--101}.
%Type = Inproceedings
\bibitem[{Barry et~al.(2015)Barry, Oleynikova, Honegger, Pollefeys and
  Tedrake}]{Barry2015fpga}
\bibinfo{author}{Barry, A.J.}, \bibinfo{author}{Oleynikova, H.},
  \bibinfo{author}{Honegger, D.}, \bibinfo{author}{Pollefeys, M.},
  \bibinfo{author}{Tedrake, R.}, \bibinfo{year}{2015}.
\newblock \bibinfo{title}{{FPGA} vs. pushbroom stereo vision for {MAVs}}, in:
  \bibinfo{booktitle}{Proceedings of the IROS Workshop on Vision-based Control
  and Navigation of Small Lightweight UAVs}.
%Type = Inproceedings
\bibitem[{Cheng et~al.(2020)Cheng, Xu, Zhu, Li, Li, Ramamoorthi and
  Su}]{Cheng2020deep}
\bibinfo{author}{Cheng, S.}, \bibinfo{author}{Xu, Z.}, \bibinfo{author}{Zhu,
  S.}, \bibinfo{author}{Li, Z.}, \bibinfo{author}{Li, L.E.},
  \bibinfo{author}{Ramamoorthi, R.}, \bibinfo{author}{Su, H.},
  \bibinfo{year}{2020}.
\newblock \bibinfo{title}{Deep stereo using adaptive thin volume representation
  with uncertainty awareness}, in: \bibinfo{booktitle}{Proceedings of the
  IEEE/CVF Conference on Computer Vision and Pattern Recognition}, pp.
  \bibinfo{pages}{2524--2534}.
%Type = Inproceedings
\bibitem[{Collins(1996)}]{Collins1996space}
\bibinfo{author}{Collins, R.T.}, \bibinfo{year}{1996}.
\newblock \bibinfo{title}{A space-sweep approach to true multi-image matching},
  in: \bibinfo{booktitle}{Proceedings of the IEEE Conference on Computer Vision
  and Pattern Recognition}, pp. \bibinfo{pages}{358--363}.
%Type = Article
\bibitem[{Davison et~al.(2007)Davison, Reid, Molton and
  Stasse}]{Davison2007monoslam}
\bibinfo{author}{Davison, A.J.}, \bibinfo{author}{Reid, I.D.},
  \bibinfo{author}{Molton, N.D.}, \bibinfo{author}{Stasse, O.},
  \bibinfo{year}{2007}.
\newblock \bibinfo{title}{{MonoSLAM}: Real-time single camera {SLAM}}.
\newblock \bibinfo{journal}{IEEE Transactions on Pattern Analysis and Machine
  Intelligence} \bibinfo{volume}{29}, \bibinfo{pages}{1052--1067}.
%Type = Book
\bibitem[{DJI(2020a)}]{DJI2020mavic}
\bibinfo{author}{DJI}, \bibinfo{year}{2020}a.
\newblock \bibinfo{title}{Matrice 2 Pro/Zoom - User Manual}.
%Type = Book
\bibitem[{DJI(2020b)}]{DJI2020matrice}
\bibinfo{author}{DJI}, \bibinfo{year}{2020}b.
\newblock \bibinfo{title}{Matrice 200 V2-Series - User Manual}.
%Type = Inproceedings
\bibitem[{Drory et~al.(2014)Drory, Haubold, Avidan and
  Hamprecht}]{Drory2014semi}
\bibinfo{author}{Drory, A.}, \bibinfo{author}{Haubold, C.},
  \bibinfo{author}{Avidan, S.}, \bibinfo{author}{Hamprecht, F.A.},
  \bibinfo{year}{2014}.
\newblock \bibinfo{title}{Semi-global matching: a principled derivation in
  terms of message passing}, in: \bibinfo{booktitle}{Proceedings of the German
  Conference on Pattern Recognition}, pp. \bibinfo{pages}{43--53}.
%Type = Inproceedings
\bibitem[{Eade and Drummond(2006)}]{Eade2006scalable}
\bibinfo{author}{Eade, E.}, \bibinfo{author}{Drummond, T.},
  \bibinfo{year}{2006}.
\newblock \bibinfo{title}{Scalable monocular {SLAM}}, in:
  \bibinfo{booktitle}{Proceedings of the IEEE Computer Society Conference on
  Computer Vision and Pattern Recognition}, pp. \bibinfo{pages}{469--476}.
%Type = Inproceedings
\bibitem[{Furukawa et~al.(2009)Furukawa, Curless, Seitz and
  Szeliski}]{Furukawa2009}
\bibinfo{author}{Furukawa, Y.}, \bibinfo{author}{Curless, B.},
  \bibinfo{author}{Seitz, S.M.}, \bibinfo{author}{Szeliski, R.},
  \bibinfo{year}{2009}.
\newblock \bibinfo{title}{Manhattan-world stereo}, in:
  \bibinfo{booktitle}{Proceedings of the IEEE Conference on Computer Vision and
  Pattern Recognition}, pp. \bibinfo{pages}{1422--1429}.
%Type = Article
\bibitem[{Furukawa and Ponce(2010)}]{Furukawa2010pmvs}
\bibinfo{author}{Furukawa, Y.}, \bibinfo{author}{Ponce, J.},
  \bibinfo{year}{2010}.
\newblock \bibinfo{title}{Accurate, dense, and robust multiview stereopsis}.
\newblock \bibinfo{journal}{IEEE Transactions on Pattern Analysis and Machine
  Intelligence} \bibinfo{volume}{32}, \bibinfo{pages}{1362--1376}.
%Type = Incollection
\bibitem[{Furutani and Minami(2021)}]{Furutani2021drones}
\bibinfo{author}{Furutani, T.}, \bibinfo{author}{Minami, M.},
  \bibinfo{year}{2021}.
\newblock \bibinfo{title}{Drones for disaster risk reduction and crisis
  response}, in: \bibinfo{booktitle}{Emerging Technologies for Disaster
  Resilience}, pp. \bibinfo{pages}{51--62}.
%Type = Inproceedings
\bibitem[{Galliani et~al.(2015)Galliani, Lasinger and Schindler}]{Galliani2015}
\bibinfo{author}{Galliani, S.}, \bibinfo{author}{Lasinger, K.},
  \bibinfo{author}{Schindler, K.}, \bibinfo{year}{2015}.
\newblock \bibinfo{title}{Massively parallel multiview stereopsis by surface
  normal diffusion}, in: \bibinfo{booktitle}{Proceedings of the IEEE
  International Conference on Computer Vision}, pp. \bibinfo{pages}{873--881}.
%Type = Inproceedings
\bibitem[{Gallup et~al.(2007)Gallup, Frahm, Mordohai, Yang and
  Pollefeys}]{Gallup2007}
\bibinfo{author}{Gallup, D.}, \bibinfo{author}{Frahm, J.M.},
  \bibinfo{author}{Mordohai, P.}, \bibinfo{author}{Yang, Q.},
  \bibinfo{author}{Pollefeys, M.}, \bibinfo{year}{2007}.
\newblock \bibinfo{title}{Real-time plane-sweeping stereo with multiple
  sweeping directions}, in: \bibinfo{booktitle}{Proceedings of the IEEE
  Conference on Computer Vision and Pattern Recognition}.
%Type = Inproceedings
\bibitem[{Gallup et~al.(2010)Gallup, Frahm and Pollefeys}]{Gallup2010piece}
\bibinfo{author}{Gallup, D.}, \bibinfo{author}{Frahm, J.M.},
  \bibinfo{author}{Pollefeys, M.}, \bibinfo{year}{2010}.
\newblock \bibinfo{title}{Piecewise planar and non-planar stereo for urban
  scene reconstruction}, in: \bibinfo{booktitle}{Proceedings of the IEEE
  Conference on Computer Vision and Pattern Recognition}, pp.
  \bibinfo{pages}{1418--1425}.
%Type = Inproceedings
\bibitem[{Goesele et~al.(2007)Goesele, Snavely, Curless, Hoppe and
  Seitz}]{Goesele2007multi}
\bibinfo{author}{Goesele, M.}, \bibinfo{author}{Snavely, N.},
  \bibinfo{author}{Curless, B.}, \bibinfo{author}{Hoppe, H.},
  \bibinfo{author}{Seitz, S.M.}, \bibinfo{year}{2007}.
\newblock \bibinfo{title}{Multi-view stereo for community photo collections},
  in: \bibinfo{booktitle}{Proceedings of the IEEE International Conference on
  Computer Vision}.
%Type = Inproceedings
\bibitem[{Gu et~al.(2020)Gu, Fan, Zhu, Dai, Tan and Tan}]{Gu2020cascade}
\bibinfo{author}{Gu, X.}, \bibinfo{author}{Fan, Z.}, \bibinfo{author}{Zhu, S.},
  \bibinfo{author}{Dai, Z.}, \bibinfo{author}{Tan, F.}, \bibinfo{author}{Tan,
  P.}, \bibinfo{year}{2020}.
\newblock \bibinfo{title}{Cascade cost volume for high-resolution multi-view
  stereo and stereo matching}, in: \bibinfo{booktitle}{Proceedings of the
  IEEE/CVF Conference on Computer Vision and Pattern Recognition}, pp.
  \bibinfo{pages}{2495--2504}.
%Type = Inproceedings
\bibitem[{Haala et~al.(2015)Haala, Rothermel and Cavegn}]{Haala2015}
\bibinfo{author}{Haala, N.}, \bibinfo{author}{Rothermel, M.},
  \bibinfo{author}{Cavegn, S.}, \bibinfo{year}{2015}.
\newblock \bibinfo{title}{Extracting 3d urban models from oblique aerial
  images}, in: \bibinfo{booktitle}{Proceedings of the IEEE Joint Urban Remote
  Sensing Event}, \bibinfo{organization}{IEEE}.
%Type = Inproceedings
\bibitem[{Han et~al.(2015)Han, Leung, Jia, Sukthankar and
  Berg}]{Han2015matchnet}
\bibinfo{author}{Han, X.}, \bibinfo{author}{Leung, T.}, \bibinfo{author}{Jia,
  Y.}, \bibinfo{author}{Sukthankar, R.}, \bibinfo{author}{Berg, A.C.},
  \bibinfo{year}{2015}.
\newblock \bibinfo{title}{{MatchNet}: Unifying feature and metric learning for
  patch-based matching}, in: \bibinfo{booktitle}{Proceedings of the IEEE
  Conference on Computer Vision and Pattern Recognition}, pp.
  \bibinfo{pages}{3279--3286}.
%Type = Book
\bibitem[{Hartley and Zisserman(2004)}]{Hartley2003}
\bibinfo{author}{Hartley, R.}, \bibinfo{author}{Zisserman, A.},
  \bibinfo{year}{2004}.
\newblock \bibinfo{title}{Multiple view geometry in computer vision}.
\newblock \bibinfo{publisher}{Cambridge University Press}.
%Type = Inproceedings
\bibitem[{Hartmann et~al.(2017)Hartmann, Galliani, Havlena, Van~Gool and
  Schindler}]{Hartmann2017learned}
\bibinfo{author}{Hartmann, W.}, \bibinfo{author}{Galliani, S.},
  \bibinfo{author}{Havlena, M.}, \bibinfo{author}{Van~Gool, L.},
  \bibinfo{author}{Schindler, K.}, \bibinfo{year}{2017}.
\newblock \bibinfo{title}{Learned multi-patch similarity}, in:
  \bibinfo{booktitle}{Proceedings of the IEEE International Conference on
  Computer Vision}, pp. \bibinfo{pages}{1586--1594}.
%Type = Article
\bibitem[{Heinrich and Mehltretter(2021)}]{Heinrich2021learning}
\bibinfo{author}{Heinrich, K.}, \bibinfo{author}{Mehltretter, M.},
  \bibinfo{year}{2021}.
\newblock \bibinfo{title}{Learning multi-modal features for dense
  matching-based confidence estimation}.
\newblock \bibinfo{journal}{The International Archives of the Photogrammetry,
  Remote Sensing and Spatial Information Sciences}
  \bibinfo{volume}{XLIII-B2-2021}, \bibinfo{pages}{91--99}.
%Type = Article
\bibitem[{Hermann et~al.(2021)Hermann, Ruf and Weinmann}]{Hermann2021realtime}
\bibinfo{author}{Hermann, M.}, \bibinfo{author}{Ruf, B.},
  \bibinfo{author}{Weinmann, M.}, \bibinfo{year}{2021}.
\newblock \bibinfo{title}{Real-time dense 3d reconstruction from monocular
  video data captured by low-cost {UAVs}}.
\newblock \bibinfo{journal}{The International Archives of the Photogrammetry,
  Remote Sensing and Spatial Information Sciences}
  \bibinfo{volume}{{XLIII}-B2-2021}, \bibinfo{pages}{361--368}.
%Type = Inproceedings
\bibitem[{Hermann et~al.(2009)Hermann, Klette and
  Destefanis}]{Hermann2009inclusion}
\bibinfo{author}{Hermann, S.}, \bibinfo{author}{Klette, R.},
  \bibinfo{author}{Destefanis, E.}, \bibinfo{year}{2009}.
\newblock \bibinfo{title}{Inclusion of a second-order prior into semi-global
  matching}, in: \bibinfo{booktitle}{Proceedings of the Pacific-Rim Symposium
  on Image and Video Technology}, pp. \bibinfo{pages}{633--644}.
%Type = Article
\bibitem[{Hernandez-Juarez et~al.(2016)Hernandez-Juarez, Chac{\'o}n, Espinosa,
  V{\'a}zquez, Moure and L{\'o}pez}]{Hernandez2016embedded}
\bibinfo{author}{Hernandez-Juarez, D.}, \bibinfo{author}{Chac{\'o}n, A.},
  \bibinfo{author}{Espinosa, A.}, \bibinfo{author}{V{\'a}zquez, D.},
  \bibinfo{author}{Moure, J.C.}, \bibinfo{author}{L{\'o}pez, A.M.},
  \bibinfo{year}{2016}.
\newblock \bibinfo{title}{Embedded real-time stereo estimation via semi-global
  matching on the gpu}.
\newblock \bibinfo{journal}{Procedia Computer Science} \bibinfo{volume}{80},
  \bibinfo{pages}{143--153}.
%Type = Inproceedings
\bibitem[{Hirschm\"{u}ller(2005)}]{Hirschmueller2005}
\bibinfo{author}{Hirschm\"{u}ller, H.}, \bibinfo{year}{2005}.
\newblock \bibinfo{title}{Accurate and efficient stereo processing by
  semi-global matching and mutual information}, in:
  \bibinfo{booktitle}{Proceedings of the IEEE Conference on Computer Vision and
  Pattern Recognition}, pp. \bibinfo{pages}{807--814}.
%Type = Article
\bibitem[{Hirschm\"{u}ller(2008)}]{Hirschmueller2008}
\bibinfo{author}{Hirschm\"{u}ller, H.}, \bibinfo{year}{2008}.
\newblock \bibinfo{title}{Stereo processing by semiglobal matching and mutual
  information}.
\newblock \bibinfo{journal}{IEEE Transactions on Pattern Analysis and Machine
  Intelligence} \bibinfo{volume}{30}, \bibinfo{pages}{328--341}.
%Type = Inproceedings
\bibitem[{Huang et~al.(2021)Huang, Yi, Huang, He, Liu and
  Liu}]{Huang2021m3vsnet}
\bibinfo{author}{Huang, B.}, \bibinfo{author}{Yi, H.}, \bibinfo{author}{Huang,
  C.}, \bibinfo{author}{He, Y.}, \bibinfo{author}{Liu, J.},
  \bibinfo{author}{Liu, X.}, \bibinfo{year}{2021}.
\newblock \bibinfo{title}{{M3VSNET}: Unsupervised multi-metric multi-view
  stereo network}, in: \bibinfo{booktitle}{Proceedings of the IEEE
  International Conference on Image Processing}, pp.
  \bibinfo{pages}{3163--3167}.
%Type = Inproceedings
\bibitem[{Huang et~al.(2018)Huang, Matzen, Kopf, Ahuja and
  Huang}]{Huang2018deepmvs}
\bibinfo{author}{Huang, P.H.}, \bibinfo{author}{Matzen, K.},
  \bibinfo{author}{Kopf, J.}, \bibinfo{author}{Ahuja, N.},
  \bibinfo{author}{Huang, J.B.}, \bibinfo{year}{2018}.
\newblock \bibinfo{title}{{DeepMVS}: Learning multi-view stereopsis}, in:
  \bibinfo{booktitle}{Proceedings of the IEEE Conference on Computer Vision and
  Pattern Recognition}, pp. \bibinfo{pages}{2821--2830}.
%Type = Inproceedings
\bibitem[{Jensen et~al.(2014)Jensen, Dahl, Vogiatzis, Tola and
  Aan{\ae}s}]{Jensen2014dtu}
\bibinfo{author}{Jensen, R.}, \bibinfo{author}{Dahl, A.},
  \bibinfo{author}{Vogiatzis, G.}, \bibinfo{author}{Tola, E.},
  \bibinfo{author}{Aan{\ae}s, H.}, \bibinfo{year}{2014}.
\newblock \bibinfo{title}{Large scale multi-view stereopsis evaluation}, in:
  \bibinfo{booktitle}{Proceedings of the IEEE Conference on Computer Vision and
  Pattern Recognition}, pp. \bibinfo{pages}{406--413}.
%Type = Inproceedings
\bibitem[{Ji et~al.(2017)Ji, Gall, Zheng, Liu and Fang}]{Ji2017surfacenet}
\bibinfo{author}{Ji, M.}, \bibinfo{author}{Gall, J.}, \bibinfo{author}{Zheng,
  H.}, \bibinfo{author}{Liu, Y.}, \bibinfo{author}{Fang, L.},
  \bibinfo{year}{2017}.
\newblock \bibinfo{title}{{SurfaceNet}: An end-to-end 3d neural network for
  multiview stereopsis}, in: \bibinfo{booktitle}{Proceedings of the IEEE
  International Conference on Computer Vision}, pp.
  \bibinfo{pages}{2307--2315}.
%Type = Inproceedings
\bibitem[{Kang et~al.(2001)Kang, Szeliski and Chai}]{Kang2001}
\bibinfo{author}{Kang, S.B.}, \bibinfo{author}{Szeliski, R.},
  \bibinfo{author}{Chai, J.}, \bibinfo{year}{2001}.
\newblock \bibinfo{title}{Handling occlusions in dense multi-view stereo}, in:
  \bibinfo{booktitle}{Proceedings of the IEEE Conference on Computer Vision and
  Pattern Recognition}, pp. \bibinfo{pages}{103--110}.
%Type = Inproceedings
\bibitem[{Khot et~al.(2019)Khot, Agrawal, Tulsiani, Mertz, Lucey and
  Hebert}]{Khot2019learning}
\bibinfo{author}{Khot, T.}, \bibinfo{author}{Agrawal, S.},
  \bibinfo{author}{Tulsiani, S.}, \bibinfo{author}{Mertz, C.},
  \bibinfo{author}{Lucey, S.}, \bibinfo{author}{Hebert, M.},
  \bibinfo{year}{2019}.
\newblock \bibinfo{title}{Learning unsupervised multi-view stereopsis via
  robust photometric consistency}, in: \bibinfo{booktitle}{Proceedings of the
  IEEE Conference on Computer Vision and Pattern Recognition Workshops}.
%Type = Inproceedings
\bibitem[{Klein and Murray(2007)}]{Klein2007ptam}
\bibinfo{author}{Klein, G.}, \bibinfo{author}{Murray, D.},
  \bibinfo{year}{2007}.
\newblock \bibinfo{title}{Parallel tracking and mapping for small {AR}
  workspaces}, in: \bibinfo{booktitle}{Proceedings of the IEEE and ACM
  International Symposium on Mixed and Augmented Reality}, pp.
  \bibinfo{pages}{225--234}.
%Type = Article
\bibitem[{Knapitsch et~al.(2017)Knapitsch, Park, Zhou and
  Koltun}]{Knapitsch2017tanks}
\bibinfo{author}{Knapitsch, A.}, \bibinfo{author}{Park, J.},
  \bibinfo{author}{Zhou, Q.Y.}, \bibinfo{author}{Koltun, V.},
  \bibinfo{year}{2017}.
\newblock \bibinfo{title}{Tanks and temples: Benchmarking large-scale scene
  reconstruction}.
\newblock \bibinfo{journal}{ACM Transactions on Graphics} \bibinfo{volume}{36},
  \bibinfo{pages}{78}.
%Type = Inproceedings
\bibitem[{Kolev et~al.(2014)Kolev, Tanskanen, Speciale and
  Pollefeys}]{Kolev2014}
\bibinfo{author}{Kolev, K.}, \bibinfo{author}{Tanskanen, P.},
  \bibinfo{author}{Speciale, P.}, \bibinfo{author}{Pollefeys, M.},
  \bibinfo{year}{2014}.
\newblock \bibinfo{title}{Turning mobile phones into 3d scanners}, in:
  \bibinfo{booktitle}{Proceedings of the IEEE Conference on Computer Vision and
  Pattern Recognition}, pp. \bibinfo{pages}{3946--3953}.
%Type = Inproceedings
\bibitem[{Kuschk and Cremers(2013)}]{Kuschk2013}
\bibinfo{author}{Kuschk, G.}, \bibinfo{author}{Cremers, D.},
  \bibinfo{year}{2013}.
\newblock \bibinfo{title}{Fast and accurate large-scale stereo reconstruction
  using variational methods}, in: \bibinfo{booktitle}{Proceedings of the IEEE
  International Conference on Computer Vision Workshops}, pp.
  \bibinfo{pages}{700--707}.
%Type = Inproceedings
\bibitem[{Newcombe and Davison(2010)}]{Newcombe2010}
\bibinfo{author}{Newcombe, R.A.}, \bibinfo{author}{Davison, A.J.},
  \bibinfo{year}{2010}.
\newblock \bibinfo{title}{Live dense reconstruction with a single moving
  camera}, in: \bibinfo{booktitle}{Proceedings of the IEEE Conference on
  Computer Vision and Pattern Recognition}, pp. \bibinfo{pages}{1498--1505}.
%Type = Inproceedings
\bibitem[{Newcombe et~al.(2011)Newcombe, Lovegrove and Davison}]{Newcombe2011}
\bibinfo{author}{Newcombe, R.A.}, \bibinfo{author}{Lovegrove, S.J.},
  \bibinfo{author}{Davison, A.J.}, \bibinfo{year}{2011}.
\newblock \bibinfo{title}{{DTAM}: Dense tracking and mapping in real-time}, in:
  \bibinfo{booktitle}{Proceedings of the IEEE International Conference on
  Computer Vision}, pp. \bibinfo{pages}{2320--2327}.
%Type = Article
\bibitem[{Nex and Rinaudo(2011)}]{Nex2011lidar}
\bibinfo{author}{Nex, F.}, \bibinfo{author}{Rinaudo, F.}, \bibinfo{year}{2011}.
\newblock \bibinfo{title}{Lidar or photogrammetry? integration is the answer}.
\newblock \bibinfo{journal}{European Journal of Remote Sensing}
  \bibinfo{volume}{43}, \bibinfo{pages}{107--121}.
%Type = Article
\bibitem[{Ni et~al.(2018)Ni, Li, Liu and Zhou}]{Ni2018second}
\bibinfo{author}{Ni, J.}, \bibinfo{author}{Li, Q.}, \bibinfo{author}{Liu, Y.},
  \bibinfo{author}{Zhou, Y.}, \bibinfo{year}{2018}.
\newblock \bibinfo{title}{Second-order semi-global stereo matching algorithm
  based on slanted plane iterative optimization}.
\newblock \bibinfo{journal}{IEEE Access} \bibinfo{volume}{6},
  \bibinfo{pages}{61735--61747}.
%Type = Article
\bibitem[{\"Ozdemir et~al.(2019)\"Ozdemir, Toschi and
  Remondino}]{Osdemir2019multi}
\bibinfo{author}{\"Ozdemir, E.}, \bibinfo{author}{Toschi, I.},
  \bibinfo{author}{Remondino, F.}, \bibinfo{year}{2019}.
\newblock \bibinfo{title}{A multi-purpose benchmark for photogrammetric urban
  3d reconstruction in a controlled environment}.
\newblock \bibinfo{journal}{The International Archives of the Photogrammetry,
  Remote Sensing and Spatial Information Sciences} \bibinfo{volume}{XLII-1/W2},
  \bibinfo{pages}{53--60}.
%Type = Article
\bibitem[{Poggi et~al.(2020)Poggi, Tosi and Mattoccia}]{Poggi2020learning}
\bibinfo{author}{Poggi, M.}, \bibinfo{author}{Tosi, F.},
  \bibinfo{author}{Mattoccia, S.}, \bibinfo{year}{2020}.
\newblock \bibinfo{title}{Learning a confidence measure in the disparity domain
  from {O(1)} features}.
\newblock \bibinfo{journal}{Computer Vision and Image Understanding}
  \bibinfo{volume}{193}, \bibinfo{pages}{102905}.
%Type = Article
\bibitem[{Pollefeys et~al.(2008)Pollefeys, Nist{\'e}r, Frahm, Akbarzadeh,
  Mordohai, Clipp, Engels, Gallup, Kim, Merrell, Salmi, Sinha, Talton, Wang,
  Yang, Stew{\'e}nius, Yang, Welch and Towles}]{Pollefeys2008}
\bibinfo{author}{Pollefeys, M.}, \bibinfo{author}{Nist{\'e}r, D.},
  \bibinfo{author}{Frahm, J.M.}, \bibinfo{author}{Akbarzadeh, A.},
  \bibinfo{author}{Mordohai, P.}, \bibinfo{author}{Clipp, B.},
  \bibinfo{author}{Engels, C.}, \bibinfo{author}{Gallup, D.},
  \bibinfo{author}{Kim, S.J.}, \bibinfo{author}{Merrell, P.},
  \bibinfo{author}{Salmi, C.}, \bibinfo{author}{Sinha, S.},
  \bibinfo{author}{Talton, B.}, \bibinfo{author}{Wang, L.},
  \bibinfo{author}{Yang, Q.}, \bibinfo{author}{Stew{\'e}nius, H.},
  \bibinfo{author}{Yang, R.}, \bibinfo{author}{Welch, G.},
  \bibinfo{author}{Towles, H.}, \bibinfo{year}{2008}.
\newblock \bibinfo{title}{Detailed real-time urban 3d reconstruction from
  video}.
\newblock \bibinfo{journal}{International Journal of Computer Vision}
  \bibinfo{volume}{78}, \bibinfo{pages}{143--167}.
%Type = Inproceedings
\bibitem[{Remondino et~al.(2013)Remondino, Spera, Nocerino, Menna, Nex and
  Gonizzi-Barsanti}]{Remondino2013dense}
\bibinfo{author}{Remondino, F.}, \bibinfo{author}{Spera, M.G.},
  \bibinfo{author}{Nocerino, E.}, \bibinfo{author}{Menna, F.},
  \bibinfo{author}{Nex, F.}, \bibinfo{author}{Gonizzi-Barsanti, S.},
  \bibinfo{year}{2013}.
\newblock \bibinfo{title}{Dense image matching: Comparisons and analyses}, in:
  \bibinfo{booktitle}{Proceedings of the Digital Heritage International
  Congress}, pp. \bibinfo{pages}{47--54}.
%Type = Article
\bibitem[{Restas(2015)}]{Restas2015drone}
\bibinfo{author}{Restas, A.}, \bibinfo{year}{2015}.
\newblock \bibinfo{title}{Drone applications for supporting disaster
  management}.
\newblock \bibinfo{journal}{World Journal of Engineering and Technology}
  \bibinfo{volume}{3}, \bibinfo{pages}{316--321}.
%Type = Inproceedings
\bibitem[{Ronneberger et~al.(2015)Ronneberger, Fischer and
  Brox}]{Ronneberger2015unet}
\bibinfo{author}{Ronneberger, O.}, \bibinfo{author}{Fischer, P.},
  \bibinfo{author}{Brox, T.}, \bibinfo{year}{2015}.
\newblock \bibinfo{title}{{U-Net}: Convolutional networks for biomedical image
  segmentation}, in: \bibinfo{booktitle}{Proceedings of the International
  Conference on Medical Image Computing and Computer-Assisted Intervention},
  pp. \bibinfo{pages}{234--241}.
%Type = Article
\bibitem[{Roth and Mayer(2019)}]{Roth2019reduction}
\bibinfo{author}{Roth, L.}, \bibinfo{author}{Mayer, H.}, \bibinfo{year}{2019}.
\newblock \bibinfo{title}{Reduction of the fronto-parallel bias for
  wide-baseline semi-global matching}.
\newblock \bibinfo{journal}{ISPRS Annals of the Photogrammetry, Remote Sensing
  and Spatial Information Sciences} \bibinfo{volume}{IV-2/W5},
  \bibinfo{pages}{69--76}.
%Type = Inproceedings
\bibitem[{Rothermel et~al.(2012)Rothermel, Wenzel, Fritsch and
  Haala}]{Rothermel2012}
\bibinfo{author}{Rothermel, M.}, \bibinfo{author}{Wenzel, K.},
  \bibinfo{author}{Fritsch, D.}, \bibinfo{author}{Haala, N.},
  \bibinfo{year}{2012}.
\newblock \bibinfo{title}{{SURE}: Photogrammetric surface reconstruction from
  imagery}, in: \bibinfo{booktitle}{Proceedings of the LowCost3D Workshop}.
%Type = Article
\bibitem[{Ruf et~al.(2017)Ruf, Erdnuess and Weinmann}]{Ruf2017cross}
\bibinfo{author}{Ruf, B.}, \bibinfo{author}{Erdnuess, B.},
  \bibinfo{author}{Weinmann, M.}, \bibinfo{year}{2017}.
\newblock \bibinfo{title}{Determining plane-sweep sampling points in image
  space using the cross-ratio for image-based depth estimation}.
\newblock \bibinfo{journal}{The International Archives of the Photogrammetry,
  Remote Sensing and Spatial Information Sciences} \bibinfo{volume}{XLII-2/W6},
  \bibinfo{pages}{325--332}.
%Type = Article
\bibitem[{Ruf et~al.(2021)Ruf, Mohrs, Weinmann, Hinz and
  Beyerer}]{Ruf2021restac}
\bibinfo{author}{Ruf, B.}, \bibinfo{author}{Mohrs, J.},
  \bibinfo{author}{Weinmann, M.}, \bibinfo{author}{Hinz, S.},
  \bibinfo{author}{Beyerer, J.}, \bibinfo{year}{2021}.
\newblock \bibinfo{title}{{ReS$^2$tAC} -- {UAV}-borne real-time {SGM} stereo
  optimized for embedded {ARM} and {CUDA} devices}.
\newblock \bibinfo{journal}{Sensors} \bibinfo{volume}{21},
  \bibinfo{pages}{3938}.
%Type = Article
\bibitem[{Ruf et~al.(2019)Ruf, Pollok and Weinmann}]{Ruf2019efficient}
\bibinfo{author}{Ruf, B.}, \bibinfo{author}{Pollok, T.},
  \bibinfo{author}{Weinmann, M.}, \bibinfo{year}{2019}.
\newblock \bibinfo{title}{Efficient surface-aware semi-global matching with
  multi-view plane-sweep sampling}.
\newblock \bibinfo{journal}{ISPRS Annals of the Photogrammetry, Remote Sensing
  and Spatial Information Sciences} \bibinfo{volume}{IV-2/W7},
  \bibinfo{pages}{137--144}.
%Type = Article
\bibitem[{Scharstein and Szeliski(2002)}]{Scharstein2002}
\bibinfo{author}{Scharstein, D.}, \bibinfo{author}{Szeliski, R.},
  \bibinfo{year}{2002}.
\newblock \bibinfo{title}{A taxonomy and evaluation of dense two-frame stereo
  correspondence algorithms}.
\newblock \bibinfo{journal}{International Journal of Computer Vision}
  \bibinfo{volume}{47}, \bibinfo{pages}{7--42}.
%Type = Inproceedings
\bibitem[{Scharstein et~al.(2017)Scharstein, Taniai and
  Sinha}]{Scharstein2018surface}
\bibinfo{author}{Scharstein, D.}, \bibinfo{author}{Taniai, T.},
  \bibinfo{author}{Sinha, S.N.}, \bibinfo{year}{2017}.
\newblock \bibinfo{title}{Semi-global stereo matching with surface orientation
  priors}, in: \bibinfo{booktitle}{Proceedings of the International Conference
  on 3D Vision}, pp. \bibinfo{pages}{215--224}.
%Type = Inproceedings
\bibitem[{Sch\"{o}nberger and Frahm(2016)}]{Schoenberger2016sfm}
\bibinfo{author}{Sch\"{o}nberger, J.L.}, \bibinfo{author}{Frahm, J.M.},
  \bibinfo{year}{2016}.
\newblock \bibinfo{title}{Structure-from-motion revisited}, in:
  \bibinfo{booktitle}{Proceedings of the IEEE Conference on Computer Vision and
  Pattern Recognition}, pp. \bibinfo{pages}{4104--4113}.
%Type = Inproceedings
\bibitem[{Sch{\"o}nberger et~al.(2016)Sch{\"o}nberger, Zheng, Frahm and
  Pollefeys}]{Schoenberger2016mvs}
\bibinfo{author}{Sch{\"o}nberger, J.L.}, \bibinfo{author}{Zheng, E.},
  \bibinfo{author}{Frahm, J.M.}, \bibinfo{author}{Pollefeys, M.},
  \bibinfo{year}{2016}.
\newblock \bibinfo{title}{Pixelwise view selection for unstructured multi-view
  stereo}, in: \bibinfo{booktitle}{Proceedings of the European Conference on
  Computer Vision}, pp. \bibinfo{pages}{501--518}.
%Type = Inproceedings
\bibitem[{Sch\"{o}ps et~al.(2017)Sch\"{o}ps, Sch\"{o}nberger, Galliani,
  Sattler, Schindler, Pollefeys and Geiger}]{Schoeps2017eth3d}
\bibinfo{author}{Sch\"{o}ps, T.}, \bibinfo{author}{Sch\"{o}nberger, J.},
  \bibinfo{author}{Galliani, S.}, \bibinfo{author}{Sattler, T.},
  \bibinfo{author}{Schindler, K.}, \bibinfo{author}{Pollefeys, M.},
  \bibinfo{author}{Geiger, A.}, \bibinfo{year}{2017}.
\newblock \bibinfo{title}{A multi-view stereo benchmark with high-resolution
  images and multi-camera videos}, in: \bibinfo{booktitle}{Proceedings of the
  IEEE Conference on Computer Vision and Pattern}, pp.
  \bibinfo{pages}{3260--3269}.
%Type = Inproceedings
\bibitem[{Sinha et~al.(2014)Sinha, Scharstein and Szeliski}]{Sinha2014}
\bibinfo{author}{Sinha, S.N.}, \bibinfo{author}{Scharstein, D.},
  \bibinfo{author}{Szeliski, R.}, \bibinfo{year}{2014}.
\newblock \bibinfo{title}{Efficient high-resolution stereo matching using local
  plane sweeps}, in: \bibinfo{booktitle}{Proceedings of the IEEE Conference on
  Computer Vision and Pattern Recognition}, pp. \bibinfo{pages}{1582--1589}.
%Type = Inproceedings
\bibitem[{Sinha et~al.(2009)Sinha, Steedly and Szeliski}]{Sinha2009}
\bibinfo{author}{Sinha, S.N.}, \bibinfo{author}{Steedly, D.},
  \bibinfo{author}{Szeliski, R.}, \bibinfo{year}{2009}.
\newblock \bibinfo{title}{Piecewise planar stereo for image-based rendering},
  in: \bibinfo{booktitle}{Proceedings of the IEEE International Conference on
  Computer Vision}, pp. \bibinfo{pages}{1881--1888}.
%Type = Inproceedings
\bibitem[{Spangenberg et~al.(2014)Spangenberg, Langner, Adfeldt and
  Rojas}]{Spangenberg2014}
\bibinfo{author}{Spangenberg, R.}, \bibinfo{author}{Langner, T.},
  \bibinfo{author}{Adfeldt, S.}, \bibinfo{author}{Rojas, R.},
  \bibinfo{year}{2014}.
\newblock \bibinfo{title}{Large scale semi-global matching on the {CPU}}, in:
  \bibinfo{booktitle}{Proceedings of the IEEE Intelligent Vehicles Symposium},
  pp. \bibinfo{pages}{195--201}.
%Type = Article
\bibitem[{Szeliski and Scharstein(2004)}]{Szeliski2004}
\bibinfo{author}{Szeliski, R.}, \bibinfo{author}{Scharstein, D.},
  \bibinfo{year}{2004}.
\newblock \bibinfo{title}{Sampling the disparity space image}.
\newblock \bibinfo{journal}{IEEE Transactions on Pattern Analysis and Machine
  Intelligence} \bibinfo{volume}{26}, \bibinfo{pages}{419--425}.
%Type = Phdthesis
\bibitem[{Wenzel(2016)}]{Wenzel2016dense}
\bibinfo{author}{Wenzel, K.}, \bibinfo{year}{2016}.
\newblock \bibinfo{title}{Dense image matching for close range photogrammetry}.
\newblock Ph.D. thesis. University of Stuttgart, Germany.
%Type = Article
\bibitem[{Wenzel et~al.(2013a)Wenzel, Rothermel, Fritsch and
  Haala}]{Wenzel2013image}
\bibinfo{author}{Wenzel, K.}, \bibinfo{author}{Rothermel, M.},
  \bibinfo{author}{Fritsch, D.}, \bibinfo{author}{Haala, N.},
  \bibinfo{year}{2013}a.
\newblock \bibinfo{title}{Image acquisition and model selection for multi-view
  stereo}.
\newblock \bibinfo{journal}{The International Archives of the Photogrammetry,
  Remote Sensing and Spatial Information Sciences} \bibinfo{volume}{40},
  \bibinfo{pages}{251--258}.
%Type = Inproceedings
\bibitem[{Wenzel et~al.(2013b)Wenzel, Rothermel, Haala and
  Fritsch}]{Wenzel2013Sure}
\bibinfo{author}{Wenzel, K.}, \bibinfo{author}{Rothermel, M.},
  \bibinfo{author}{Haala, N.}, \bibinfo{author}{Fritsch, D.},
  \bibinfo{year}{2013}b.
\newblock \bibinfo{title}{{SURE} -- the {IFP} software for dense image
  matching}, in: \bibinfo{booktitle}{Proceedings of the Photogrammetric Week},
  pp. \bibinfo{pages}{59--70}.
%Type = Inproceedings
\bibitem[{Yan et~al.(2020)Yan, Wei, Yi, Ding, Zhang, Chen, Wang and
  Tai}]{Yan2020dense}
\bibinfo{author}{Yan, J.}, \bibinfo{author}{Wei, Z.}, \bibinfo{author}{Yi, H.},
  \bibinfo{author}{Ding, M.}, \bibinfo{author}{Zhang, R.},
  \bibinfo{author}{Chen, Y.}, \bibinfo{author}{Wang, G.}, \bibinfo{author}{Tai,
  Y.W.}, \bibinfo{year}{2020}.
\newblock \bibinfo{title}{Dense hybrid recurrent multi-view stereo net with
  dynamic consistency checking}, in: \bibinfo{booktitle}{Proceedings of the
  European Conference on Computer Vision}, pp. \bibinfo{pages}{674--689}.
%Type = Inproceedings
\bibitem[{Yao et~al.(2018)Yao, Luo, Li, Fang and Quan}]{Yao2018mvsnet}
\bibinfo{author}{Yao, Y.}, \bibinfo{author}{Luo, Z.}, \bibinfo{author}{Li, S.},
  \bibinfo{author}{Fang, T.}, \bibinfo{author}{Quan, L.}, \bibinfo{year}{2018}.
\newblock \bibinfo{title}{{MVSNet}: Depth inference for unstructured multi-view
  stereo}, in: \bibinfo{booktitle}{Proceedings of the European Conference on
  Computer Vision}, pp. \bibinfo{pages}{767--783}.
%Type = Inproceedings
\bibitem[{Yao et~al.(2019)Yao, Luo, Li, Shen, Fang and Quan}]{Yao2019recurrent}
\bibinfo{author}{Yao, Y.}, \bibinfo{author}{Luo, Z.}, \bibinfo{author}{Li, S.},
  \bibinfo{author}{Shen, T.}, \bibinfo{author}{Fang, T.},
  \bibinfo{author}{Quan, L.}, \bibinfo{year}{2019}.
\newblock \bibinfo{title}{Recurrent {MVSNet} for high-resolution multi-view
  stereo depth inference}, in: \bibinfo{booktitle}{Proceedings of the IEEE/CVF
  Conference on Computer Vision and Pattern Recognition}, pp.
  \bibinfo{pages}{5525--5534}.
%Type = Inproceedings
\bibitem[{Yi et~al.(2020)Yi, Wei, Ding, Zhang, Chen, Wang and
  Tai}]{Yi2020pyramid}
\bibinfo{author}{Yi, H.}, \bibinfo{author}{Wei, Z.}, \bibinfo{author}{Ding,
  M.}, \bibinfo{author}{Zhang, R.}, \bibinfo{author}{Chen, Y.},
  \bibinfo{author}{Wang, G.}, \bibinfo{author}{Tai, Y.W.},
  \bibinfo{year}{2020}.
\newblock \bibinfo{title}{Pyramid multi-view stereo net with self-adaptive view
  aggregation}, in: \bibinfo{booktitle}{Proceedings of the European Conference
  on Computer Vision}, pp. \bibinfo{pages}{766--782}.
%Type = Inproceedings
\bibitem[{Zabih and Woodfill(1994)}]{Zabih1994}
\bibinfo{author}{Zabih, R.}, \bibinfo{author}{Woodfill, J.},
  \bibinfo{year}{1994}.
\newblock \bibinfo{title}{Non-parametric local transforms for computing visual
  correspondence}, in: \bibinfo{booktitle}{Proceedings of the European
  Conference on Computer Vision}, pp. \bibinfo{pages}{151--158}.
%Type = Article
\bibitem[{Zbontar and LeCun(2016)}]{Zbontar2016stereo}
\bibinfo{author}{Zbontar, J.}, \bibinfo{author}{LeCun, Y.},
  \bibinfo{year}{2016}.
\newblock \bibinfo{title}{Stereo matching by training a convolutional neural
  network to compare image patches.}
\newblock \bibinfo{journal}{Journal of Machine Learning Research}
  \bibinfo{volume}{17}, \bibinfo{pages}{2287--2318}.
%Type = Inproceedings
\bibitem[{Zhao et~al.(2020)Zhao, Liang, Feng, Ding, Sinha, Zhang and
  Shen}]{Zhao2020fp}
\bibinfo{author}{Zhao, J.}, \bibinfo{author}{Liang, T.}, \bibinfo{author}{Feng,
  L.}, \bibinfo{author}{Ding, W.}, \bibinfo{author}{Sinha, S.},
  \bibinfo{author}{Zhang, W.}, \bibinfo{author}{Shen, S.},
  \bibinfo{year}{2020}.
\newblock \bibinfo{title}{{FP-Stereo}: Hardware-efficient stereo vision for
  embedded applications}, in: \bibinfo{booktitle}{Proceedings of the IEEE
  International Conference on Field-Programmable Logic and Applications}, pp.
  \bibinfo{pages}{269--276}.

\end{thebibliography}
